\documentclass[11pt,twocolumn]{article}
\usepackage{times}

\usepackage{hyperref}

\usepackage{graphicx}
\usepackage{amsmath}
\usepackage{amssymb}
\usepackage{booktabs}
\usepackage{nicefrac}
\usepackage{algorithm}
\usepackage[algo2e]{algorithm2e}
\usepackage{multirow}
\usepackage{mwe}
\usepackage{url}
\usepackage{array}
\usepackage{xifthen} 
\usepackage{soul}
\usepackage[total={19cm, 9in}]{geometry}
\usepackage{setspace}

\usepackage[capitalize]{cleveref}
\crefname{section}{Sec.}{Secs.}
\Crefname{section}{Section}{Sections}
\Crefname{table}{Table}{Tables}
\crefname{table}{Tab.}{Tabs.}

\newcommand{\igwh}[3]{\includegraphics[width=#1, height=#2]{#3}}
\newcommand{\igw}[2]{\includegraphics[width=#1]{#2}}
\newcommand{\igh}[2]{\includegraphics[height=#1]{#2}}
\newcommand{\etal}{et al.}

\newcommand{\CustomCite}[3][\etal]{%
  \ifthenelse{\isempty{#1}}%
    {#2 \cite{#3}}    
    {#2 #1 \cite{#3}} 
}

\usepackage{comment}
\usepackage{color,soul}

\title{\Large\bf Pola4All: survey of polarimetric applications and an open-source \\toolkit to analyze polarization}

\author{
Joaquin Rodriguez$^{~1}$ \hspace{0.2in}
Lew-Fock-Chong Lew-Yan-Voon$^{~1}$ \\
Renato Martins$^{~2,3}$   \hspace{0.2in}
Olivier Morel$^{~1}$ \vspace{0.05in}\\
$^1$Université de Bourgogne, ImViA UR 7535\hspace{0.2in}
$^2$Université de Bourgogne, CNRS UMR 6303 ICB \hspace{0.04in}\\
$^3$Université de Lorraine, CNRS, Inria, LORIA\vspace{0.05in} \\
\small{Emails: {\tt \{joaquin-jorge.rodriguez,lew.lew-yan-voon,renato.martins,olivier.morel\}@u-bourgogne.fr}}
}

\newcommand{\changes}[1]{#1}

\usepackage{fancyhdr}
\fancypagestyle{firstpage}{
  \fancyhf{} 
  \fancyhead[C]{This is a preprint of a published article in Journal of Electronic Imaging (JEI).\\ (The final publication will be available soon)} 
  \headsep = 0.7cm
  \fancyfoot[C]{1}
}

\begin{document}

\date{}
\maketitle

\thispagestyle{firstpage}
\begin{abstract}
Polarization information of the light can provide rich cues for computer vision
and scene understanding tasks, such as the type of material, pose, and shape of
the objects. With the advent of new and cheap polarimetric sensors, this imaging
modality is becoming accessible to a wider public for solving problems such as
pose estimation, 3D reconstruction, underwater navigation, and depth estimation.
However, we observe several limitations regarding the usage of this
sensorial modality, as well as a lack of standards and publicly available
tools to analyze polarization images. Furthermore, although polarization camera
manufacturers usually provide acquisition tools to interface with their cameras,
they rarely include processing algorithms that make use of the polarization
information. In this paper, we review recent advances in applications that
involve polarization imaging, including a comprehensive survey of recent
advances on polarization for vision and robotics perception tasks. We also
introduce a complete software toolkit that provides common standards to
communicate with and process information from most of the existing micro-grid
polarization cameras on the market. The toolkit also implements several image
processing algorithms for this modality, and it is publicly available on GitHub:
\url{https://github.com/vibot-lab/Pola4all_JEI_2023}.
\end{abstract}




\begin{spacing}{1}   
\section{Introduction}
The polarization of the light is present in several real-world physical
phenomena. Light coming from rainbows in the sky, reflections from water on
highways, and monitors and cellphones based on LCD screens are typical examples.
Light polarization is naturally generated when an unpolarized light source (e.g.,
light bulbs or the sun) hits a surface and is reflected. The polarized light
generated in this way can be of two types: specular, when the reflection of the
light is in a single direction; or diffuse, when the reflection is in all
directions. The way the reflected wave oscillates depends on the characteristics
and the shape of the material. This relationship between the observed light and
the object properties is a key feature that vision algorithms can use to improve
their accuracy with respect to another one that uses only texture information.
These additional features can be leveraged, for instance, to improve object
detection and scene segmentation results, detect mirrors and other surfaces that
polarize the light, or uniquely identify places in a room that will serve as
landmarks in navigation algorithms. It is worth noting that polarization cues
are the main sources of information used by many biological agents such as
insects and bees for their orientation in space\cite{BioInspiredPola}.

The introduction of micro-grid polarization sensors, such as Sony Polarsens,
boosted the research in the polarization domain since they are capable of
capturing \changes{the intensity, the color and the linear polarization information} in a
single snapshot, and they also allow measurements outside laboratory conditions.
However, the number of approaches leveraging polarization for
performing computer vision and robotics tasks is, unfortunately, still quite limited. For these
reasons, to promote the usage of polarization cameras in robotics and computer
vision tasks, we provide in this paper a comprehensive review of the latest
advances in the field of polarization imaging. The revised papers have been
chosen to show the potential of this modality to improve the results given by
RGB-only methods, especially in challenging situations, e.g., in the presence of
transparent and textureless objects. Furthermore, to push even more the research
in this field, we break the barrier of the practical problems by introducing a
complete acquisition and processing software toolkit. This toolkit comes with a
graphical user interface, and is capable of processing images coming from
commonly available RGB-polarization sensors, such as the Sony Polarsens,
regardless the camera manufacturer. The aim of the software is to provide
standard acquisition and processing tools for researchers and practitioners in
the field to facilitate the visualization and analysis of polarization images.
In summary, the main contributions of this paper are:

\begin{itemize}
    \item A comprehensive review of the most recent advances in the
        polarization imaging field for computer vision tasks. The selected works
        include the most representative application cases in which the
        polarization information contributes outstandingly to an improvement of
        the results compared to those that would have been obtained with an
        RGB-only camera.

    \item We provide a complete acquisition and processing toolkit, including a
        graphical user interface to fill the gap of a lack of a common software
        for RGB-polarization cameras. The objective of this tool is twofold. On
        one hand, it is a toolkit that any researcher can use as a base
        framework, to analyze and understand the images obtained by any
        RGB-polarization camera. On the other hand, due to its careful design,
        it enables easy development and inclusion of additional applications. We
        include in this paper some different use cases of the developed software
        with the available features to highlight the potential usages.

\end{itemize}

In what follows, we start with a brief introduction to the theory of the
polarization state of light. We then present the main concepts used by all the
reviewed papers, allowing the reader to better understand the contribution of
polarization for each approach. Finally, we describe the developed software
toolkit made publicly available to the community.
\section{\label{sec:PolaIntro}Polarization background}
\subsection{Mathematical polarization model}
Light is an electromagnetic wave of high frequency, and when it propagates
through space, it can be defined by its amplitude, its frequency, and the way it
moves as it travels. The intensity of the wave is equivalent to the brightness
of the light, the frequency is equivalent to its color, and the way it moves
transversely to the propagation direction defines its polarization state. Let us
consider the projection of the oscillation of the electric field vector of the
light wave on a plane perpendicular to the propagation direction. The light is
said to be linearly polarized if this projection gives a line. It is said to be
circularly polarized if the projection describes a circle. If the vector moves
in all directions in a random manner, the light is said to be unpolarized
\cite{born2013principles}. Furthermore, a combination of linearly and
circularly polarized light gives an elliptically polarized wave, and a light
that has both an unpolarized and a polarized component is a partially polarized
light. This last case is the most common type of polarized light that can be
found in the nature.

There exist several models to depict mathematically the polarization state, but
the most commonly adopted one is the Stokes model \cite{PolarizedLightBook}. This
model defines the light wave as a 4D vector
$\mathbf{S}=\left[S_{0},S_{1},S_{2},S_{3}\right]^T$. $S_{0}$ represents the
total light intensity (polarized and unpolarized). $S_{1}$ and $S_{2}$ describe
the amount of light that is linearly polarized horizontally / vertically, and in
the direction of $\pm 45^\circ$, respectively. $S_{3}$ represents the amount of
light that is circularly polarized. Using this model, it is possible to define
two important physical polarization  parameters:
\begin{equation}
    \begin{array}{ccc}
        \rho=\dfrac{\sqrt{S_{1}^2+S_{2}^2+S_{3}^2}}{S_{0}} & \mbox{~and~} & \phi=\dfrac{1}{2}\arctan{\left(\dfrac{S_{2}}{S_{1}}\right)}, \\
    \end{array}
    \label{eq:PolaFromStokes}
\end{equation}
where $\rho$ is called the Degree of Polarization (DoP) which represents the
portion of the light that is polarized, and $\phi$ is the Angle of Polarization
(AoP). This angle represents the orientation \changes{of the line segment}, or of the ellipse,
when the light is respectively linearly or elliptically polarized.
Circularly polarized light is rare in nature\cite{CircPolarizedUnused}, thus
considering only the first three components of the Stokes vector is a good
approximation to model the polarization state. Therefore, in most applications,
the $S_{3}$ component is set to zero, and the Stokes model is represented by a
3D vector of these physical variables as:
\begin{equation}
    \mathbf{S}=\left(\begin{array}{c}
        S_{0} \\
        S_{0} \rho \cos\left(2\phi\right)\\
        S_{0} \rho \sin\left(2\phi\right)\\
    \end{array}\right).
    \label{eq:StokesAndPola}
\end{equation}
If the linear components of the Stokes vector are used, we refer to the AoP and
the DoP as the angle of linear polarization (AoLP), and the degree of linear
polarization (DoLP), respectively.

\subsection{Polarization measurement}
An advantage of the Stokes model is that the effect produced by an object
(either by transmission, by reflection, or by scattering) on the incident wave
can be modeled by a Mueller matrix \textbf{M} \cite{PolarizedLightBook}. More
specifically, a Stokes vector $\mathbf{S}_{in}$ that interacts with an object
whose Mueller matrix is $\mathbf{M}$ is converted into a Stokes vector
$\mathbf{S}_{out}$ according to the following  equation:
\begin{equation}
    \mathbf{S}_{out}=\mathbf{M}\mathbf{S}_{in}.
\end{equation}
If the full-Stokes vector is used, the matrix \textbf{M} has a shape of $4
\times 4$ elements. However, if only the linear part of the Stokes vector is
considered, then this matrix has $3 \times 3$ components. Until now, the only
way to measure the Stokes vector components is by using an indirect measurement
method. For the linear components of the light, this process consists in
measuring the received intensity when the light passes through a Linear
Polarization Filter (LPF) at different orientations. An LPF is an optical device
that allows only the waves that have the same orientation as the filter axis to
pass through. Any other wave is filtered with a gain that has a sine curve
shape. The maximum gain occurs when the filter orientation matches the Angle of
Linear Polarization (AoLP) of the incident light, and the minimum gain will
occur when these angles are separated by $\pi / 2$ radians. An optical device as
the one described is modeled by the following Mueller matrix \cite{Mueller}:
\begin{equation}
    \mathbf{M}=\dfrac{1}{2}\left[\begin{array}{ccc}
        q + r &
        \left(q-r\right)C_{2\theta} &
        \left(q-r\right)S_{2\theta} \\

        \left(q - r\right)C_{2\theta} &
        m_{11} &
        m_{12} \\

        \left(q - r\right)S_{2\theta} &
        m_{21} &
        m_{22} \\
    \end{array}\right],
\end{equation}
where $q$ and $r$ are the major and minor light transmittance of the linear
polarizer, respectively, $\theta$ is the orientation of the filter, and:
\begin{equation*}
    \begin{array}{c}
        S_{2\theta}=\sin\left(2\theta\right) , C_{2\theta}=\cos\left(2\theta\right), \\
        m_{11}=\left(q + r\right)C_{2\theta}^2 + 2\sqrt{qr}S_{2\theta}^2, \\
        m_{22}=\left(q + r\right)S_{2\theta}^2 + 2\sqrt{qr}C_{2\theta}^2, \\
        m_{21}=m_{12}=\left(q + r - 2\sqrt{qr}\right)S_{2\theta}C_{2\theta}. \\
    \end{array}
\end{equation*}
If a camera is used to take the measurements of the filtered light, then, only
the first component of $\mathbf{S}_{out}$ can be retrieved, which corresponds to
the total intensity of the observed light $S_{0}^{out}$. Thus, only the first
line of the Mueller matrix $\mathbf{M}$ should be considered:
\begin{equation}
    S_{0_{\theta}}^{out}=\dfrac{1}{2}\left[\begin{array}{ccc}
        q + r & \left(q-r\right)C_{2\theta} & \left(q-r\right)S_{2\theta} \end{array}\right]
        \mathbf{S_{in}},
    \label{eq:IntenMeasurement}
\end{equation}
where $S_{0_{\theta}}^{out}$ is the $S_{0}$ component of the output Stokes
vector, when the filter axis is oriented at an angle of $\theta$ radians. If the
filter is considered ideal, then $q=1$ and $r=0$, and \cref{eq:IntenMeasurement}
becomes:
\begin{equation}
    S_{0_{\theta}}^{out}=\dfrac{1}{2}\left[\begin{array}{ccc} 1 & \cos\left(2\theta\right) & \sin\left(2\theta\right) \end{array}\right]\mathbf{S_{in}}.
\end{equation}

In general, $S_{0_{\theta}}^{out}=I_{\theta} - d$, where $I_{\theta}$ is the
readout intensity given by a pixel, and $d$ is the pixel offset, often called
dark current. Most works ignore $d$ since in commercial cameras, this
value is negligible compared to the camera measurement\cite{25_Connor}. Thus,
we have in general that:
\begin{equation}
    I_{\theta}=\dfrac{1}{2}\left[\begin{array}{ccc} 1 & \cos\left(2\theta\right) & \sin\left(2\theta\right) \end{array}\right]\mathbf{S_{in}}.
    \label{eq:usedEq}
\end{equation}

\begin{figure*}[!t]
     \centering
     \begin{tabular}{cc}
         \includegraphics[width=3.5cm]{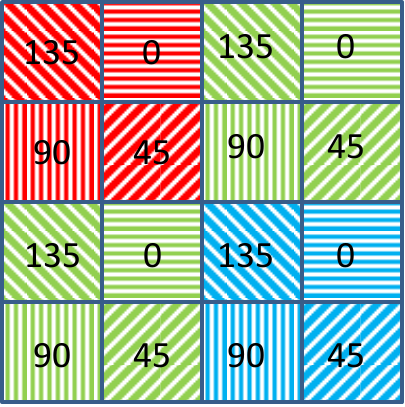} & \hspace{2cm} \includegraphics[width=6cm]{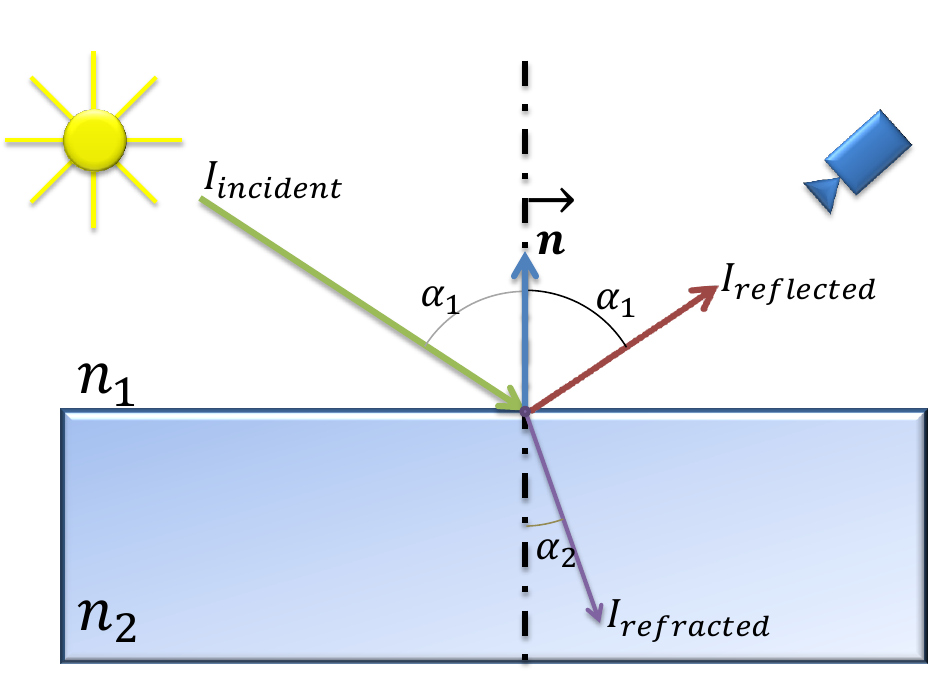} \\
         (a) & \hspace{2cm} (b) \\ 
     \end{tabular}
     \caption[RGB-polarization sensing unit and light-material interaction sketch at the surface level.]
     {(a) Micro-grid sensor scheme with four super-pixels with polarizer orientations of
        $0^\circ$, $45^\circ$, $90^\circ$, and $135^\circ$ arranged following a
        Bayer pattern of a typical RGB-polarization sensor
        \cite{our_calib_paper}. (b) Sketch of the interaction between a light ray coming from the air
        with refractive index $n_{1}$ and an object with refractive index $n_{2}$.
        When the incident light ray hits a point on the surface of the object, a
        portion of the light is reflected and a portion  of it is refracted or
        transmitted in the second medium. The law of reflection states that the
        angle of reflection is equal to the angle of incidence $\alpha_{1}$.
        Regarding the refracted light, its direction of travel will change, and
        it will be equal to $\alpha_{2}$. }
     \label{fig:PixelPattern}
 \end{figure*}

To find the vector $\mathbf{S}_{in}$, several measurements at different angles
$\theta$ are required. This can be done by using a division of focal plane
(DoFP) polarization sensor composed of super-pixels. A super-pixel is a
matrix of $2\times2$ pixels on top of which are four linear polarization
micro-filters, one on each pixel and oriented at angles of $0^\circ$,
$45^\circ$, $90^\circ$, and $135^\circ$. Moreover, to capture the color
information, the super-pixels are organized according to the Bayer pattern where
each of its constitutive pixels shares the same color filter as shown in
\Cref{fig:PixelPattern} (a). This method has the advantage of capturing all the
required information in a single shot: color and polarization. The choice of the
four polarizer orientation values for the super-pixel is demonstrated by
\CustomCite[]{Tyo}{demonstration_microgrid_angles}, in which it concludes that using
equidistant angles in the range $\left[0^\circ,180^\circ\right]$ optimizes the
SNR of the computed Stokes vector.

For a DoFP sensor, four different orientations are used
$\left(\theta_{1},~\theta_{2},~\theta_{3},~\theta_{4}\right)=\left(0^\circ,~45^\circ,~90^\circ,~135^\circ\right)$,
thus four different intensity measurements are obtained
$\left(I_{\theta_{1}},~I_{\theta_{2}},~I_{\theta_{3}},~I_{\theta_{4}}\right)=\left(I_{0^\circ},~I_{45^\circ},~I_{90^\circ},~I_{135^\circ}\right)$.
Then, four expressions of \cref{eq:usedEq} are obtained which when stacked
together gives a linear system in the form $\mathbf{I}=\mathbf{AS}_{in}$ with:

\begin{equation}
\begin{array}{c}
    \left[\begin{array}{c}
        I_{\theta_{1}}   \\
        I_{\theta_{2}}  \\
        I_{\theta_{3}}  \\
        I_{\theta_{4}} \\
    \end{array}\right]=\dfrac{1}{2}\left[\begin{array}{ccc}
        1 & \cos\left(2\theta_{1}\right) & \sin\left(2\theta_{1}\right) \\
        1 & \cos\left(2\theta_{2}\right) & \sin\left(2\theta_{2}\right) \\
        1 & \cos\left(2\theta_{3}\right) & \sin\left(2\theta_{3}\right) \\
        1 & \cos\left(2\theta_{4}\right) & \sin\left(2\theta_{4}\right) \\
    \end{array}\right]\mathbf{S}_{in}, \\
    \end{array}
\end{equation}
where $\mathbf{I}$ is the intensity measurements vector, $\mathbf{A}$ is called
the pixel matrix, and $\mathbf{S}_{in}$ is the Stokes vector we want to
estimate. Then, it is possible to find the Stokes vector $\mathbf{S}_{in}$ by
computing the pseudo-inverse of $\mathbf{A}$, which results in the following
analytical form:
\begin{equation}
    \mathbf{S}_{in}=\left(\begin{array}{c}
        \dfrac{I_{0^\circ}+I_{45^\circ}+I_{90^\circ}+I_{135^\circ}}{2} \\
        I_{0^\circ}-I_{90^\circ} \\
        I_{45^\circ}-I_{135^\circ} \\
    \end{array}
    \right).
    \label{eq:StokesFromI}
\end{equation}

\subsection{\label{sec:fresnel}Naturally generated polarization}
An important property of the polarization state is that it conveys information
about the shape and the \changes{composition} of the objects when they are made of
\changes{dielectric materials}. After hitting a surface, an incident wave will create two new waves
\cite{PolarizedLightBook}: a reflected and a refracted wave. In general, when
a camera observes an object, the captured light is the result of the reflection.

Furthermore, for insulator materials that produce specular light, there is a
single angle in which the reflected light will be 100\% linearly polarized
\cite{PolarizedLightBook}. Any other direction will produce
partially polarized light. The angle at which this type of reflection occurs
depends on the index of refraction, which is a value related to the material
type and the wavelength of the incident light. The interaction of the light, the
material, and the observed intensity is depicted in \Cref{fig:PixelPattern} (b). In
this image, $n_{1}$ and $n_{2}$ are the indexes of refraction of the top and
bottom medium respectively, $\alpha_{1}$ the angle of the reflected light with
respect to the normal vector $\mathbf{n}$ to the surface, and $\alpha_{2}$ the
angle of the refracted light ray with respect to the opposite to the normal
vector. Another important property regarding the reflected light is that the
polarization state is related to the surface orientation. It is for this reason
that, through Fresnel theory \cite{FresnelEquationsBook}, if the ratio
$\lambda=\nicefrac{n_{2}}{n_{1}}$ is known, by measuring the AoP and the DoP, it
is possible to retrieve the normal to the surface at each point of the object.

In summary, the polarization state of the light can provide rich geometric cues
about the shape, pose and material of objects, notably in challenging conditions
to classic imaging approaches, such as in the presence of highly reflective
objects' surfaces (mirrors, windows, water, ... ) or transparent/translucent
objects. \changes{We let the reader refer to these references:}
\cite{FresnelEquationsBook,depth_estimation_under_sun,Zhu_2019_CVPR,PolaRelPosPred}.
\section{Deep-learning background}
\changes{In this section, we introduce the basic concepts of deep learning
algorithms, with the aim of facilitating the understanding of the subsequent
sections of the paper.}

\subsection{Overview of deep-learning networks}
\changes{In the fields of computer vision and robotics, data-driven algorithms are
the most performant ones,  more particularly, deep-learning algorithms. These
methods are based on a network model developed to solve a task through a
training process where  the network would learn to interpret the input data.
Once the network is trained, it can be used on new and unseen data to perform
the intended task.

The training process is an optimization routine in which a set of input data,
known as the training data, is passed through a large equation with coefficients
that can be tuned and the corresponding outputs determined. Next, an error
function, called \textit{the loss function} is computed based on the input data,
the output data, and the constraints with which the output must comply. Finally,
the equation coefficients, or network weights, are updated to try to reduce the
loss function value. This process of updating the weights is called \textit{the
back-propagation algorithm}} \cite{BackProp}. \changes{Once the weights have been
updated, the process is repeated for a certain number of iterations, or epochs.}

\changes{The theory of deep neural networks has existed for several years, but it is
only after Krizhevsky} \etal \cite{FirstDeepLearning} \changes{that these algorithms
have become popular. The authors have shown that Convolutional Neural Networks
(CNN) are capable of outperforming hand-crafted theoretical algorithms. A CNN is
composed of several layers, and each of them includes several convolutional,
optimization, and normalization operations. At the output of each layer, we
obtain the different images called \textit{feature maps}. The deeper the layer
in the network, the higher the level of information it represents. In other
words, the output of the layers close to the network input represents low-level
features (edges, lines, circles), and the output of the layers close to the
output of the network represents high-level features such as object labels
(chair, building, car, dog, etc), depth of the objects in the scene with respect
to the camera coordinate frame, a text description of what the scene represents,
or the pose of an object with respect to the camera frame. In a CNN, a large
number of convolutions are done to the input image, and the training process is
in charge of finding the convolution kernels coefficients to convert the input
image into the expected output image.}

\subsection{Popular developed models}
\changes{In the last decade, several deep neural network algorithms have been
developed. They significantly improved the performance of their predecessors in
various aspects such as achieving higher accuracies, maintaining the same
precision with fewer parameters, and faster training times. Examples of
convolutional neural networks are the VGGNet }\cite{VGG}, \changes{MobileNet}
\cite{MobileNet}, \changes{EfficientNet} \cite{EffNet}, \changes{DeepLabV3}
\cite{DeepLabV3}, \changes{and ResNet} \cite{ResNets}. \changes{Even though there is a
vast variety of models, in general, the models can be divided into two main
blocks: an encoder block and a decoder block. The encoder is composed of layers
designed to extract valuable information from the input data (also called
features). The features are then passed to the decoder for interpretation and
determination of the output data. Many authors use a known encoder block and
tailor a specialized decoder module to meet their specific needs. Additionally,
when dealing with color data, the encoder blocks of the most popular networks
are already included in the deep learning frameworks, and they have already been
pre-trained with the largest datasets available. This way, the optimization
routine will converge faster than training the same model with a tiny dataset.

Very recently, a new type of neural network architecture has been developed,
that challenges the existing convolutional neural networks. It is known as
Transformers, and it has been presented by Vaswani} \etal \cite{Transformers}.
\changes{This new type of network does not use the convolution operation, and it has
been shown to outperform all the known networks in the field of Natural Language
Processing. Furthermore, Dosovitskiy} \etal \cite{Vit}. \changes{showed that
Transformers can also provide more accurate results in the Computer Vision
field. The network is based on what is called \textit{the attention mechanism},
and more precisely, on the \textit{self-attention} mechanism. This method
computes the dot product between the different entries, and the output is an
indication of the importance of the data at a given position for the task.
Therefore, the attention mechanism can be used to ignore the areas where no
important data is present and to keep the regions where the data is valuable to
accomplish the final task. The disadvantage of this type of architecture is that
it requires a large amount of data to outperform CNN. If the dataset used for
training is not large enough, the performance of Transformers is poor.}

\subsection{Polarization and multi-modal networks}
\changes{In data-driven approaches, it is not common to use polarization data
exclusively, either because they are very noisy due to the acquisition process,
or because there are not enough constraints based purely on the polarization
theory for the network to converge. Therefore, polarization data are usually
mixed with other modalities in a multi-modal network. This type of networks take
several sources of data (at least color and polarization), and try to mix them
efficiently. The process of mixing data in this way is called \textit{data
fusion} in the literature, and in general, one of the following three
configurations is used: early, middle, or late fusion.

In the early fusion architecture, the data is mixed before entering the encoder
network for feature extraction. This fusion mechanism is performed in general
through operations that contain learnable parameters as the network, and these
parameters are adjusted during the training process. In late fusion, the data
from each modality passes through different encoders. The weights for each
modality can be shared between the encoders, but it is not a common practice
since each modality extracts different types of patterns from the inputs. Once
the input modalities have passed through the encoders, the high-level feature
maps are fused into a single one that is then passed to the decoder module.
Finally, in the middle fusion architecture, as for the late fusion model, the
data from each modality is inputted to an independent encoder, but in this case,
the feature maps in the encoder at the different hierarchical levels are fused.
In other words, at each network layer of the encoder, the corresponding feature
maps are fused. Middle fusion is the heaviest fusion architecture in terms of
parameters and forward pass time. Nonetheless, it ensures a complete mixture of
the data at all levels in the network, avoiding information loss while the data
passes from one layer to another.

In what follows, we will review the different approaches that make use of the
polarization information in different ways, by using distinct approaches such as
optimization algorithms, hand-crafted features, and data-driven approaches.}

\section{\label{sec:relWorks}Related work}
In this section, we discuss recent advances in the field of polarization imaging
within both model-based and data-driven strategies. \changes{We have reviewed the
latest applications in the computer vision and robotics field that utilize
polarimetry between the years 2016 and 2022. Older applications have not been
considered because the most significant advancements began after the release of
the first micro-grid division-of-focal plane (DoFP) sensor around 2014. Since
then, DoFP sensors have been adopted and widely used as it allows real-time
imaging. In 2018, Sony introduced a DoFP sensor, named PolarSens, which became
the core device of many commercial polarization cameras. It offers much better
quality data and facilitates the development of more performant algorithms and
real-time applications beyond laboratory conditions.}. The works have been
grouped into four representative categories and complementary tasks: image
enhancement, segmentation, surface depth and normal estimation, and pose
estimation. \Cref{tab:PapersSynthesis} shows a summary of the papers included in
each group. \changes{Most reviewed works focus on the application fields of computer
vision and robotic vision. Particularly, we focus our study on the context of
scene understanding in both ground and underwater environments}. Polarization
has also been extensively used in the field of remote sensing, notably in
combination with synthetic aperture Radar (SAR) data. Applications in remote
sensing are not the focus of this survey and a good review of the recent
advances of applications of polarization in this field using data-driven
algorithms can be found in \cite{ReviewPolaSAR}.

\begin{table*}
    \smaller
    \centering
    \begin{tabular}{>{\centering\arraybackslash}ccc}
        \toprule
        \textbf{Survey section } & \textbf{Application} & \textbf{Papers} \\
        \midrule
        \multirow{3}{2.5cm}{\centering Image enhancement} & White balance, Demosaicking & \cite{CVPR_2022_dolp_color_const,EARI,JointChormaticPolaInterpolation,MonochromeDemosaicingPola}\\
        & ~~Dehazing, Reflection removal, Image restoration ~~ & \cite{iplnet,dehazingpaper,semanticnet,Lei_2020_CVPR,UnderwaterRestoration,UnderwaterImgRestoDL,PolaNLOS,UnderwaterRestorat2019,UnderwaterImgRestorat2020} \\
        & Specular-diffuse separation & \cite{PolaDiffuseSpecSep} \\
        \midrule
        \multirow{3}{2.5cm}{\centering Image segmentation} & Material-based segmentation & \cite{CVPR2022MutiModMatSeg} \\
        & Glass segmentation & \cite{glass_segm_paper,TransparentObjectSegmentation} \\
        & Urban scenes segmentation & \cite{RoadDetectionLWIR,RoadDetection,RGBP_SemSeg} \\
        \midrule
        \multirow{3}{2.5cm}{\centering Surface normal and depth estimation} & Normal map estimation &    \cite{Ichikawa_2021_CVPR,normal_pola_in_the_wild_Lei_2022_CVPR,Fukao_2021_CVPR,ECCV_DeepSfP_Ba,Deschaintre_2021_CVPR} \\
        & Depth estimation & \cite{depth_estimation_under_sun,Zhu_2019_CVPR,DepthFromPOlarization_urban_sim,Blanchon2021P2DAS} \\
        & 3D reconstruction / scene rendering & \cite{PolaDenseMapping,MultiViewInvRender,pBRDF}\\
        \midrule
        \multirow{3}{2.5cm}{\centering Pose estimation} & Object's pose & {\centering\cite{PolaRelPosPred,Gao_Pola_pose_pred,HumanPoseEstimation}} \\
        & Optical flow &  \cite{OpticalFlowPola} \\
        & Underwater solar-tracking & \cite{solarTrackingAlgo} \\
        \bottomrule 
    \end{tabular}
    \caption{Summary of the reviewed works (2016 - 2022) of this polarization survey.}
    \label{tab:PapersSynthesis}
\end{table*}

\subsection{Image enhancement}
\begin{figure}[!t]
    \centering
    \begin{tabular}{cccc}
        \vspace{0.1cm}
        (a) &
        \raisebox{-.5\height}{\igwh{2.7cm}{2.7cm}{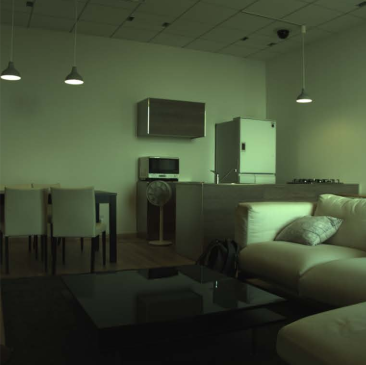}} &
        \raisebox{-.5\height}{\igwh{2.7cm}{2.7cm}{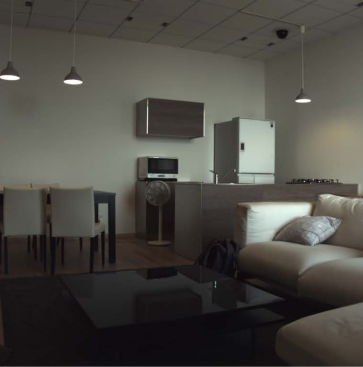}} &
        \raisebox{-.5\height}{\igwh{2.7cm}{2.7cm}{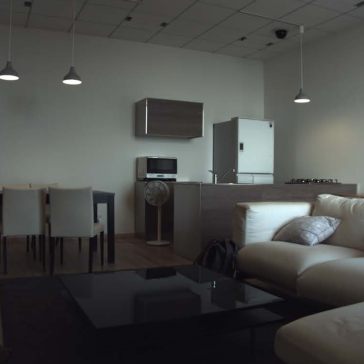}} \\
        \vspace{0.1cm}
        (b) &
        \raisebox{-.5\height}{\igwh{2.7cm}{2.7cm}{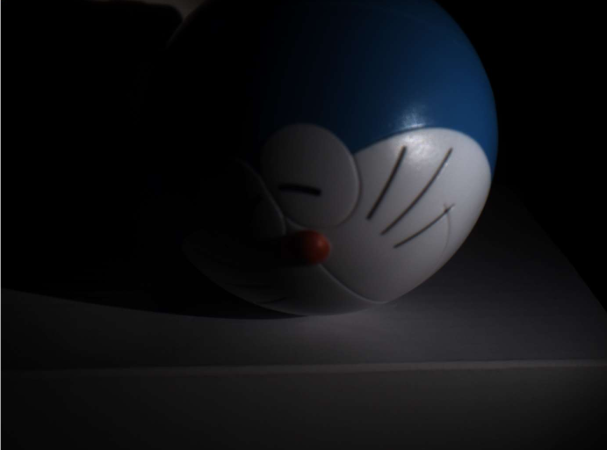}} &
        \raisebox{-.5\height}{\igwh{2.7cm}{2.7cm}{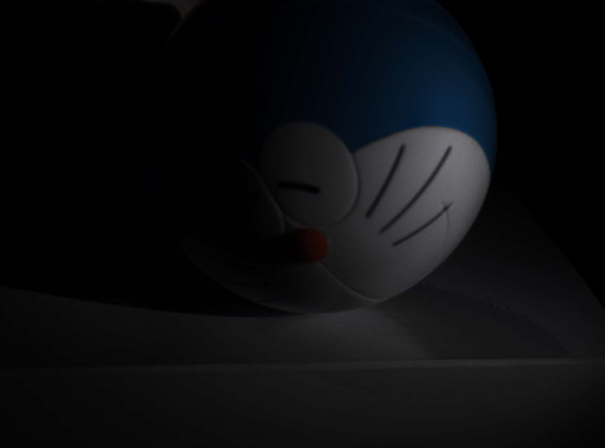}} &
        \raisebox{-.5\height}{\igwh{2.7cm}{2.7cm}{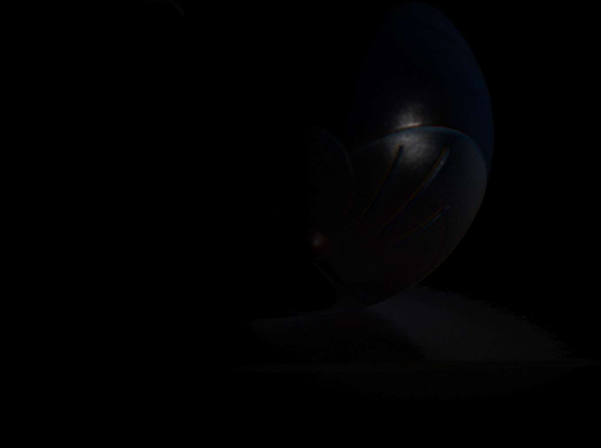}} \\
        \vspace{0.1cm}
        (c) &
        \raisebox{-.5\height}{\igwh{2.7cm}{2.7cm}{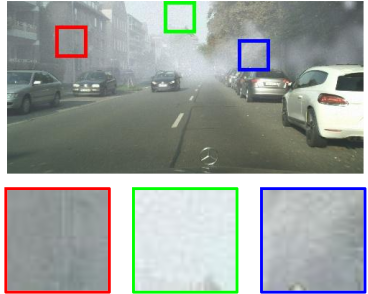}} &
        \raisebox{-.5\height}{\igwh{2.7cm}{2.7cm}{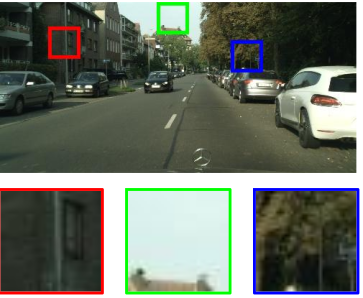}} &
        \raisebox{-.5\height}{\igwh{2.7cm}{2.7cm}{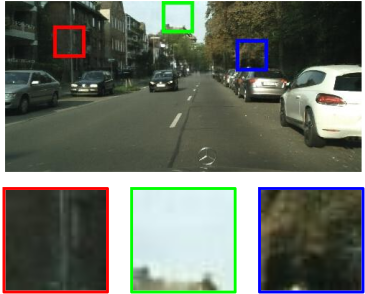}} \\
    \end{tabular}
    \caption[Image enhancement results examples.]
    {Results of some of the reviewed algorithms for image enhancement.
       From left to right:
      (a) Input image, ground truth, and algorithm results of DoLP-based color
        constancy \cite{CVPR_2022_dolp_color_const}.
      (b) Input image, obtained diffuse image, and obtained specular image of
        Polarization-guided specular reflection separation
        \cite{PolaDiffuseSpecSep}.
      (c) Input image, ground truth, and resulting images of the polarization dehazing method
        \cite{dehazingpaper}. Images courtesy of the respective works.}
    \label{fig:ImgEnhancementExam}
\end{figure}

In real-world applications, changes in viewing conditions can strongly impact 
the performance of computer vision algorithms. Thus, enhancing the quality of the
visual information is often a required step to keep the accuracy of the
developed computer vision system. The type of image quality improvement depends
on the application itself. In some cases, this implies having high-quality
measurements independently of the camera used, which can be achieved through
camera calibration \cite{25_Connor,our_calib_paper,Mueller,PowellCalibrab}. In
others, the improvement can be related to removing highly bright, specular
reflections, requiring a separation between this type of reflection from the diffuse ones
\cite{CVPR_2022_dolp_color_const,SpecDiffuseSepCVRPR2023,SpecDiffuseSep2004}. In
more complex cases, the background structure needs to be recovered when an
atmospheric phenomenon is present such as mist or fog, as shown for some
examples in \Cref{fig:ImgEnhancementExam}. In most cases, the physical
constraints defined by the polarization state of the light can be used to
improve the results obtained by conventional cameras.

In this context, \CustomCite{Ono}{CVPR_2022_dolp_color_const} present a white
balance algorithm for RGB-polarization sensors based on the achromaticity of the
Stokes vector in the visible spectrum. \CustomCite{Rodriguez}{our_calib_paper}
relax the experiment setup to calibrate micro-grid color-polarization sensors to
achieve a flat-field response in all the polarization parameter images.
\CustomCite{Wen}{PolaDiffuseSpecSep} solve the separation of
specular from diffuse reflections with a model-based optimization strategy
\cite{ADMM}. Their pipeline is independent of the illumination source by
exploiting polarization and chromaticity images.
\CustomCite{Wen}{JointChormaticPolaInterpolation} propose to jointly demosaic
RGB and polarization information to obtain high-quality, 12-channeled
RGB-polarization images by using a sparse representation model. The model is
obtained through an optimization algorithm that uses the ADMM scheme. Similarly,
\CustomCite{Morimatsu}{EARI} obtain high-quality, polarization images by
extrapolating the residual interpolation for RGB images \cite{origRI} to the
monochrome and color polarization sensors. They achieve their results by
changing the guidance image so that it is edge-aware, and by making use of the
raw polarization intensity measurements. \CustomCite{Tanaka}{PolaNLOS} achieve
better quality images by improving the condition number of the transport matrix
in comparison with conventional, passive Non-Line-Of-Sight (NLOS) system. This
is done by using the polarization leakage effect model produced by the oblique
reflection over a filter oriented at the Brewster angle
\cite{FresnelEquationsBook} of a wall.

Using hand-crafted theories provides good results when the scene and the effect
to analyze are not complex, since a high-precision mathematical model of the
problem can be established. When this is not possible, data-driven algorithms can be
used, as they have the capability to learn complex theories during training. For
example, they can handle scenes with several objects at the same time, or model effects
for which no known mathematical model exists. Data-driven approaches as
described in \CustomCite{Lei}{Lei_2020_CVPR} have been designed to remove the
reflections produced by different types of glasses by using polarization theory.
The input to their network architecture, composed of pre-trained U-net and
VGG-19 networks, is an image which is a combination of the raw measurements,
split by polarization channel and polarization parameters (I, $\rho$, $\phi$),
as presented in \cref{eq:PolaFromStokes}. \CustomCite{Zhou}{dehazingpaper} use a
single polarization image and a deep learning network composed of two U-net
models and two autoencoders to dehaze urban scenes. \CustomCite{Hu}{iplnet}
developed a data-driven approach and a dataset to increase the brightness and
quality of images under low-illumination conditions. They created a
convolutional neural network that works in two steps based on the raw
measurements of the camera: firstly, an enhancement in the intensity domain for
all the color channels, and then each color channel is treated by a separate
network. \CustomCite{Liu}{semanticnet} proposed a Generative Adversarial Network
(GAN) architecture to fuse the DoLP and the intensity images into a single
intensity image. By dividing the image into background and foreground, the
network fuses these two polarization images into a single image that has an
increased and better contrast than the original intensity image. The results
produced by this network can be used to train other networks, i.e., to perform
an improved data augmentation, in order to obtain models with increased
generalization capabilities than the ones obtained when trained only with the original
intensity images. Despite the outstanding results of data-driven algorithms
with respect to the optimization-based approaches, the quality of these
results depends on the data and the type of model used. Particularly for the data,
if not all the cases have been considered in the images provided during training,
the missing cases might produce in less accurate results during testing.

Several relevant image enhancing approaches have also been proposed to deal with
the challenging scenario of underwater imaging.
\CustomCite{Li}{UnderwaterRestoration} aim to improve the contrast of underwater
images due to turbidity by using polarization and an optimization strategy. They
propose to split the Stokes vector into three contributions (diffuse, specular,
and scattered light), since they claim that the scattering reflection underwater
cannot be neglected. In the same direction,
\CustomCite{Hu}{UnderwaterImgRestoDL} present a novel CNN based on residual
blocks that fuse the polarization features to restore the contrast of underwater
images. By using the raw measurements of three polarization channels of a
monochrome polarization camera, they are able to see through turbid water, and
obtain a clear image of the hidden objects.
\CustomCite{Amer}{UnderwaterRestorat2019} propose a static pipeline to increase
the image quality for underwater applications. Based on the active
cross-polarization technique, and an optimized version of the Dark Channel
Prior, they achieve contrast improvement for underwater imaging, with a single
snapshot in real-time. \CustomCite[and Zhao]{Shen}{UnderwaterImgRestorat2020}
developed an iterative pipeline to jointly improve the image contrast and
denoise the image. With two polarization images taken with a rotative filter at
$0^\circ$ and $90^\circ$, they compute the transmittance and the irradiance maps
in underwater conditions for each color channel. Then, they establish an
iterative process to refine these results by using an adaptative bilateral
filter, and an adaptative color correction routine.

Despite the remarkable improvement brought by polarization to image enhancement,
as compared to  similar applications for RGB-only cameras, several challenges
still remain. For example, \CustomCite{Ono}{CVPR_2022_dolp_color_const}
outperform different baselines in many scenes, but it is left as future work to
improve the results obtained when the sky occupies a large portion of the image.
Similarly, \CustomCite{Zhou}{dehazingpaper} retrieve the hidden structure of
objects behind the haze with good accuracy in real-world situations, after
training the network with computer generated images. Despite this, the authors
claim that the model does not produce adequate reconstructions for fog and mist
due to the physical phenomena produced by these perturbations that are not the
same as for the haze. It is important to note that one of the barriers in the
polarization image enhancement field is the lack of standard benchmarks. Indeed,
most works had to create a dataset to demonstrate their contributions. Some of
the created datasets, which often required a huge amount of work, can be reused
as in \CustomCite{Lei}{Lei_2020_CVPR}, where the authors realized acquisitions
in a large variety of environments, and with different types of glasses. On the
other hand, other applications have been demonstrated using small, non-available
datasets, that may not necessarily cover all the required cases
\cite{JointChormaticPolaInterpolation}, or they have created a polarization
dataset based on RGB ones and a mathematical model of the polarization effect
\cite{dehazingpaper}.

\subsection{Image segmentation}
\begin{figure}[!t]
    \centering
    \begin{tabular}{cccc}
        \vspace{0.1cm}
        (a) &
        \raisebox{-.5\height}{\igwh{2.7cm}{2.7cm}{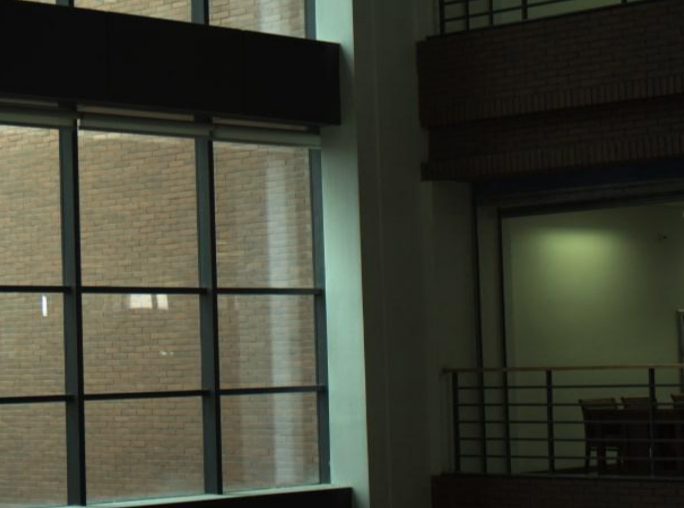}} &
        \raisebox{-.5\height}{\igwh{2.7cm}{2.7cm}{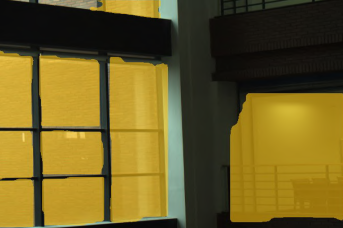}} &
        \raisebox{-.5\height}{\igwh{2.7cm}{2.7cm}{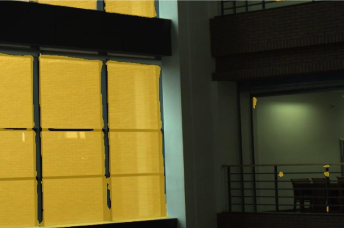}} \\
        \vspace{0.1cm}
        (b) &
        \raisebox{-.5\height}{\igwh{2.7cm}{2.7cm}{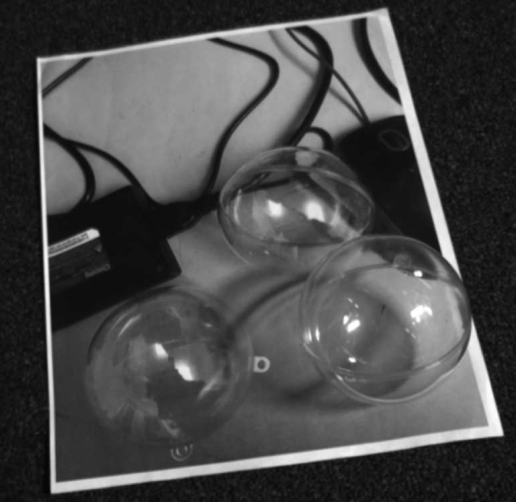}} &
        \raisebox{-.5\height}{\igwh{2.7cm}{2.7cm}{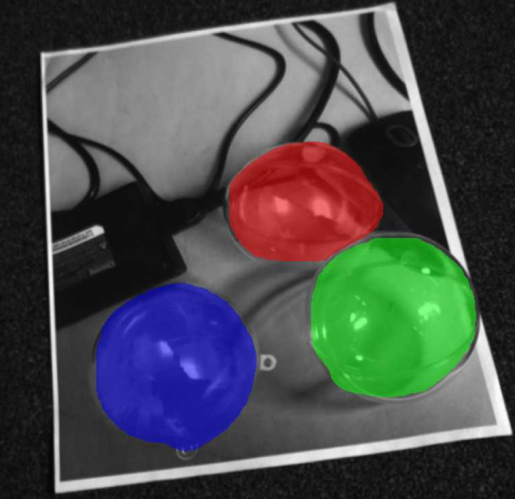}} &
        \raisebox{-.5\height}{\igwh{2.7cm}{2.7cm}{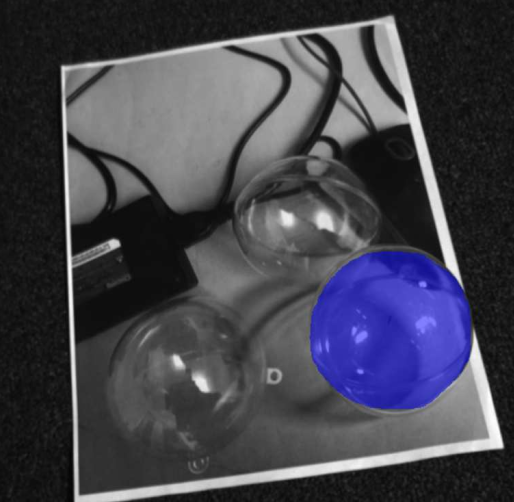}} \\
        \vspace{0.1cm}
        (c) &
        \raisebox{-.5\height}{\igwh{2.7cm}{2.7cm}{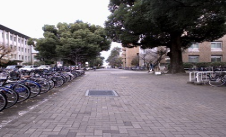}} &
        \raisebox{-.5\height}{\igwh{2.7cm}{2.7cm}{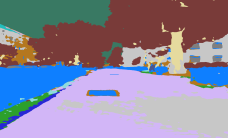}} &
        \raisebox{-.5\height}{\igwh{2.7cm}{2.7cm}{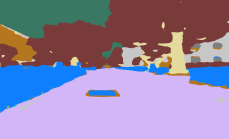}} \\
    \end{tabular}
    \caption[Image segmentation results examples.]
    {Segmentation examples with polarization cues. From left to right:
      (a) Input image, results obtained with an RGB-only method, and the result
        of glass Segmentation using intensity and spectral polarization information\cite{glass_segm_paper}.
      (b) Input image with transparent objects, RGB-only method object
        segmentation results, and results obtained with the polarization data-driven method\cite{TransparentObjectSegmentation}.
      (c) Input image, RGB-only method results, and the results of the
        Multimodal Material Segmentation algorithm
        \cite{CVPR2022MutiModMatSeg}. Images courtesy of the respective works.}
    \label{fig:imgSegmExamples}
\end{figure}
The polarization state of the light is directly linked to the object's material
and shape as introduced in \Cref{sec:fresnel}. This property can provide
insightful and complementary information to guide object segmentation approaches
in scenarios where only the surface color is not discriminant. Indeed, the index
of refraction depends on the internal structure of the objects, and on the
wavelength of the incident light as defined by the Fresnel equations
\cite{born2013principles}. It is for this reason that material classification is
one of the most fruitful application of polarization theory. Previously, this
task was accomplished using hand-crafted features, in controlled acquisition
conditions, using rotative filters, and considering a single object to be
analyzed at a time
\cite{WolffMatClass1990,WolffMatClass1996,MatClassification2008,MatClassif2007}.
Nowadays, with the advances in sensors and the advent of data-driven algorithms,
object characterization with polarization cues has been ported to more complex,
constraint-relaxed scenarios.

In the domain of infra-red imaging, \CustomCite{Li}{RoadDetectionLWIR} succeed
in efficiently detecting the road area in urban scenes by using the
zero-distribution prior in the AoLP and the difference in the DoLP of the
objects to increase the accuracy of the segmentation. This information is
further used in a visual tracking algorithm to continuously track the road
online. \changes{In an extension of their previous work,}
\CustomCite{Li}{RoadDetection} \changes{use the zero-distribution prior of the AoLP
to create a coarse map of the road. Then, they developed a deep-learning network
to refine the coarse road map. Their network consists of two branches that
analyze different aspects of the scene. The main branch receives the information
captured by an infra-red camera, already converted into a fake color image, i.e., a
3-channel image result of stacking the AoLP, the DoLP, and the total intensity
together, and then converted from the HSV to the RGB color space. The objective
of this branch is to extract multi-modal features of the scene. The other branch
or polarization-guided branch also receives the AoLP and the DoLP of the scene,
but it does not receive the intensity image. Instead, the coarse map obtained
from the zero-distribution prior of the AoLP is provided. By doing so, the
authors aim to guide the network based on the polarization properties of the
road, and not of the entire scene}. In the visible spectrum,
\CustomCite{Xiang}{RGBP_SemSeg} developed a fusion network to combine color and
polarization data to better segment objects of urban scenes. They tested several
combinations of polarization information with attention mechanisms and concluded
that using only color and the AoLP is the best combination to improve the
results. \CustomCite{Kalra}{TransparentObjectSegmentation} improve the
instance-semantic segmentation network Mask R-CNN \cite{MaskRCNN} to handle
transparent objects (as shown in \Cref{fig:imgSegmExamples}) by adding
monochrome polarization cues to the original mid-fusion pipeline. Each
polarization parameter image (intensity, AoLP, and DoLP) is fed into a different
backbone encoder, and the fusion of the feature maps is performed at each
encoder level. \CustomCite{Mei}{glass_segm_paper} extend the work presented in
\cite{TransparentObjectSegmentation} by using RGB polarization cues, instead of
monochrome, with the aim of segmenting glasses in urban scenes. Since they use
an RGB-Polarization camera, they can measure RGB intensity, RGB DoLP, and RGB
AoLP. Each of the two RGB polarization images (DoLP and AoLP) \changes{is} balanced
by using an attention mechanism. Then, these two results, and the intensity RGB
image are fed into three independent Conformer encoders \cite{Conformers} and
fused using local and global guidance. In a more complex scenario,
\CustomCite{Liang}{CVPR2022MutiModMatSeg} \changes{build} a network to fuse RGB,
infra-red, and polarization cues to produce an outdoor scene segmentation based
on the object material type. The proposed pipeline is composed of two core
elements: a network that will classify the objects present in one class of a
subset of the segmentation labels from the CityScapes dataset
\cite{Cordts2016Cityscapes}; and a region-based filter selection module that
chooses the modality that provides the most relevant information for determining
the type of material of the constitutive elements of each detected object. The
full network is composed of four encoders, one for the RGB intensity, one for
the AoLP image, one for the DoLP image, and one for the infra-red image.

All these works outperform RGB systems when the polarization information is
added to each developed pipeline. A higher gain in performance is also often
obtained when the network is adapted to correctly process the AoLP and the DoLP,
and not when an RGB network is trained with polarization images. This is why
most of these works propose carefully designed fusion schemes. However, the lack
of datasets including polarization information in the field of image
segmentation poses limitations on the development of polarization-based
approaches. It is important to highlight that all the previously discussed works
have presented their own dataset to show that polarization is a path to consider
in image segmentation. For instance,
\CustomCite{Kalra}{TransparentObjectSegmentation} used a private dataset
acquired in a very specific environment, focused on the particular application
with a pick-and-place robotic arm. \CustomCite{Xiang}{RGBP_SemSeg} provide a
small-scale dataset of RGB-polarization images captured in various urban scenes.
Although informative, the dataset contains only 394 annotated images segmented
into 9 different classes. \CustomCite{Mei}{glass_segm_paper} introduce a
medium-scale dataset, with 4511 images annotated only for the labels glass and
no-glass. Similarly, \CustomCite{Liang}{CVPR2022MutiModMatSeg} made publicly
available a dataset of semantic segmentation of urban scenes, with multi-modal
sensors, but it only includes 500 labeled images, and
\CustomCite{Li}{RoadDetectionLWIR} did it for road segmentation, and with their
personalized infra-red polarization camera. Thus, there is a need for a common
large-scale benchmark to evaluate the performance of these different
segmentation algorithms to trace the direction toward a generalization of the
polarization modality.

\subsection{\label{sec:depth}Surface normal and depth estimation}
\begin{figure}[!t]
    \centering
    \begin{tabular}{cccc}
        (a) &
        \raisebox{-.5\height}{\igwh{2.7cm}{2.7cm}{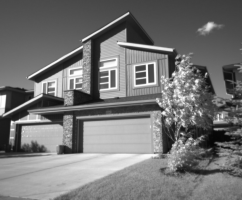}} &
        \raisebox{-.5\height}{\igwh{2.7cm}{2.7cm}{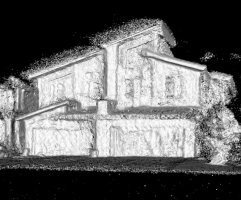}} &
        \raisebox{-.5\height}{\igwh{2.7cm}{2.7cm}{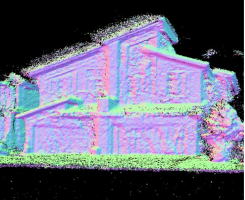}} \\
        \vspace{0.1cm}
        (b) &
        \raisebox{-.5\height}{\igwh{2.7cm}{2.7cm}{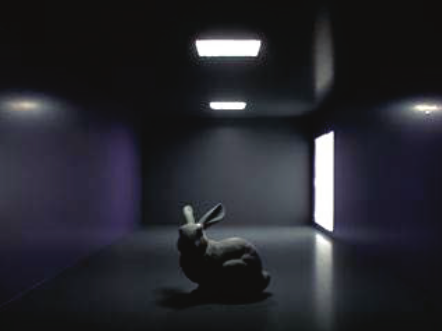}} &
        \raisebox{-.5\height}{\igwh{2.7cm}{2.7cm}{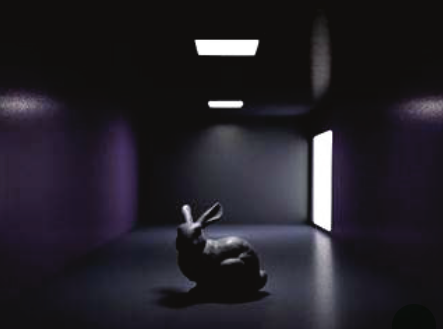}} &
        \raisebox{-.5\height}{\igwh{2.7cm}{2.7cm}{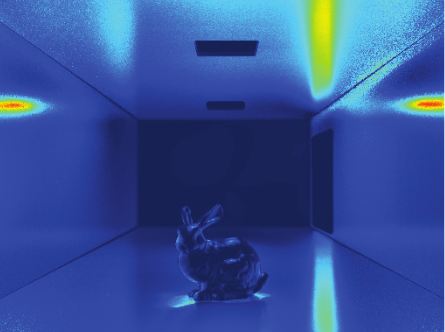}} \\
        (c) &
        \raisebox{-.5\height}{\igwh{2.7cm}{2.7cm}{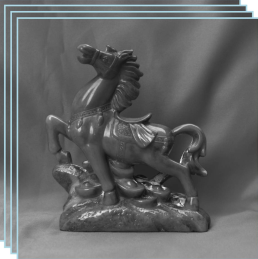}} &
        \raisebox{-.5\height}{\igwh{2.7cm}{2.7cm}{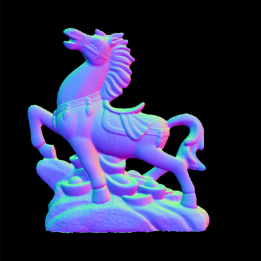}} &
        \raisebox{-.5\height}{\igwh{2.7cm}{2.7cm}{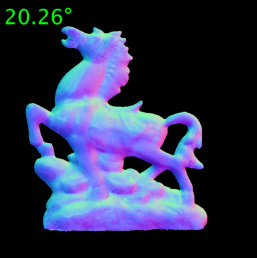}} \\
    \end{tabular}
    \caption[Surface normal and depth estimation results examples.]
    {Results of some of the reviewed algorithms for surface normal and depth estimation.
        From left to right:
        (a) Intensity input image, reconstructed mesh, and estimated normal map
          by the method introduced by \cite{PolaDenseMapping}.
        (b) Real RGB image, corresponding rendered image, and the corresponding rendered
            degree of linear polarization. This result corresponds to the scene
            rendering technique implemented in \cite{pBRDF}.
        (c) Input polarization images, ground truth, and estimated normal maps.
            These results are courtesy of \cite{ECCV_DeepSfP_Ba}.}
    \label{fig:SurfaceDepthEst}
\end{figure}

Polarization is well known to encode shape information of the different objects
being observed. Most existing approaches consider an orthographic projection of
the incoming light, and if the refractive index of the object is known, then the
normal vectors to the surface can be estimated, and through integration, the
depth map can be retrieved. These approaches are again based on the Fresnel's
theory, which has been briefly introduced in \Cref{sec:fresnel}, which
establishes the relationship between the degree of linear polarization with the
zenith angle of the normal, and the angle of linear polarization with the
azimuth angle. Even though effective in several works and scenarios
\cite{SfPIntegration2003,SfPIntegration2005,SfPIntegration2007,SfPIntegration2012},
the fact that both the object's refraction index and the light direction must be
known makes this approach limited to laboratory strict conditions. Additionally,
the relationship between polarization and the normal vector has geometric
ambiguities. Therefore, one important research direction is to reduce the
constraints and priors required for the acquisition while maintaining low
reconstruction errors.

\CustomCite{Ba}{ECCV_DeepSfP_Ba} propose a learning-based approach to estimate
the normal map of objects as shown in \Cref{fig:SurfaceDepthEst}. The ambiguous
normal maps from Fresnel's theory are used as priors given directly to a deep
neural network as inputs. \CustomCite{Fukao}{Fukao_2021_CVPR} present a shape
from polarization algorithm that uses a stereo pair of polarization cameras. The
coarse map from the stereo vision is refined by filtering the normal maps with a
belief propagation scheme. They exploit Fresnel's theory, and an improved
modeling of the micro-facet reflection effect by considering that it is a linear
combination of the diffuse and the specular lobe reflection. Similarly,
\CustomCite{Ichikawa}{Ichikawa_2021_CVPR} relaxed the constraints for shape from
polarization by using the Rayleigh and the Perez models to estimate the sun
polarization state and direction on a clear day. Then, through mathematical
optimization, the normal and shading maps are obtained. The additional cues
about the incident Stokes vector serve to determine how the object modulates the
incident light (Mueller calculus), and jointly with the shading constraints, the
normal map can be estimated. \CustomCite{Deschaintre}{Deschaintre_2021_CVPR}
propose a 3D object shape estimation, jointly with a spatially variable BRDF
model estimation, by using a single-view polarization image fed to a U-Net based
network architecture. The full input of the model is the intensity image, the
normalized Stokes map, and the normalized diffuse color which encodes the object
reflectance information. \CustomCite{Lei}{normal_pola_in_the_wild_Lei_2022_CVPR}
propose a deep-learning network to estimate the normal map of complex scenes.
Their aim is to improve the accuracy limits by incorporating viewing encoding as
input to the network, which accounts for the non-orthographic projection.
\changes{This input is an image where each pixel represents the direction of the
incident light. When estimating the normal vectors from polarization under the
orthographic assumption, all the incident light rays are supposed to be colinear
to the $Z$ axis of the camera coordinate frame. Thus, all the zenith angles are
measured with respect to a common coordinate frame. When using a perspective
lens, the zenith angle given by the polarization theory is measured with respect
to the direction of propagation of the light, which in this case, will be
different for each pixel. By providing the viewing encoding, the authors claim
the network will understand the viewing direction of the polarization state and
use this information to improve the results of a network that works under the
orthographic assumption with a perspective lens}. The other inputs to the model
are the raw measurements of the camera separated by polarization channel, the
AoLP, the DoLP, and the total intensity. Their network is also grounded by an
architecture similar to a U-Net model, with a multi-head self-attention module
in the bottleneck. \CustomCite{Smith}{depth_estimation_under_sun} define the
shape from polarization problem as a large linear system of equations. They
combine the physics theory of polarization with the geometry of the problem to
formulate the depth equations directly, without passing through the computation
of the normals. \CustomCite{Berger}{DepthFromPOlarization_urban_sim} present a
depth estimation algorithm that uses the polarization cues in a stereo vision
system. They improve the correspondence matching by adding the AoLP-normal
constraint to the intensity similarity function. In the same direction,
\CustomCite[and Smith]{Zhu}{Zhu_2019_CVPR} propose a hybrid RGB-polarization
acquisition system to obtain a dense depth reconstruction. By classifying the
pixels into specular or diffuse, they make use of normal vectors obtained from
Fresnel's theory to improve the estimation of the normal maps obtained from the
stereo images. \CustomCite{Blanchon}{Blanchon2021P2DAS} extend the monocular
depth estimation network Monodepthv2 \cite{monodepth2} to consider polarization
information by adding the azimuthal constraint to the deep-learning loss.
\CustomCite{Zhao}{MultiViewInvRender} extend the multi-view reconstruction
system \cite{origMVIR} by adding polarization cues to the optimization. They
introduce a continuous function that has four minimum values, each of them at
one of the ambiguous normal azimuth possibilities of Fresnel's theory.
\CustomCite{Kondo}{pBRDF} developed a polarimetric BRDF model that does not
constrain the illumination nor the camera position during acquisition. This
model is used to synthesize polarization images out of the RGB images, easing
the dataset creation for data-driven algorithms. By acquiring images with
different illumination of known Stokes vectors, they use Mueller calculus to
model the object reflectance. \CustomCite{Shakeri}{PolaDenseMapping} produce a
dense 3D reconstruction by using polarization cues. This is done by optimizing
an initial depth map obtained from MiDaS \cite{MIDAS_depth_estimation}, and the
coarse depth map from COLMAP \cite{COLMAP}. The optimization routine constraints
the normals with the ones from Fresnel's theory. The initial depth map is used
to disambiguate the polarization normals, and the coarse map is used to
regularize the optimization routine, since they are metrically correct but
sparse.

The previously presented works have all been developed for shape/depth
estimation while leveraging the polarization information. Combining the
polarization state of the light with any geometry problem developed for the RGB
space results in a significant improvement in accuracy and image quality. This
is because the polarization measurements are dense since they are provided
pixel-wise, which means that the normal constraints are also dense. Thus,
passive, high-quality far-field 3D reconstructions can be retrieved using a
multi-modal RGB-polarization camera, which cannot be done with active sensors
such as LiDAR or Microsoft Kinect. However, these polarization constraints are
still often dependent on knowing priors about the material (metallic vs
insulator) and the reflection type (specular vs diffuse), thus sometimes they
can produce low-quality results in the wild. To overcome this problem, some
works decide to use only diffuse reflections \cite{Deschaintre_2021_CVPR}, or to
classify the pixels into either diffuse dominant or specular dominant (such as
in \cite{depth_estimation_under_sun,PolaDenseMapping}). To deal with more
complex cases, a better modeling of the reflection effect might be required
\cite{Fukao_2021_CVPR}. For data-driven algorithms, this field also suffers from
the lack of large-scale datasets that can be used as benchmarks for research.
Most papers propose their own dataset by doing acquisitions
\cite{normal_pola_in_the_wild_Lei_2022_CVPR}, or they model the polarization
state assuming artificial conditions to simulate polarization over already
existing RGB images \cite{pBRDF}.

In summary, the polarization clearly provides valuable cues in the field of
shape estimation and 3D reconstruction but two main problems need to be
addressed. Foremost, the lack of large-scale, standard evaluation benchmarks,
that hinders the development of techniques using this modality in the current
era of data-driven algorithms. On the other hand, there is no generic model that
can effectively handle generic types of reflections over any type of material.
Therefore the challenge of interpreting the measured data and determining which
model to use remains open.

\subsection{Pose estimation}
\begin{figure}[!t]
    \centering
    \begin{tabular}{cccc}
        \vspace{0.1cm}
        (a) &
        \raisebox{-.5\height}{\igh{2.7cm}{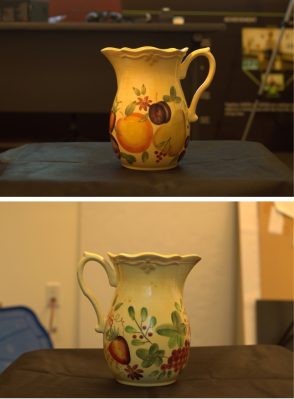}} &
        \raisebox{-.5\height}{\igwh{2.7cm}{2.7cm}{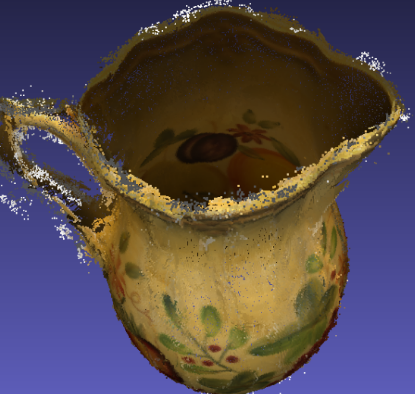}} &
        \raisebox{-.5\height}{\igwh{2.7cm}{2.7cm}{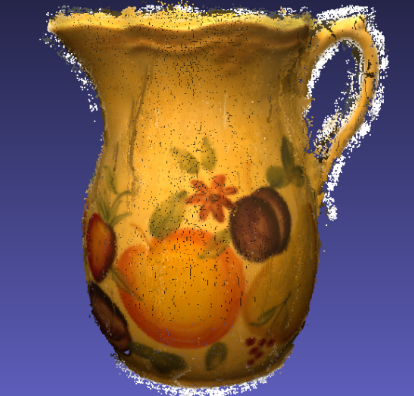}} \\
        \vspace{0.1cm}
        (b) &
        \raisebox{-.5\height}{\igh{2.7cm}{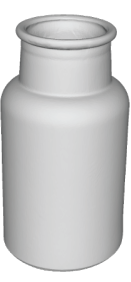}} &
        \raisebox{-.5\height}{\igwh{2.7cm}{2.7cm}{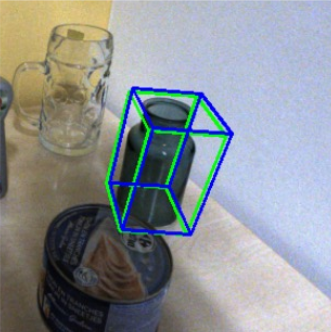}} &
        \raisebox{-.5\height}{\igwh{2.7cm}{2.7cm}{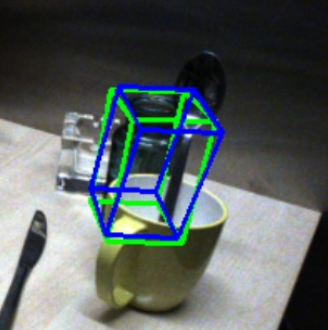}} \\
    \end{tabular}
    \caption[Pose estimation results examples.]
    {Results of some of the reviewed algorithms for pose estimation.
        From left to right:
        (a) Input RGB image, reconstructed surface with the algorithm provided
            in \cite{PolaRelPosPred}.
        (b) Glass vase model, and the detected pose from two viewpoints. Results
            courtesy of \cite{Gao_Pola_pose_pred}.
        }
    \label{fig:PoseEstimation}
\end{figure}

The polarization of light can also play a significant role in object pose
estimation since it provides valuable geometric constraints in the determination
of the vectors normal to the observed surfaces (as discussed in \Cref{sec:depth}).
Hence, this additional information can be used to overcome
the ill-posedness condition of many RGB problems, such as estimating the
relative rotation and translation of textureless objects between two images.
Particularly, these additional constraints are useful when the objects to
analyze are highly reflective and translucent since the polarization
measurements are independent of the intensity of the light.
\CustomCite{Cui}{PolaRelPosPred} use the normal vectors estimated from the
Fresnel equations to add geometric constraints for pose estimation (some
visualizations of the pose estimation can be seen in \Cref{fig:PoseEstimation}).
With this additional information, only two corresponding points in two views are
required to estimate the rotation matrix and the translation vector. In the same
direction, \CustomCite{Gao}{Gao_Pola_pose_pred} propose a data-driven algorithm
to find the pose transformation of an object in the image with respect to the
camera coordinate frame. The algorithm uses three ambiguous normals as inputs to
one encoder, and the polarization parameters into another. Then, the features
are fused at different levels, and given to the decoder. \CustomCite[and
Schechner]{Tzabari}{OpticalFlowPola} present a static approach in which they use
the AoLP and the DoLP to expand the optical flow theory. This new component
accounts for the rotation speed estimation, which cannot be done with the
classical optical-flow approach. \CustomCite{Hu}{solarTrackingAlgo} utilize a
monochrome polarization camera to build a complete pipeline to estimate the
sun's position based on the DoLP and the AoLP measured \changes{underwater}. Jointly using the
Snell and Fresnel theories, they revert the ray bending caused by the change in
medium and handle the problem as if the measurements were done outside the
water. \CustomCite{Zou}{HumanPoseEstimation} push forward the accuracy in the
human shape and pose estimation by building a two steps network with
polarization cues. Assuming  the human cloth to be diffuse dominant, they
retrieve the human features by using the raw polarization intensity images, and
the ambiguous normal maps obtained from Fresnel theory. The first network
produces a high-quality normal map, and the second one uses this result, jointly
with the output of the SMPL human shape model \cite{SMPL_body_model} to
estimate the final shape and pose of a clothed person.

Due to the geometric nature of the pose estimation problem, the polarization
state of the light provides valuable clues that can be used in any computer
vision algorithm in this field. The applications included in this section of the
review demonstrate the potential of accuracy gain obtained with the polarization
constraint, while at the same time relaxing the other algorithms hypothesis. In
\cite{PolaRelPosPred}, it is shown that only two points are required to estimate
the pose transformation between two views, resulting in an improvement in speed
and accuracy when added to any structure from motion algorithm. Without any
requirement in the type of clothes to be used, \cite{HumanPoseEstimation} were
able to estimate the human pose with a lower error and fewer constraints than
competitors.
In underwater applications, \changes{Global Positioning System (GPS)
signals cannot be used because their intensities decrease rapidly
with the depth in the water. To address this issue,} \CustomCite{Hu}{solarTrackingAlgo}
\changes{propose an autonomous underwater navigation system that uses
polarization instead of a GPS signal. In their system, the camera's
global position is estimated by applying geometrical constraints
that link the sun's position to its known trajectory} \cite{PolaGlobalPos1,
PolaGlobalPos2}. However, several limitations still remain when doing pose
estimation with polarization. For example, in \cite{Gao_Pola_pose_pred}, only
the position of one object can be done each time, while others adopt known
object materials and physical properties. Furthermore, most algorithms only
consider one type of reflection (either diffuse or specular), which limits their
generalization to any type of scene.

\section{\label{sec:PolaSoft}Pola4all: Polarization Toolkit}
From the presented study on the advances in polarization imaging, we have
observed that in all the reviewed applications there are no common standards
either for data structure, acquisition, or visualization tools. We can also
notice a lack of a common library to boost the development, evaluation, and
deployment of algorithms using polarization. For these reasons, this section
presents a software library that allows to visualize and analyze polarization
images, as well as to develop and integrate any custom application in a
structured manner. In what follows, we briefly describe the software, including
its basic components and the implemented image-processing algorithms. The
objective is to contribute to the community by providing several tools
integrated with a single, common graphical user interface (GUI) software that
can interact with any DoFP RGB-polarization camera available on the market.
Additionally, this software is meant to provide access to all the developed
algorithms that showcase the power of polarization information, making it easier
for anyone interested in working with polarization modality to get started. The
library has been developed in C++ Object Oriented Programming language and is
built on top of three main libraries/frameworks: OpenCV, ROS and QT5. \changes{An overview of
the final graphical user interface can be found in the GitHub repository of this project.}

\begin{figure}[!t]
    \centering
    \igw{\columnwidth}{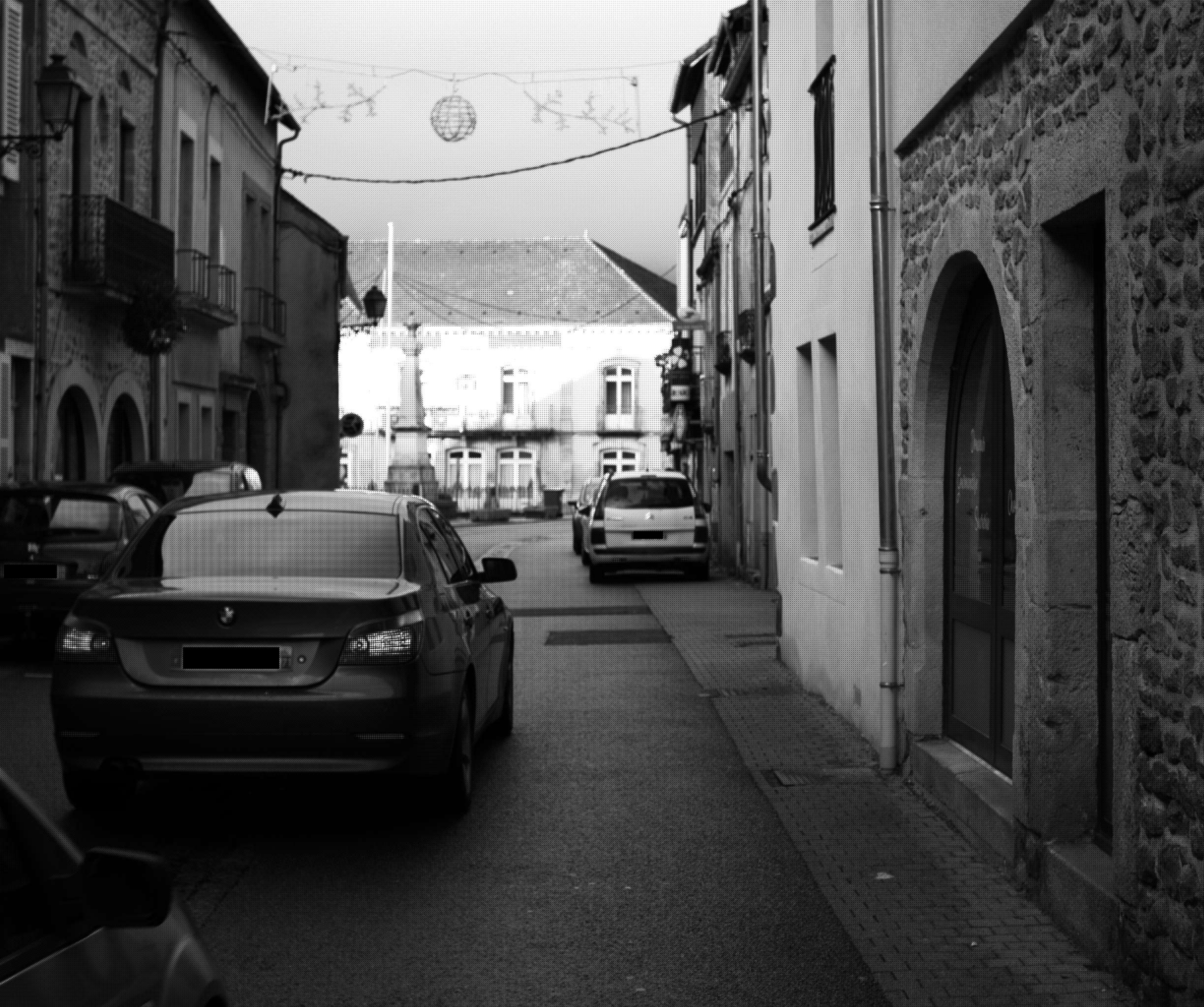}
    \caption[Raw reference image used for the software tests.]
    {Raw image of an urban scene used as a reference for all the
        algorithms implemented in the toolkit Pola4All.}
    \label{fig:UrbanSceneRaw}
\end{figure}

\changes{Throughout this section, we will show the output images of our program. To be
able to compare the polarization properties through the proposed software
library toolkit, the same test image of an urban scene is used in all the
experiments. This image, corresponding to the raw image obtained from the
camera, is shown in } \Cref{fig:UrbanSceneRaw}.
\changes{The library is composed of several modules designed to be independent so as to
enable easy maintenance and debugging. With such a structure the addition or removal
of a functionality is straightforward. Regarding the architecture of the code,
it is composed of two general components: the camera server, and the GUI client.
They are detailed in the documentation of the library repository.}

\subsection{\label{sec:CoreComps}Core components}
The first component is a ROS camera server package that interacts directly with
the camera: getting images, and changing (or querying) its parameters such as
the pixel gain, exposure time, and frame rate. It has been designed to support
the integration of existing modern micro-grid polarization cameras. The second
component of the software is the graphical user interface, which works as a
client of the ROS server. The final user interface, with a detailed explanation,
is included in the GitHub repository \cite{github_repo}.
\changes{This interface allows the user to perform all the required tasks involving the camera}
such as changing the image parameters, and the super-pixels filter
configuration. It also allows image processing, raw image display, and sensor
calibration. To analyze the calibration performance, plotting functions are
provided.

\subsection{\label{sec:PolarimetricProcMod}Basic processing}
\changes{The basic polarimetric processing techniques included in the toolkit are
detailed below.}

\begin{figure*}[!t]
    \centering
    \begin{tabular}{cccc}
        \igw{0.24\textwidth}{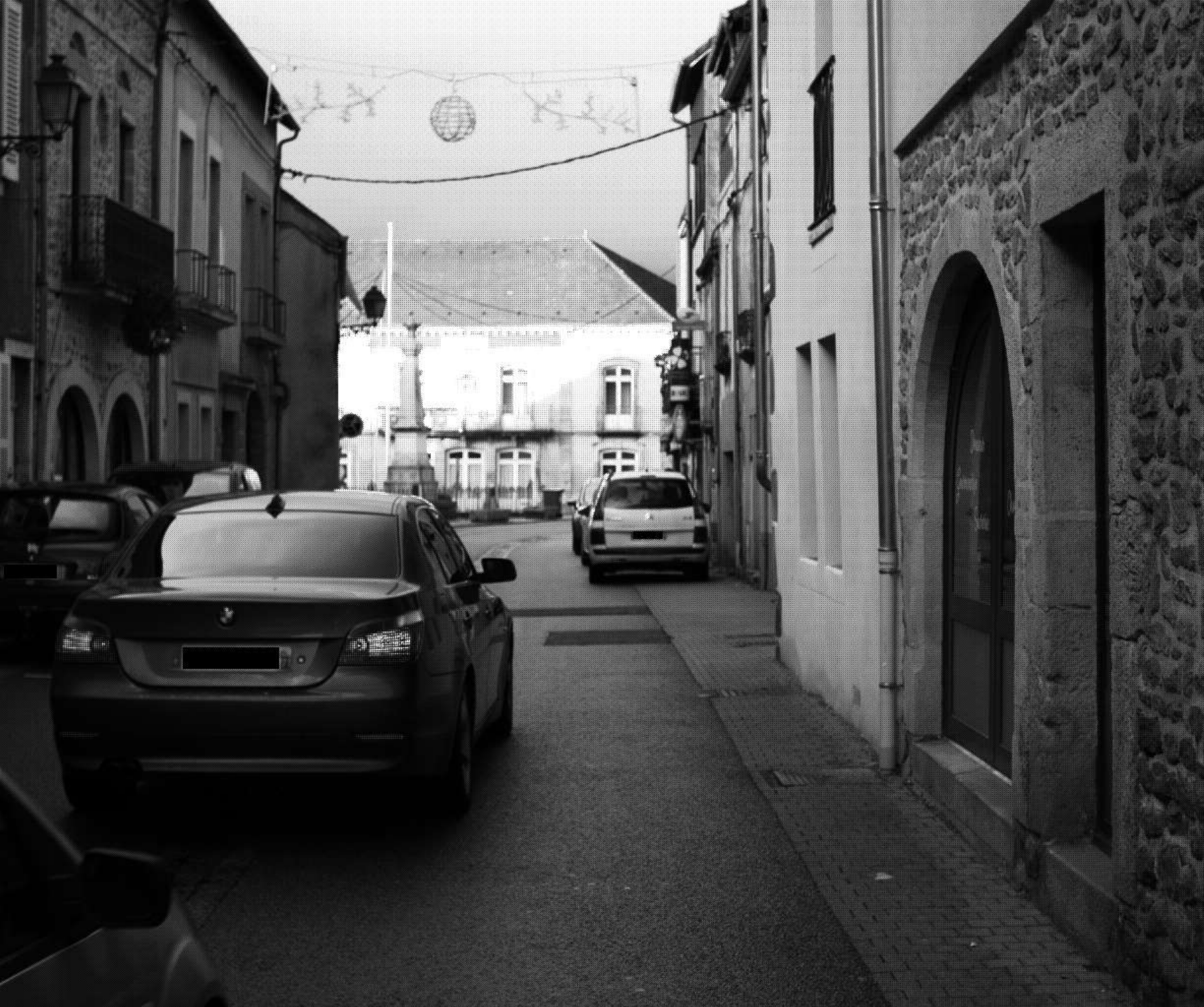} & 
        \igw{0.24\textwidth}{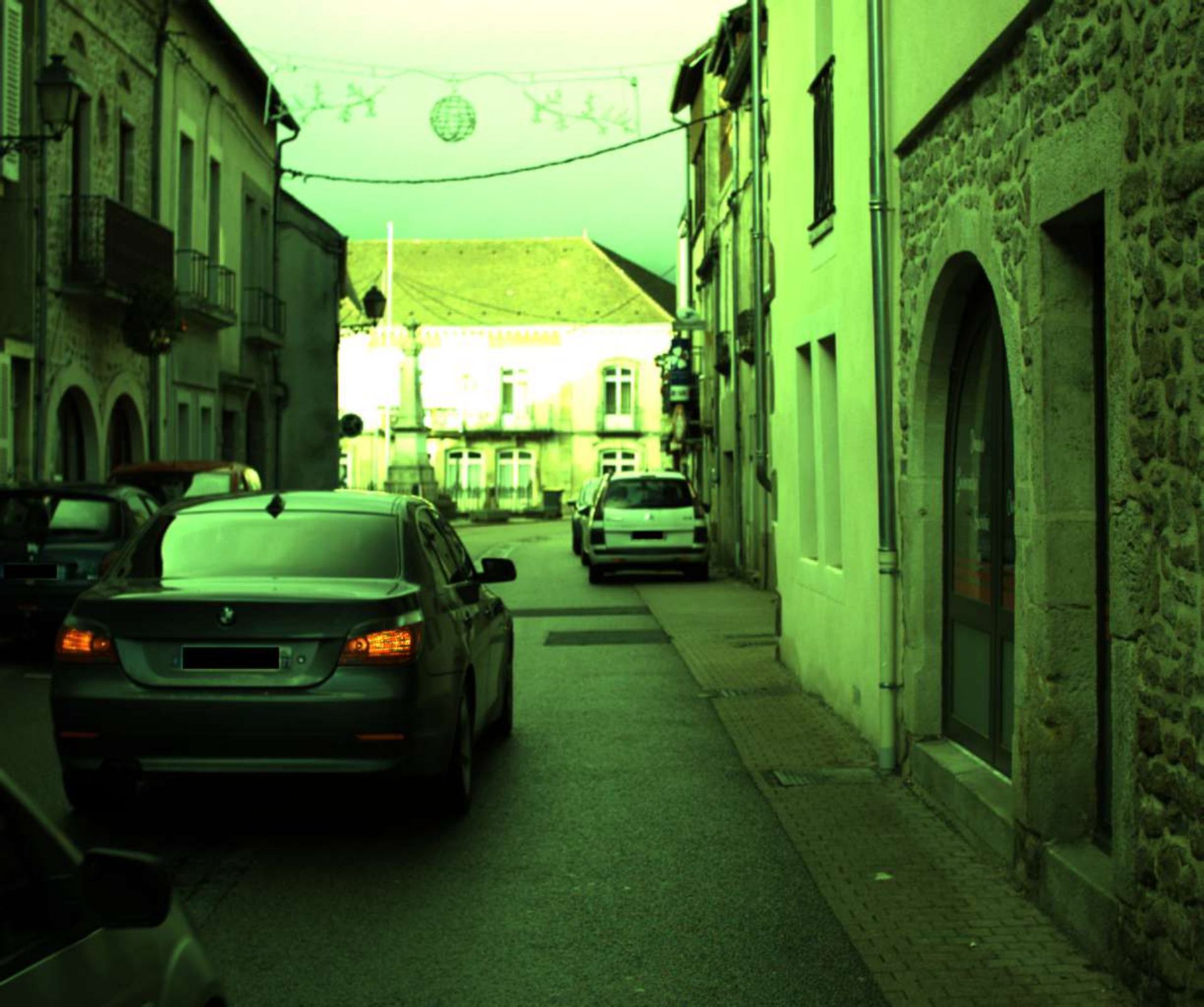} &
        \igw{0.24\textwidth}{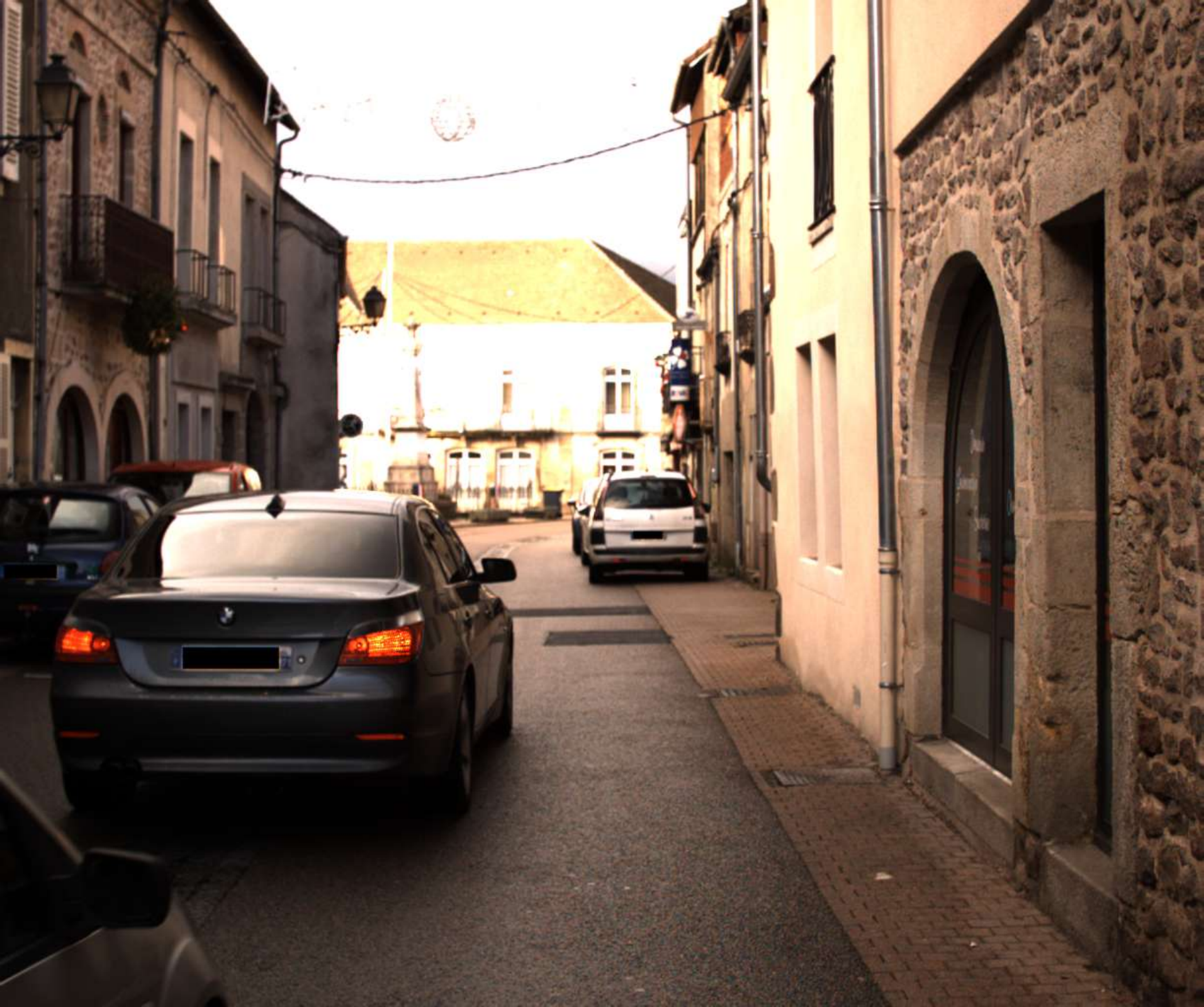} & 
        \igw{0.24\textwidth}{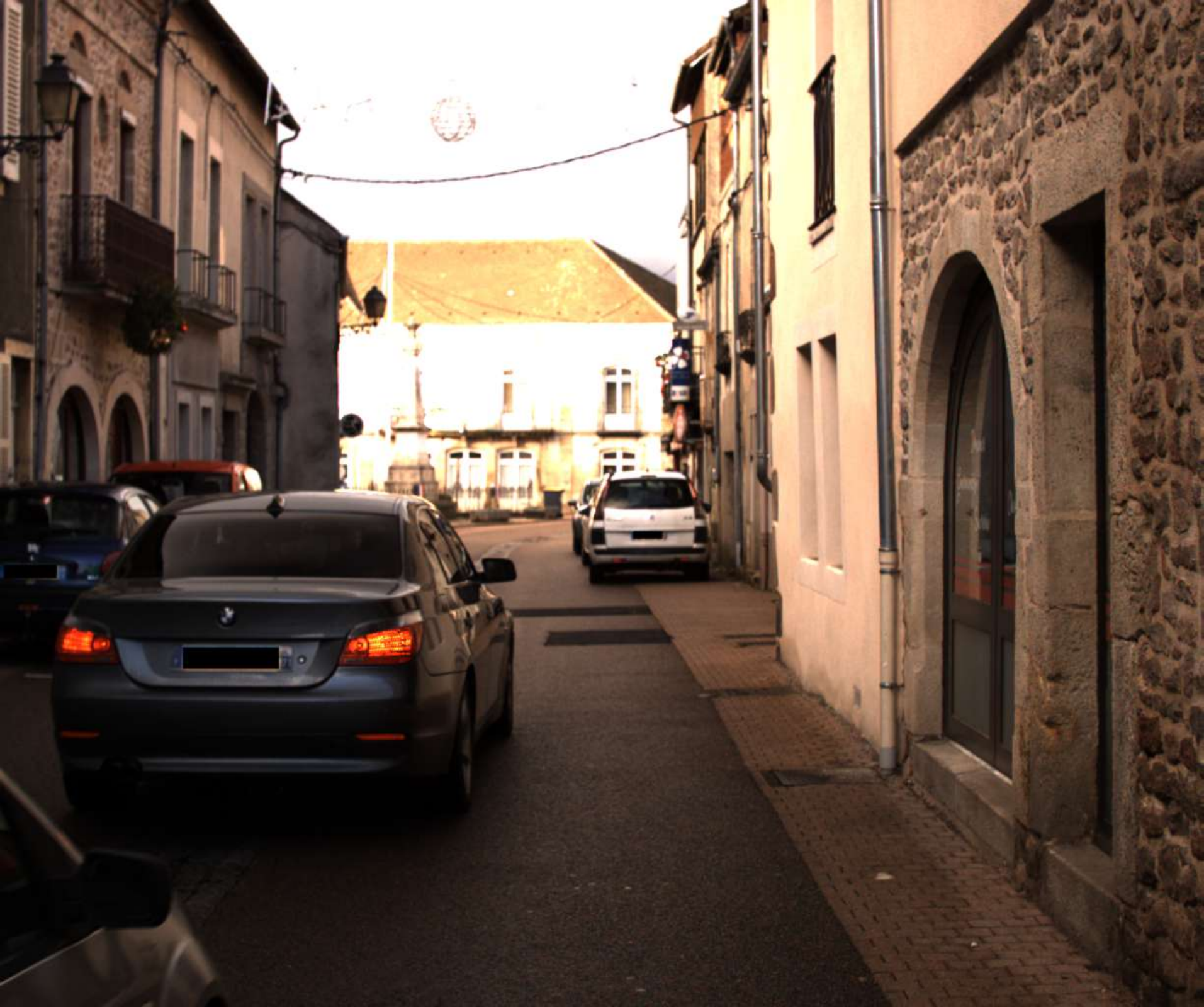} \\
        (a) & (b) & (c) & (d) \\
    \end{tabular}
    \caption[Raw, demosaicked, and white balanced versions of the reference image.]
    {   (a) Raw, mosaiced version of the reference image, for the polarization
            channel $0^\circ$.
        (b) Demosaiced version of the reference image, without white-balance,
            for the $0^\circ$ polarization channel.
        (c) White-balanced, demosaiced images for the polarization channel
            $0^\circ$.
        (d) White-balanced, demosaiced images for the polarization channel
            $135^\circ$.
    }
    \label{fig:RawDemosBalnc}
\end{figure*}

\paragraph{Raw split images}: It produces as output four images, \changes{one per
filter orientation independently of the color. An example of the image for
filter orientation of $0^\circ$ and red channel is shown in}
\Cref{fig:RawDemosBalnc} (a).

\paragraph{\label{sec:PolaColor}Polarized color images}:
It produces as output four images \changes{which are the demosaiced version of the
output from the Raw split images mode obtained using the bilinear interpolation
algorithm. Since the raw measurements of the camera are used, they are not
white-balanced. Thus, the resulting images exhibit a greenish aspect, as shown
in }\Cref{fig:RawDemosBalnc} \changes{(b) for the $0^\circ$ polarization-red color
channel. To this image the white-balance algorithm described in
}\Cref{sec:WhiteBalance} \changes{can be applied to obtain the resulting
white-balanced image shown in }\Cref{fig:RawDemosBalnc} \changes{(c). Polarized color
images allow to identify regions in the scene where light is polarized. To
illustrate this, the demosaiced, white-balanced image for the channel
$135^\circ$ is shown in }\Cref{fig:RawDemosBalnc} \changes{(d). Note that the
windshields of the cars appear dark in this image compared to the one
corresponding to filter orientation of $0^\circ$. This indicates that the light
reflected by the windshields is polarized because polarized light is filtered by
polarimetric filters according to their orientations. As explained in}
\Cref{sec:PolaIntro}, \changes{the intensity of linearly polarized light that passes
through a rotating linear polarizer will exhibit a sine wave curve with respect
to the rotation angle. This sine wave is maximum when the orientation of the
filter is equal to the angle of the linearly polarized light and it decreases as
the filter rotates. The intensity reaches a minimum when the orientation of the
filter is shifted by $\pi / 2$ radians with respect to the angle of linear
polarization of the light. In the case of the RGB-polarization camera, four
filter orientations are considered. If any of these orientations matches the
AoLP of the incoming polarized light, then the corresponding image will have a
bright spot, and the one that is shifted by $\pi / 2$ radians will have a dark
spot at the same position. Since the windshields of the cars in }
\Cref{fig:RawDemosBalnc} \changes{(d), for a filter orientation of $135^\circ$, is
dark where as in }\Cref{fig:RawDemosBalnc} \changes{(c), for a filter orientation of
$0^\circ$, it is rather bright, this means that the $135^\circ$ filter has
filtered out a great part, if not all, of the polarized reflections. Thus, we
can deduce that the reflected light has an AoLP close to $45^\circ$, and that it
is highly linearly polarized.}

\paragraph{Original color}: Only one image is returned. This image is
equivalent to the one obtained with a conventional RGB camera. It corresponds to
the first component of the Stokes vector $S_{0}$. It is computed by applying
\cref{eq:StokesFromI} to the output of the Raw split images mode. \changes{The obtained
image is then demosaiced to obtain the three-channel color image.}

\begin{figure*}[!t]
    \centering
    \begin{tabular}{ccc}
        \igw{0.24\textwidth}{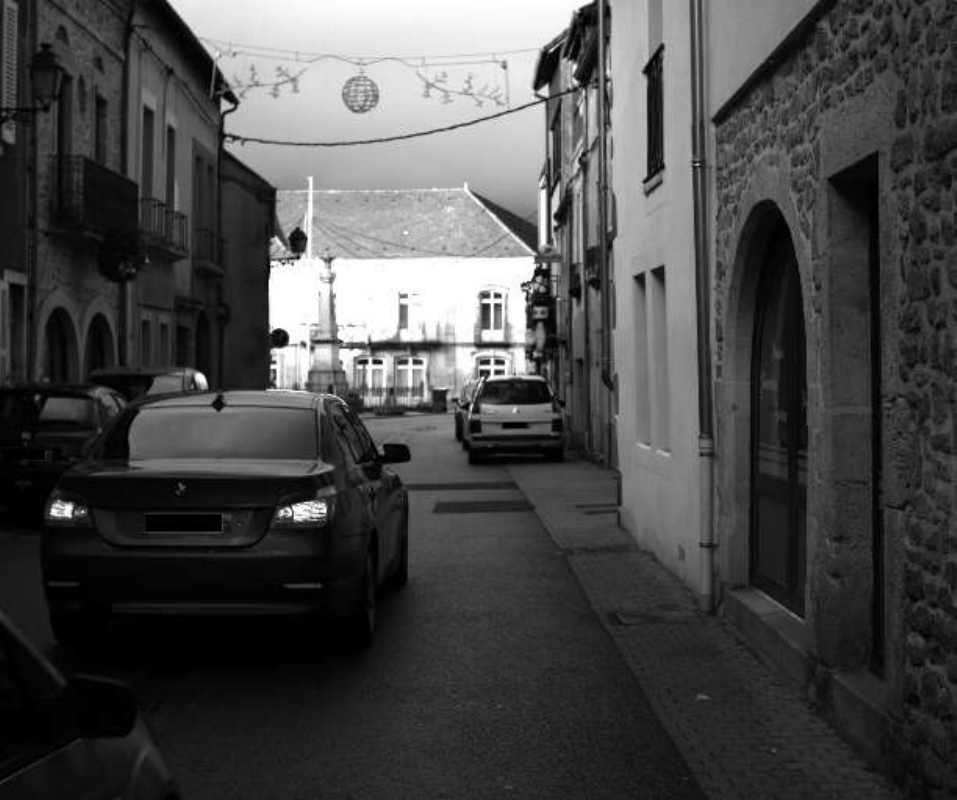} &
        \igw{0.24\textwidth}{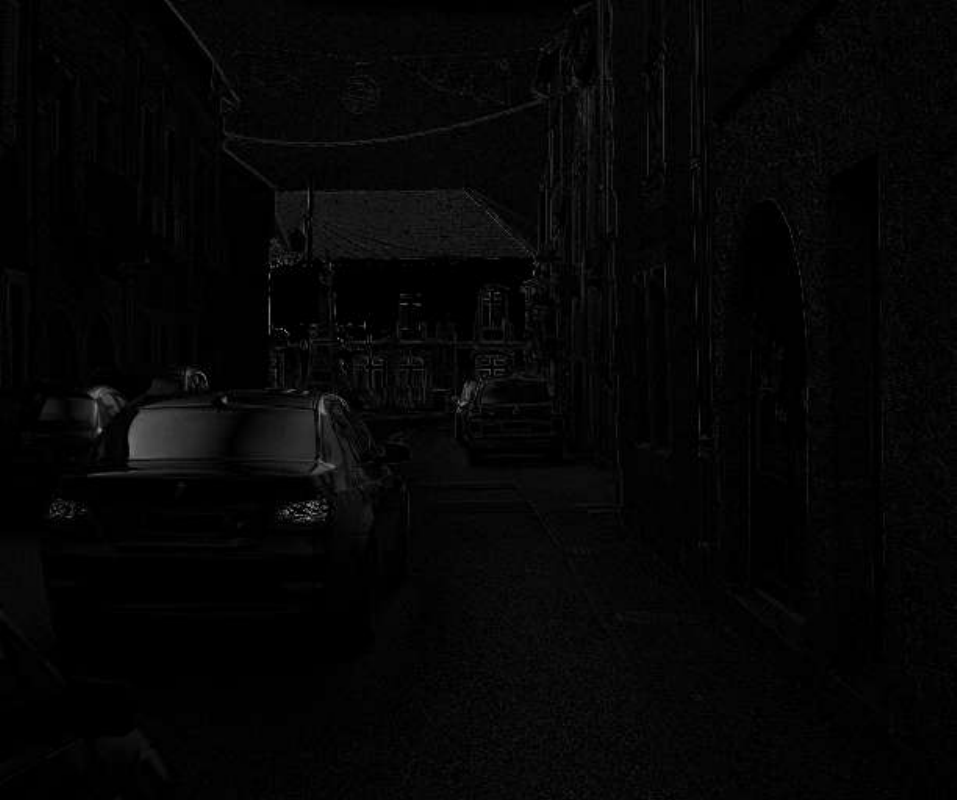} &
        \igw{0.24\textwidth}{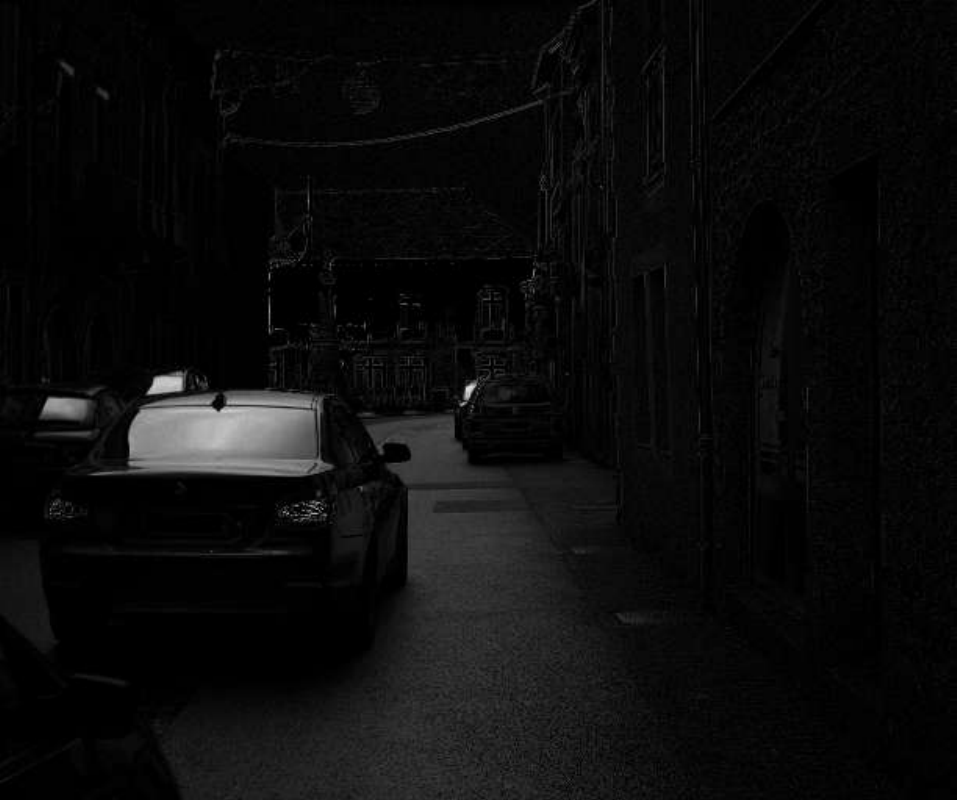} \\
        (a) & (b) & (c) \\
    \end{tabular}
    \begin{tabular}{cccc}
        \igw{0.24\textwidth}{images-software/02_raw_aolp_red.pdf} &
        \igw{0.24\textwidth}{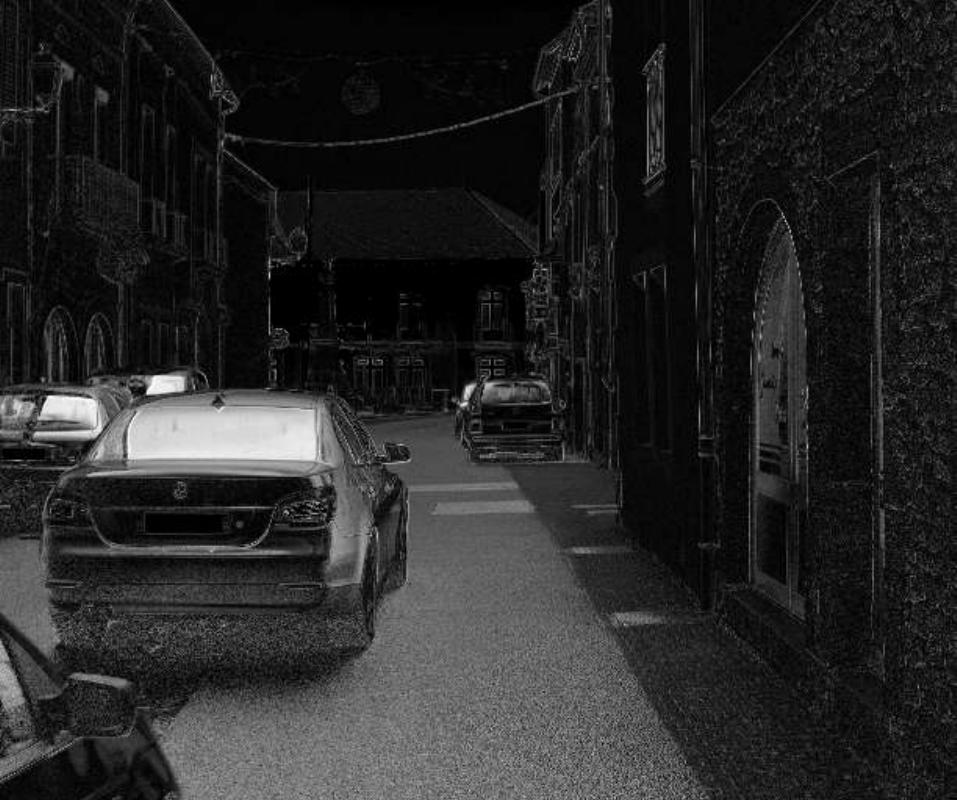} &
        \igw{0.24\textwidth}{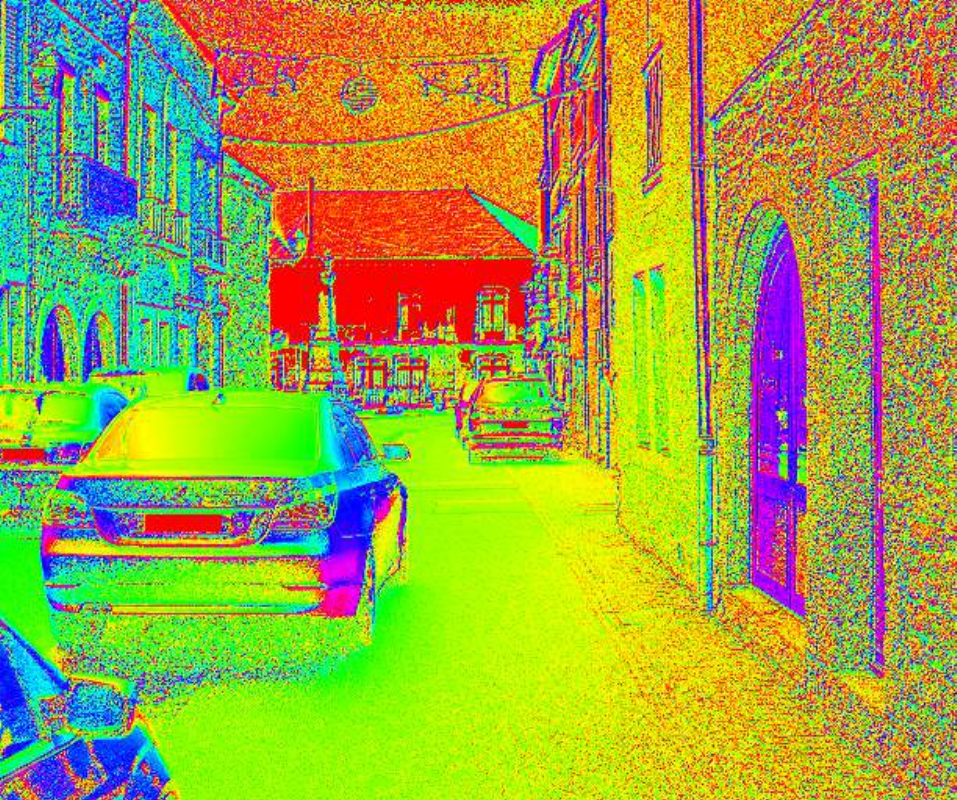} &
        \igw{0.24\textwidth}{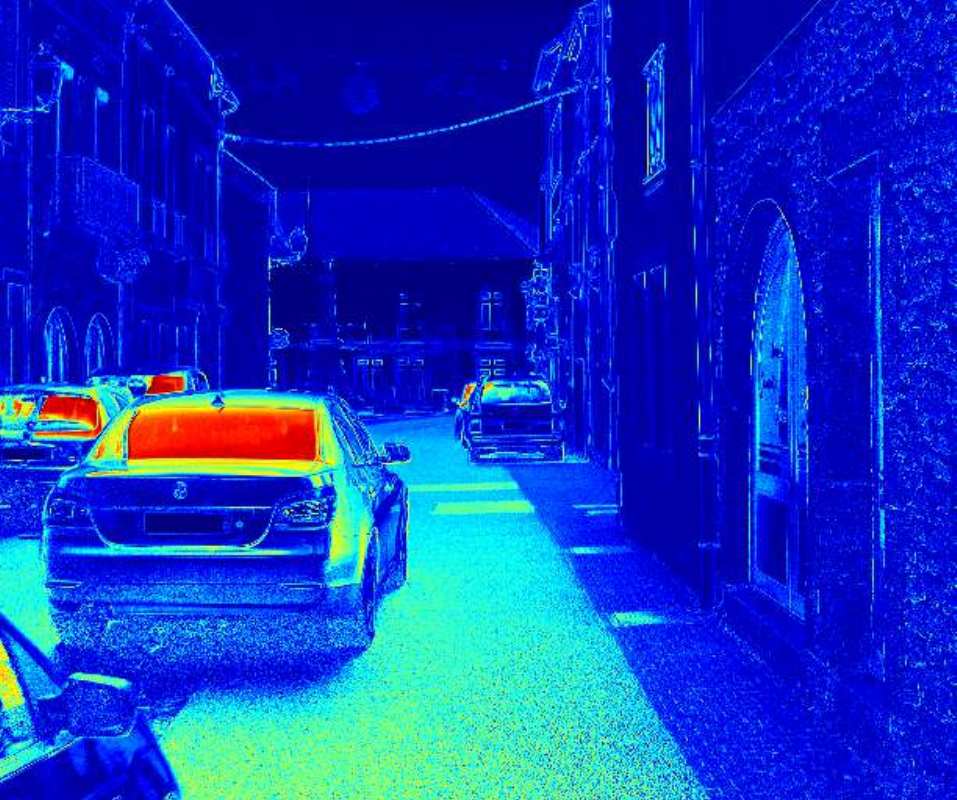} \\
        (d) & (e) & (f) & (g) \\
    \end{tabular}
    \caption[Stokes and polarization parameters of the reference image.]
    {\changes{Polarization images for the red channel. (a) (b) (c) $S_{0,1,2}$
        components of the Stokes vector. The parameter $S_{0}$ corresponds also
        to the total intensity image. (d) (e) Raw AoLP, and Raw DoLP. (f) Raw
        AoLP of the red channel, colored with the HSV palette. (g) Raw DoLP of
        the red channel, colored with the Jet palette.}}
    \label{fig:StokesIRhoPhi}
\end{figure*}

\paragraph{\label{item:StokesComputation}Stokes images}: After separating the
input image by polarization channel, the three Stokes components images can be
obtained by applying \cref{eq:StokesFromI}. Since the polarization state of the
light depends on the frequency of the light, the Stokes vectors are split by
color channel. As a consequence, this function will provide $3\times4=12$
images. Four color channels are considered since the Bayer patterns consist of
$2\times2$ patterns of red, green, green, and blue color filters. \changes{The three
linear components, $S_{0}$, $S_{1}$ and $S_{2}$, of the Stokes vector for the
red channel are shown in }\Cref{fig:StokesIRhoPhi} \changes{(a) to (c). The images
for the other channels are included in the GitHub repository
}~\cite{github_repo}. \changes{The $S_{0}$ image is the red channel image of the original color image,
while the $S_{1}$ and $S_{2}$ images are functions of the orientation of the
polarizer filters. Since $S_{1}$ and $S_{2}$ can have negative values, it is their
absolute values that are considered. In the same way as polarized color images,
Stokes component images can be used to detect regions with polarized light.
Indeed, polarized light will be characterized by a bright or high Stokes value
region in one of the $S_{1}$ and $S_{2}$ images and a corresponding dark or low
Stokes value region in the other image. Conversely, regions that appear dark in
both images correspond to areas where the reflected light is weakly polarized.
In} \Cref{fig:StokesIRhoPhi} \changes{(a) and (c), we can deduce that the windshields
exhibit highly polarized reflections. This is consistent with what we have
already deduced from the polarization color images.}

\paragraph{Raw I - Rho - Phi}: As also explained in \Cref{sec:PolaIntro}, the
Stokes vector can be represented as a function of three physical parameters: the
total intensity $I$, the degree of linear polarization $\rho$, and the angle of
linear polarization $\phi$. The equations to compute these parameters as a
function of the Stokes vector are given in \cref{eq:PolaFromStokes}. \changes{Again,
these three physical parameters can be computed for each color channel, and thus
12 images are also returned in this mode. Since each parameter has a different
interval of values, all of them have been normalized to be in the range}
$\left[0,255\right]$. \changes{The resulting images for the red channel only are
shown in} \Cref{fig:StokesIRhoPhi} \changes{(a), (d), and (e), respectively. These
images are a good representation of all the objects that reflect linearly
polarized light. Note that a pixel value in the AoLP image has a meaning only if
its corresponding pixel in the DoLP image has a non-zero value. In this set of
images, it is possible to confirm that the road, the windshields, part of the
body of the car, and some door glasses have a large DoLP. Thus, the AoLP and
DoLP features are extremely valuable and can be used, for example, in Deep
Learning models to improve the accuracy of the network results for these
objects.}

\paragraph{I - Rho - Phi}: In the Raw I - Rho - Phi mode, \changes{all the images are
single channeled, thus they are displayed in gray-scale}. Particularly for the
angle of linear polarization, this is not an adequate representation, since it
is a circular variable. A proper representation would be one that assigns the
same color to the maximum and minimum values. In this mode, this is done by
creating a color palette based on the HSV color space. Let $X$ be a gray-scale
value in the range $\left[0,255\right]$. The HSV palette is defined as a
function that assigns a 3D vector $\mathbf{C_{hsv}}$ to each value $X$, such
that:
\begin{equation}
    \mathbf{C_{hsv}}=\left[\dfrac{179X}{255},255,255\right]^T
\end{equation}
Then, the obtained three-channeled image is converted from the HSV to the RGB
color space. On the other hand, the degree of linear polarization is colorized
using the Jet palette, which maps blue colors to low values, green colors to
middle values, and red colors to high values. As in the Raw I - Rho - Phi mode,
12 images are returned in this mode. \changes{The results of this mode for the
colored AoLP and DoLP are shown in } \Cref{fig:StokesIRhoPhi} \changes{(f) and (g)
respectively, for the red channel.}

\paragraph{Fake colors}: When dealing with polarization imaging, a
correspondence is established between the HSV (hue, saturation, and value) color
space and the intensity, degree of linear polarization, and angle of linear
polarization\cite{WolffFirstPolaPaper}. Since the AoLP is a circular variable,
it is considered to be the hue of the color. The saturation is the purity of
that color, meaning that a saturation of 100\% is a pure color, and a saturation
of 0\% means that it is a gray-level value. Similarly, the degree of linear
polarization indicates the "purity" of the light: if it is totally unpolarized,
$\rho$ is equal to zero, and if it is totally linearly polarized, $\rho$ is
equal to 1. Finally, if a conventional color image is considered, and the value
channel is extracted from it in the HSV space, a gray-scale version of the
original image is obtained. This information is similar to the total intensity
measured by the $I$ parameter. Thus, a color image can be obtained if the $I$,
$\phi$, and $\rho$ images are stacked together, considered to be in the HSV
space, and then converted back to the RGB space. The result obtained in this way
is called a fake color image. The colors obtained with this processing algorithm
have the following properties:
\begin{itemize}
    \item Unpolarized light is represented by gray-scale colors.
    \item Highly polarized light will be colored.
    \item The colors in the image depend mainly on the AoLP, thus,
        to the surface orientation.
\end{itemize}

\begin{figure}[!t]
    \centering
    \begin{tabular}{cc}
        \igw{0.24\textwidth}{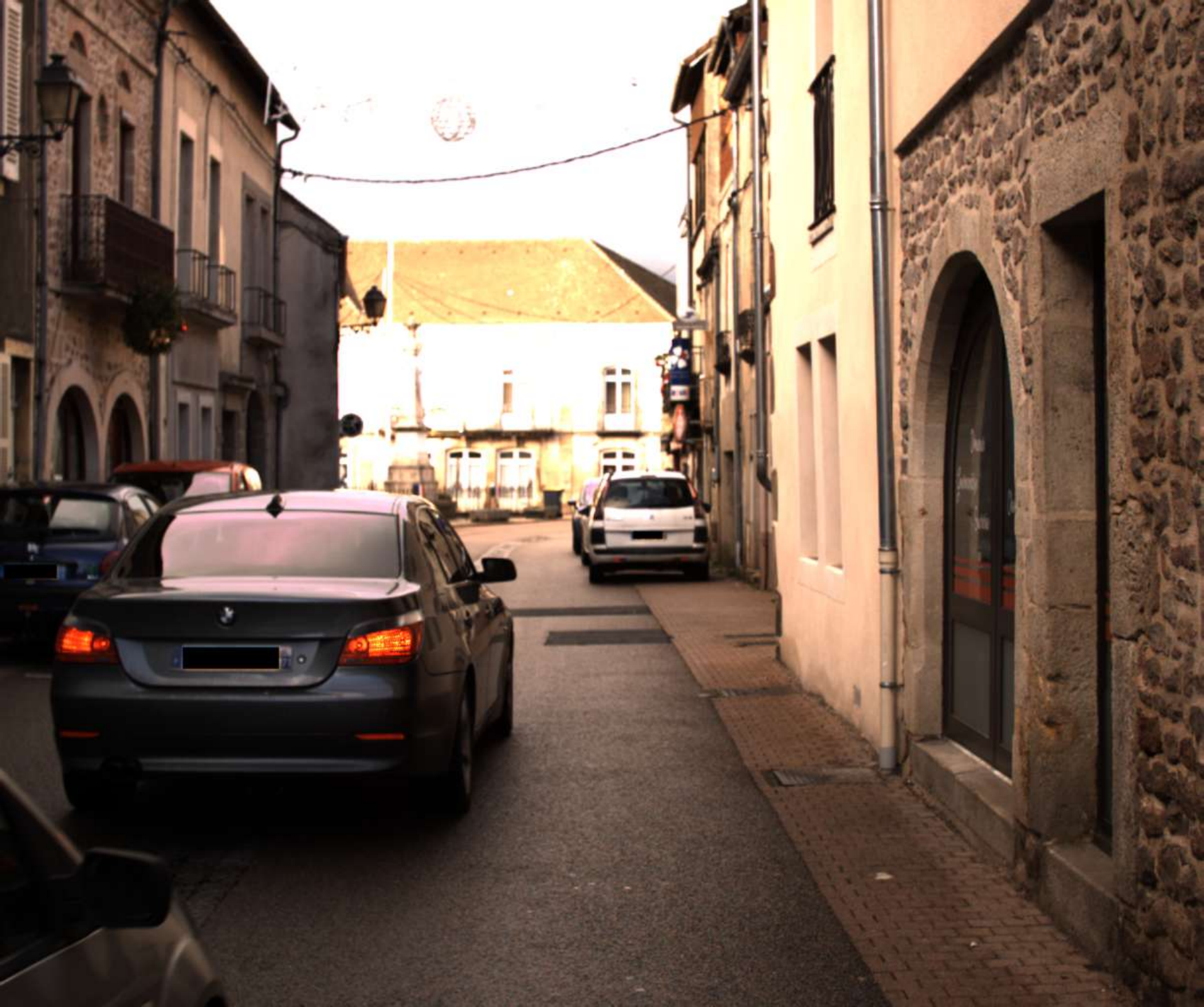} & \igw{0.24\textwidth}{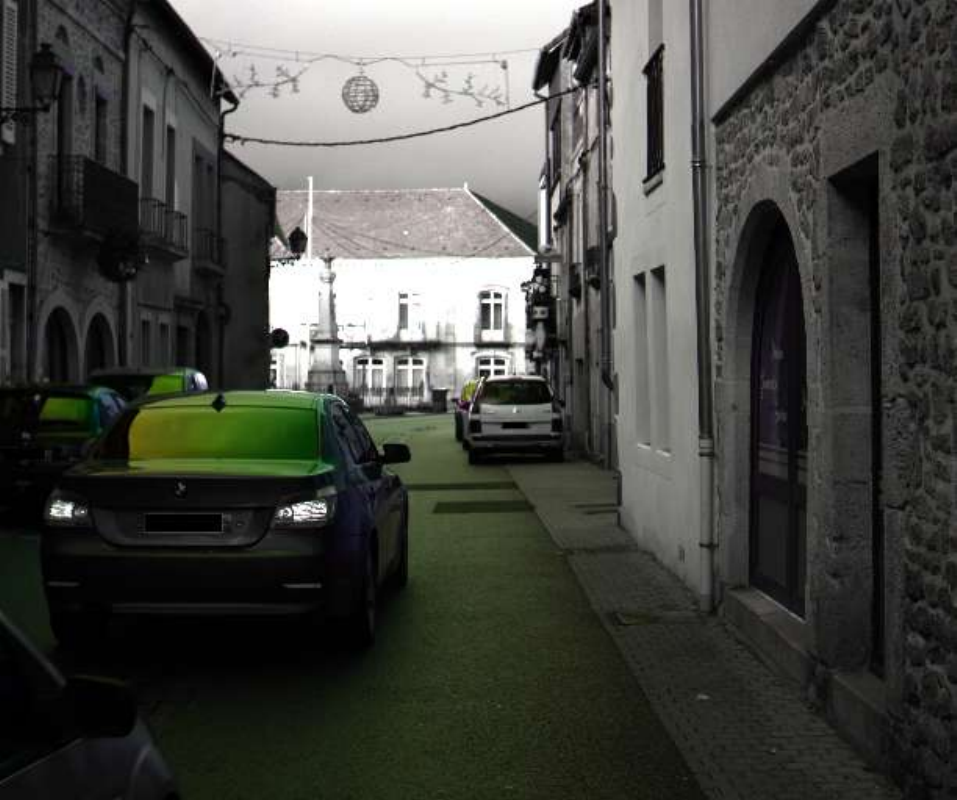} \\
        (a) & (b) \\
    \end{tabular}
    \caption[Color and fake colors version of the reference image.]
    {(a) White-balanced, color image. (b) Fake colors image representation for the red channel.}
    \label{fig:FakeColors}
\end{figure}

This processing is interesting since it helps to quickly identify the
objects that reflect polarized light. As explained before, the
polarization parameters depend on the frequency, thus the fake color images
are separated by color channel. Therefore, this operation returns four
images. \changes{The resulting image for the red channel is shown in}
\Cref{fig:FakeColors}.

\subsubsection{\label{sec:WhiteBalance}White-balance module}
The observed light by a camera depends on two parameters: the observed object
reflectance, and the illumination of the color
\cite{CVPR_2022_dolp_color_const}. Thus, a white-balance algorithm needs to be
used in order to restore the true colors in the scene. In some polarization
cameras, and particularly in the Basler RGB-polarization camera, this type of
algorithm is not correctly implemented. Thus, our software includes an
implementation of a white-balance algorithm. The white balance is a gain applied
to each color channel to get the true color \changes{of the objects despite the
environment where the image is taken}. In our implementation, the automatic
white search is done globally, and not in a user-defined region of interest
(ROI). The algorithm computes the average of all the color channels of a single
orientation, and it searches for the pixels whose average is the highest. If
there is a white piece in the scene, even with the color gains unbalanced, its
average will be the highest. Thus, the pixel whose average is the
highest is considered as white. Then, the highest channel value is left
untouched (gain equal to 1), and the other channel gains are computed such that
their values equal the highest channel value. This algorithm of automatic white
balance is constrained to work with no high-level saturated images. It can be
deactivated and the different gains set manually.
\changes{An example of the input image to this algorithm, and its corresponding output are shown in}
\Cref{fig:RawDemosBalnc} \changes{(b) and (c), respectively}.

\subsection{Polarimetric camera calibration module}
This module is an implementation of the polarimetric camera calibration
algorithm described in \cite{our_calib_paper}. The calibration algorithm is used
to compute a series of matrices that are applied by a correction function to
rectify the measurement errors due to manufacturing imperfections. In this way,
two super-pixels of the same color channel that receive the same light source
will provide the same output measurements. The module also provides a function
to correct the image, once the calibration algorithm has been run. This function
takes a raw image from the camera, and if the calibration matrices have been
computed, the function will return another image, with the same structure as the
input, but with all the pixel measurements corrected by the calibration
algorithm.

\begin{figure*}
    \centering
    \begin{tabular}{cccc}
        \includegraphics[width=0.24\textwidth]{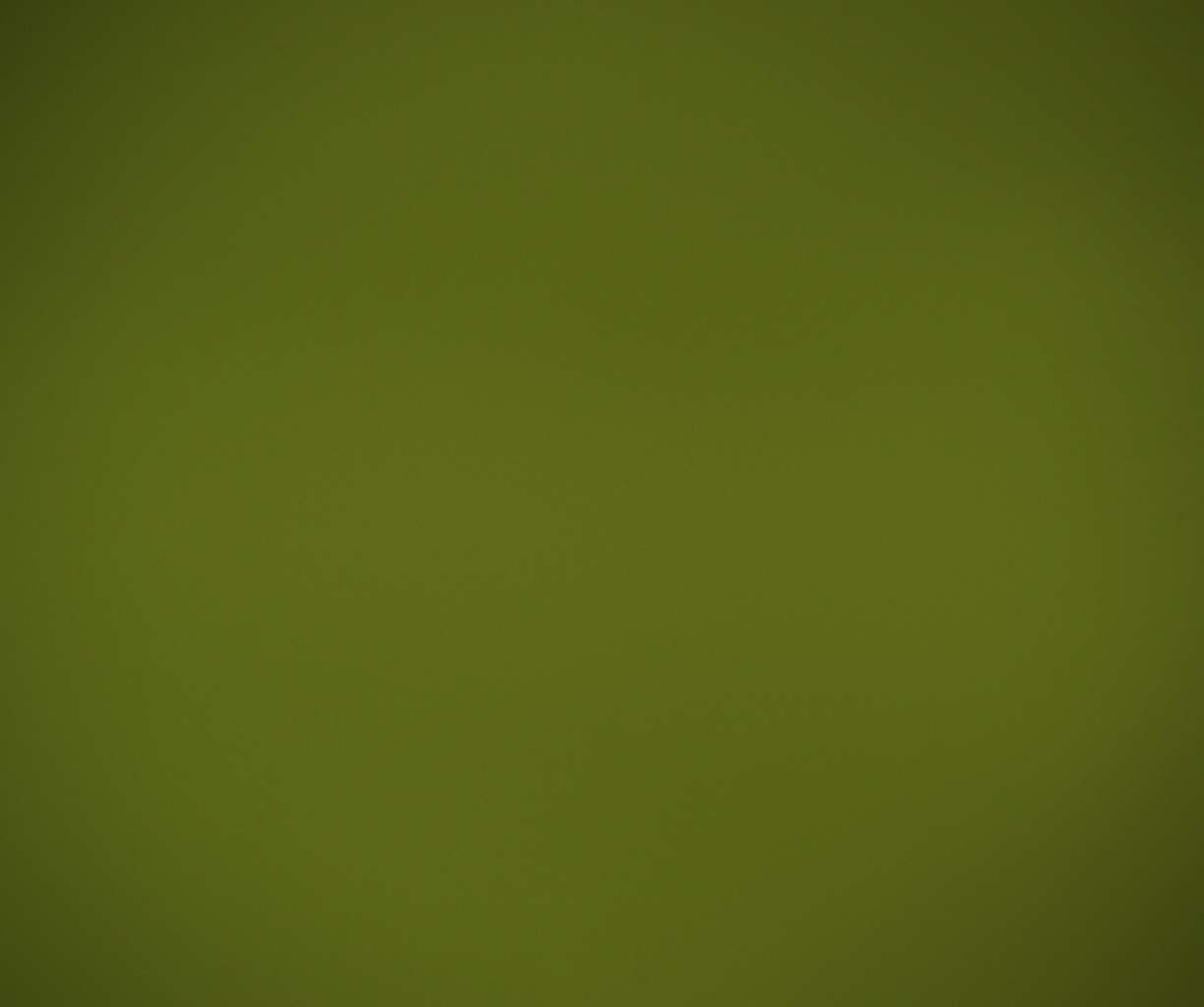} &
        \includegraphics[width=0.24\textwidth]{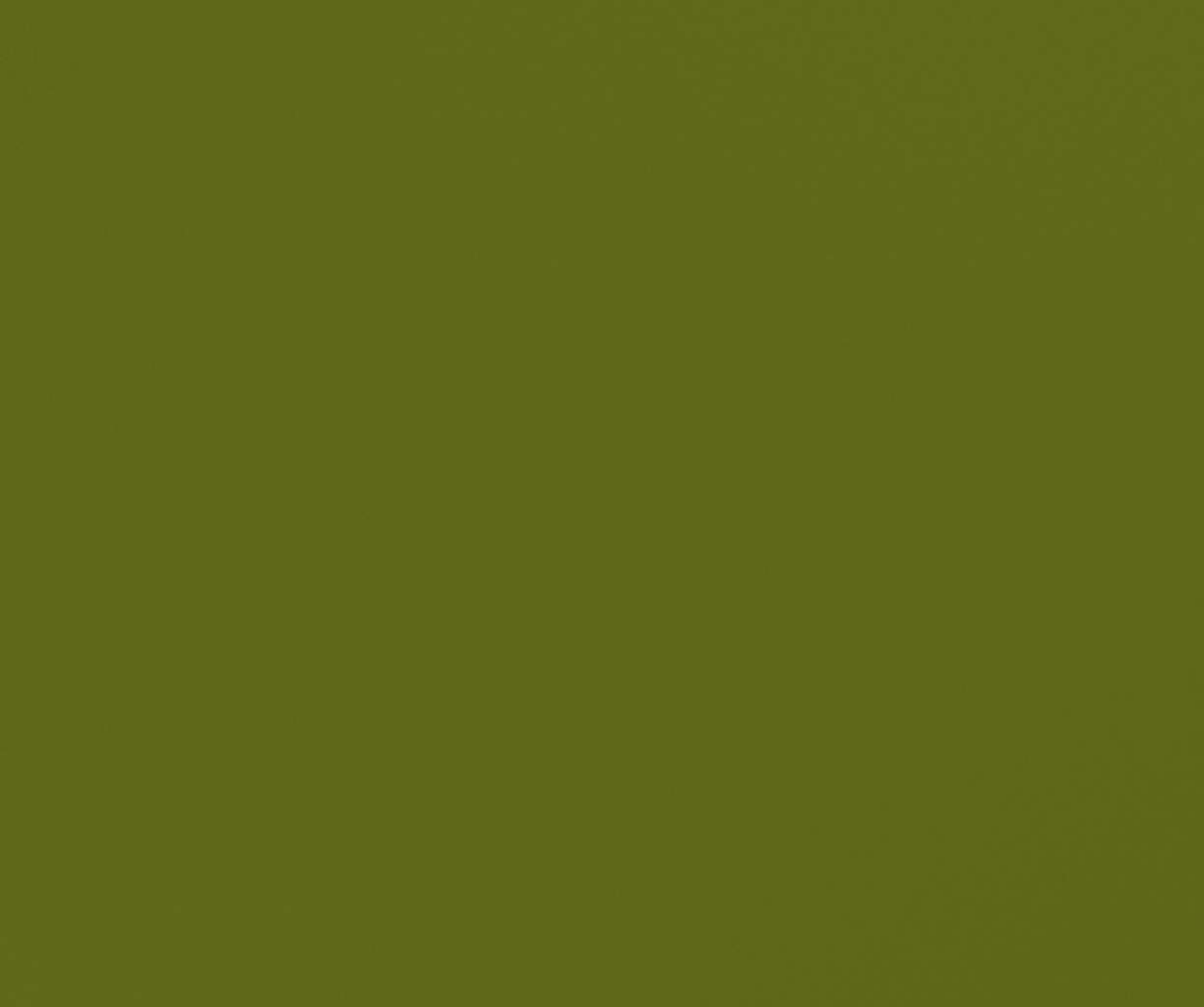} &
        \includegraphics[width=0.24\textwidth]{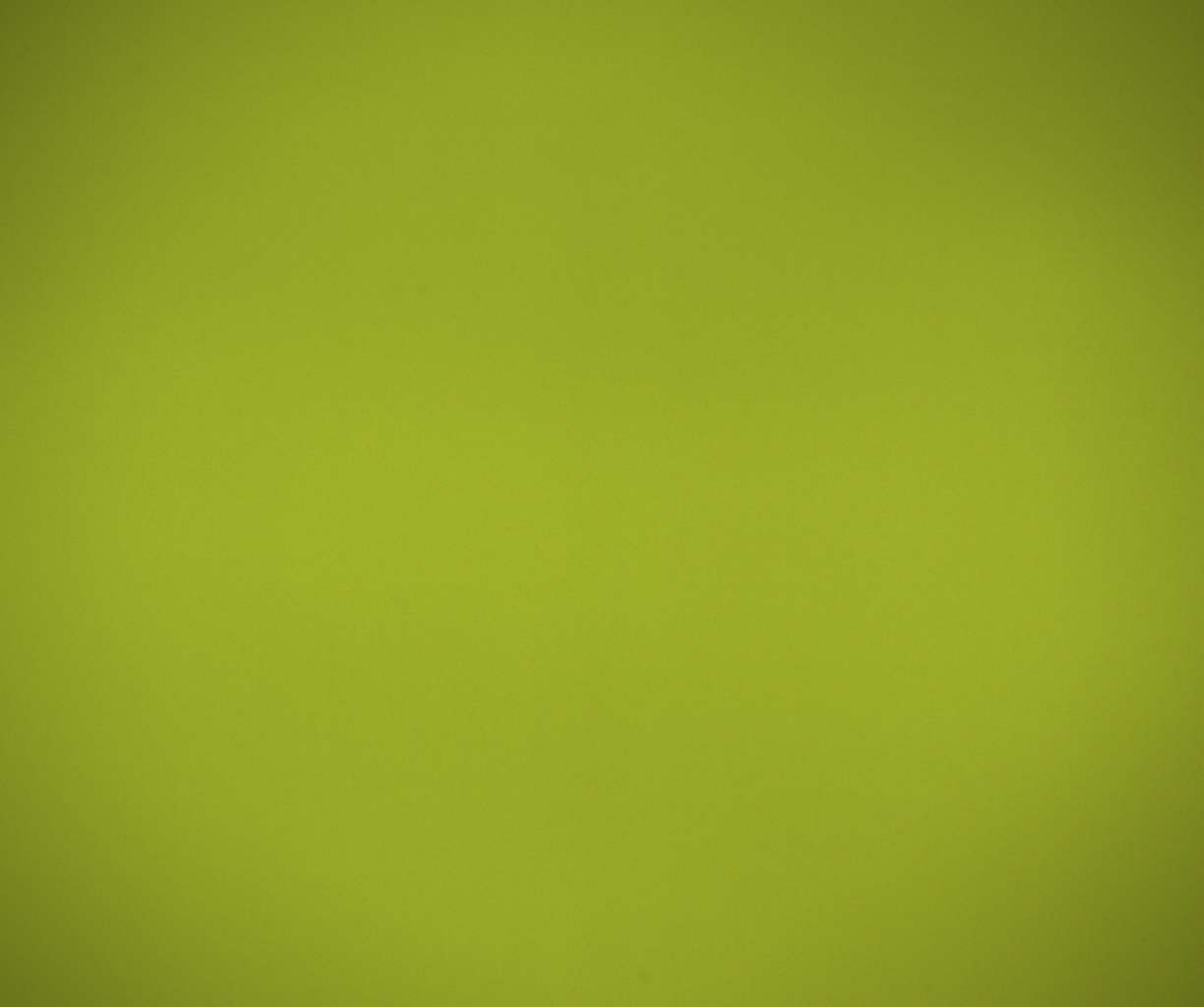} &
        \includegraphics[width=0.24\textwidth]{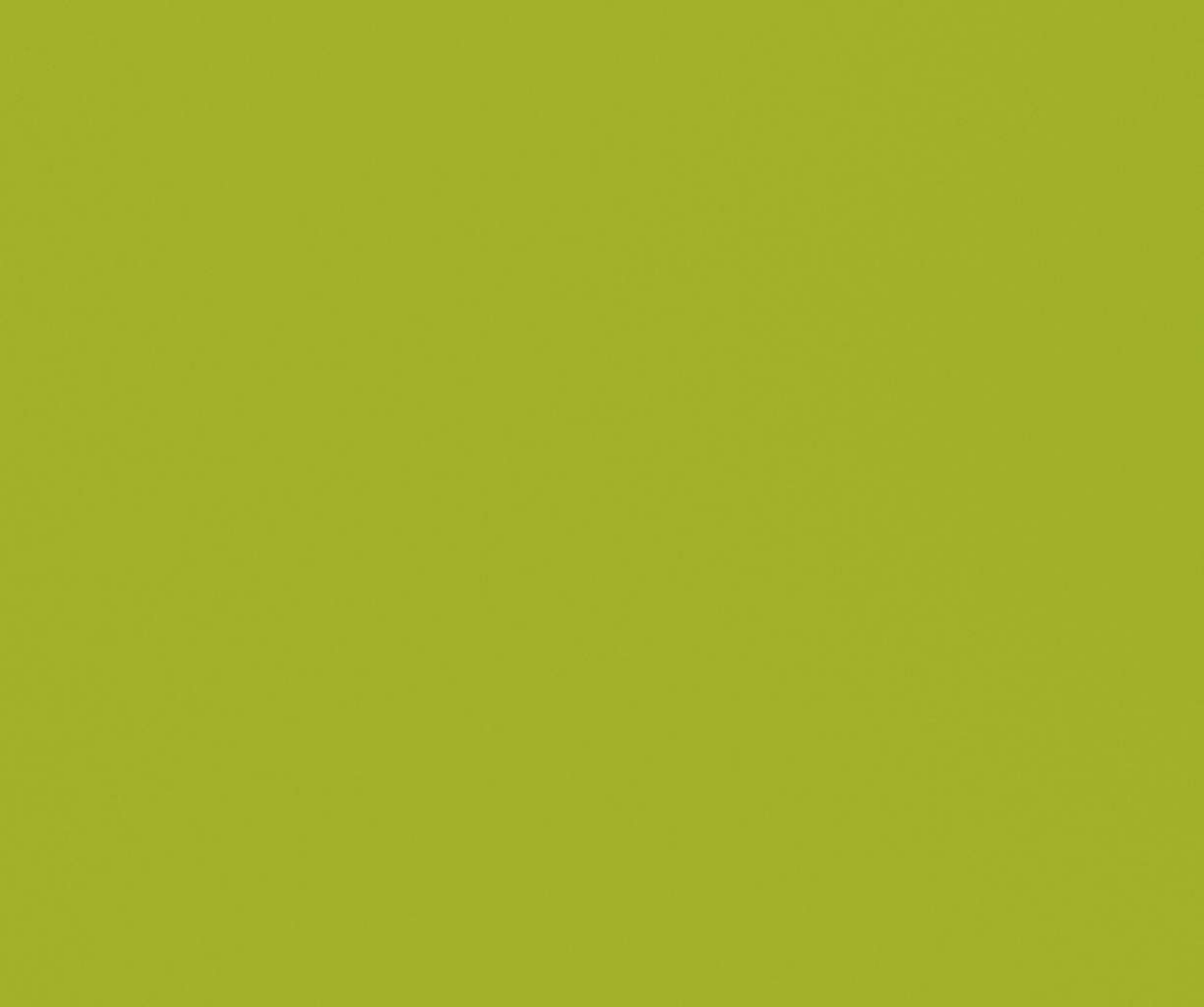} \\
        (a) & (b) & (c) & (d) \\
    \end{tabular}
    \caption[Calibration effects over the calibration light.]
    {\changes{Calibration effects over a uniform, linearly polarized light, with an
    AoLP of $40^\circ$. These images are the demosaiced, RGB polarization
    channels. (a) and (c) Uncalibrated polarization channels $0^\circ$ and $45^\circ$, respectively.
    (b) and (d) Calibrated polarization channels $0^\circ$ and $45^\circ$, respectively.}}
    \label{fig:LightCalibEffects}
\end{figure*}

\changes{The calibration problem can be solved by taking several images of a
uniform and linearly polarized light. If the light source is uniform but
unpolarized, it can be polarized using a linear polarization filter. By turning
the filter, the light received by the camera at each filter position will have a
different AoLP. A sample of a polarized light source with an AoLP of $40^\circ$
is shown in }\Cref{fig:LightCalibEffects}, \changes{for the $0^\circ$, and $45^\circ$
polarization channels. These images correspond to the results before and after
applying the calibration algorithm.}

\changes{The calibration procedure will compute the pixel parameters given the model
defined in }\cite{our_calib_paper}, \changes{i.e., $\left(T_{i}, P_{i},
\theta_{i}\right)$. In this model: ${T_{i}}$ is the pixel gain; ${P_{i}}$ is a
parameter that accounts for the non-ideality of the micro-polarization filter
implemented on the pixel; $\theta_{i}$ is the effective orientation of the
micro-polarization filter of the pixel; $i$ is the position of the pixel
considered. From Mueller calculus, these parameters have an ideal value of
$\left(T_{i}, P_{i}\right)=\left(0.5, 1.0\right)$ for all $i$, and
$\theta_{i}\in\left\{0^\circ, 45^\circ, 90^\circ, 135^\circ\right\}$. However,
in general, each pixel will have a set of parameters that will be different from
these ideal values. Thus, the calibration algorithm allows to compensate for
these differences.}

\changes{To assess the acquisition quality, the different plot functions of the
software can be used. With these plots, the user can determine if the
acquisition is correct, and if the camera measurements are valid. It can also
assess the quality of the sensor and confirm the effectiveness of the correction
on the camera measurements.}

\begin{figure*}
    \centering
    \begin{tabular}{cccc}
    \igw{0.24\textwidth}{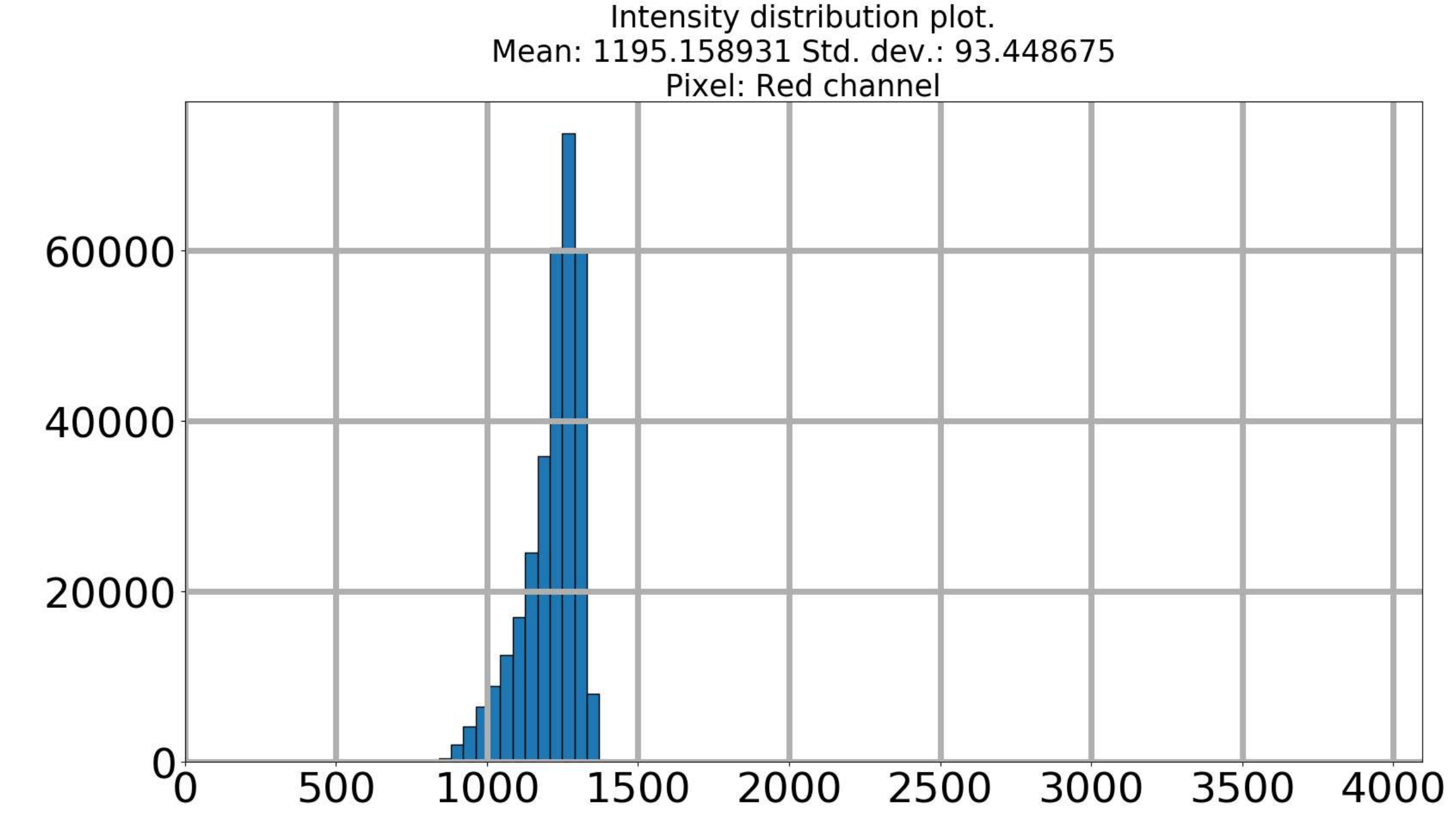} &
    \igw{0.24\textwidth}{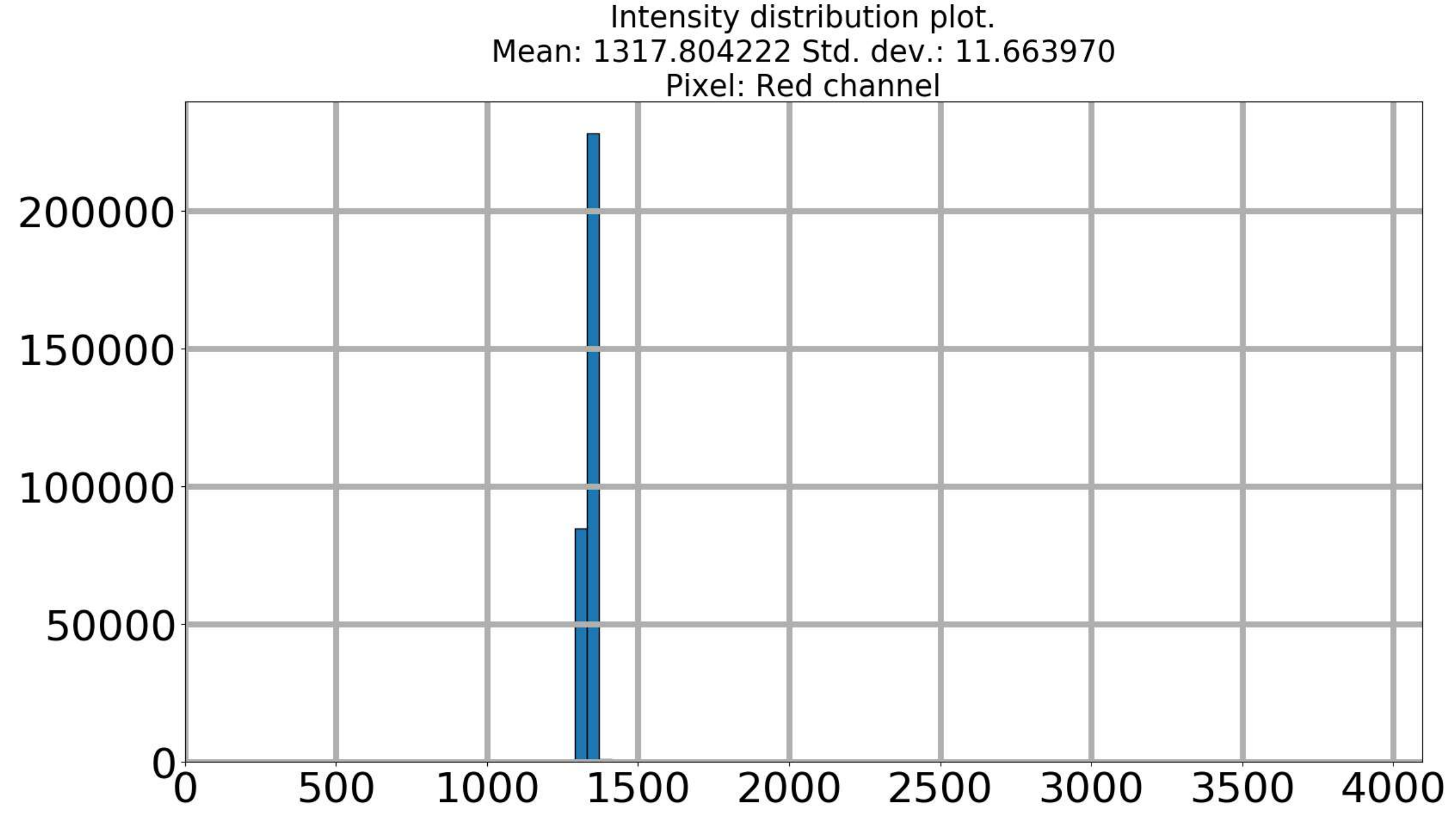} &
    \igw{0.24\textwidth}{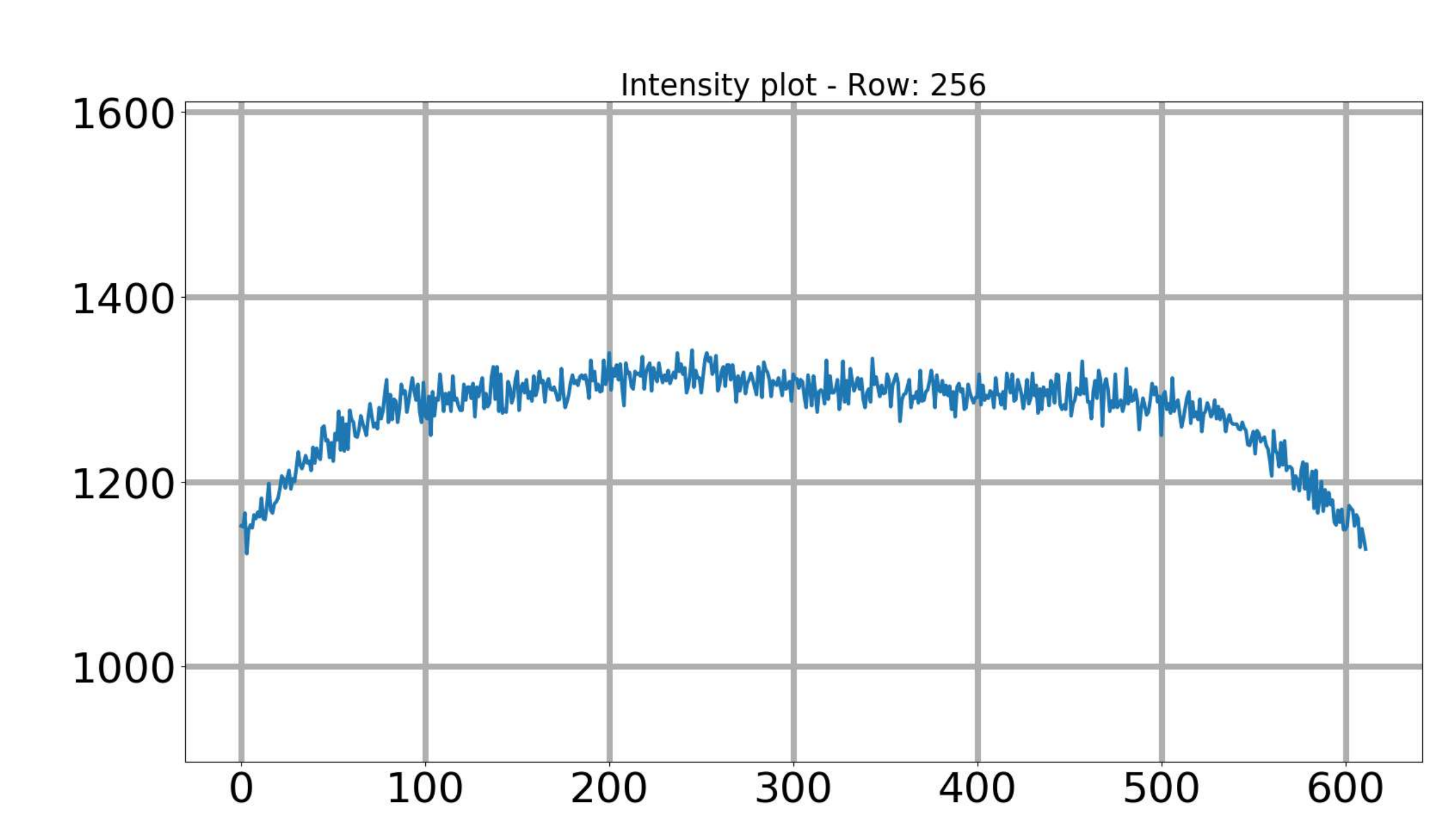} &
    \igw{0.24\textwidth}{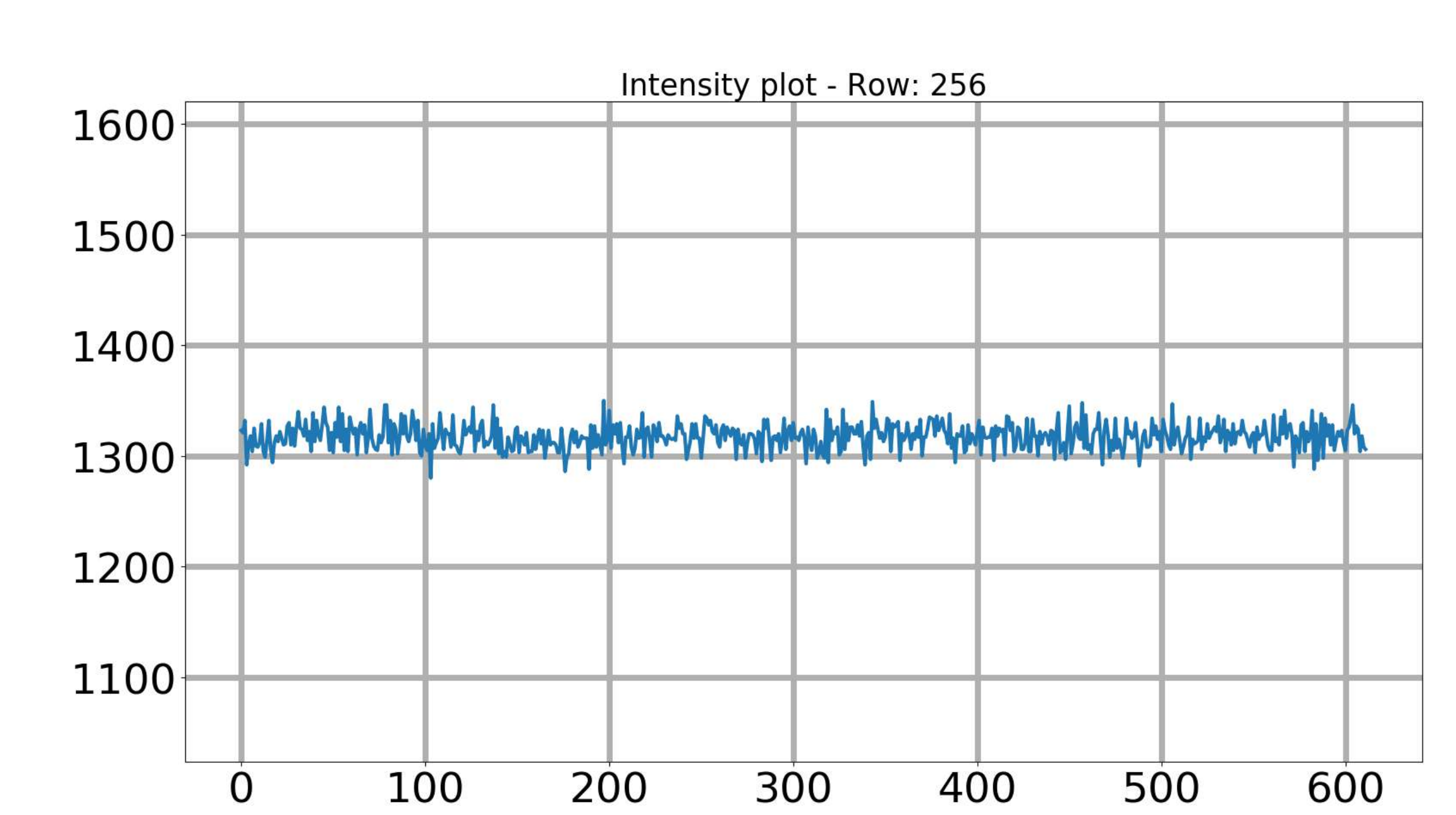} \\
    (a) & (d) & (g) & (j) \\

    \igw{0.24\textwidth}{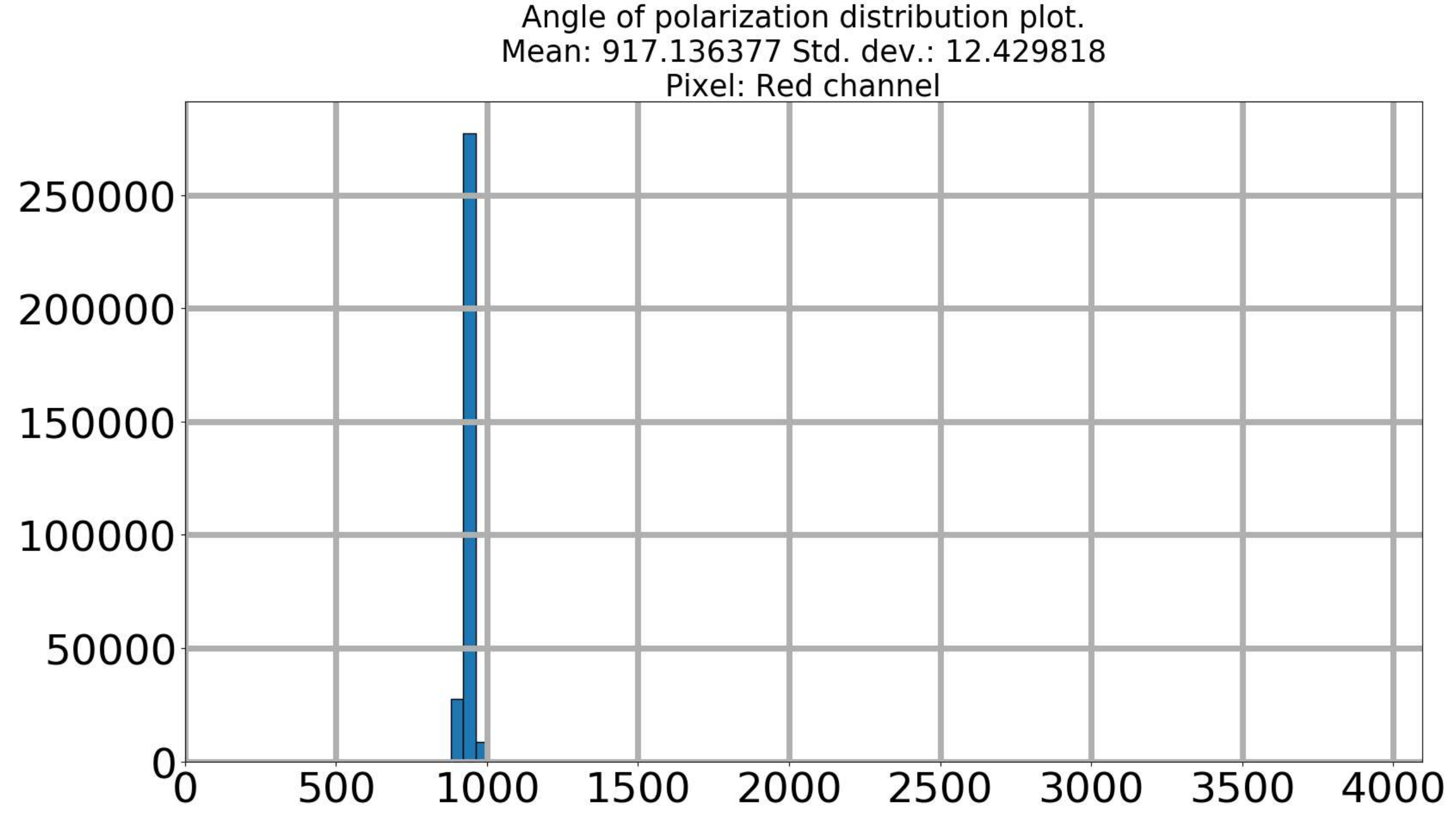} &
    \igw{0.24\textwidth}{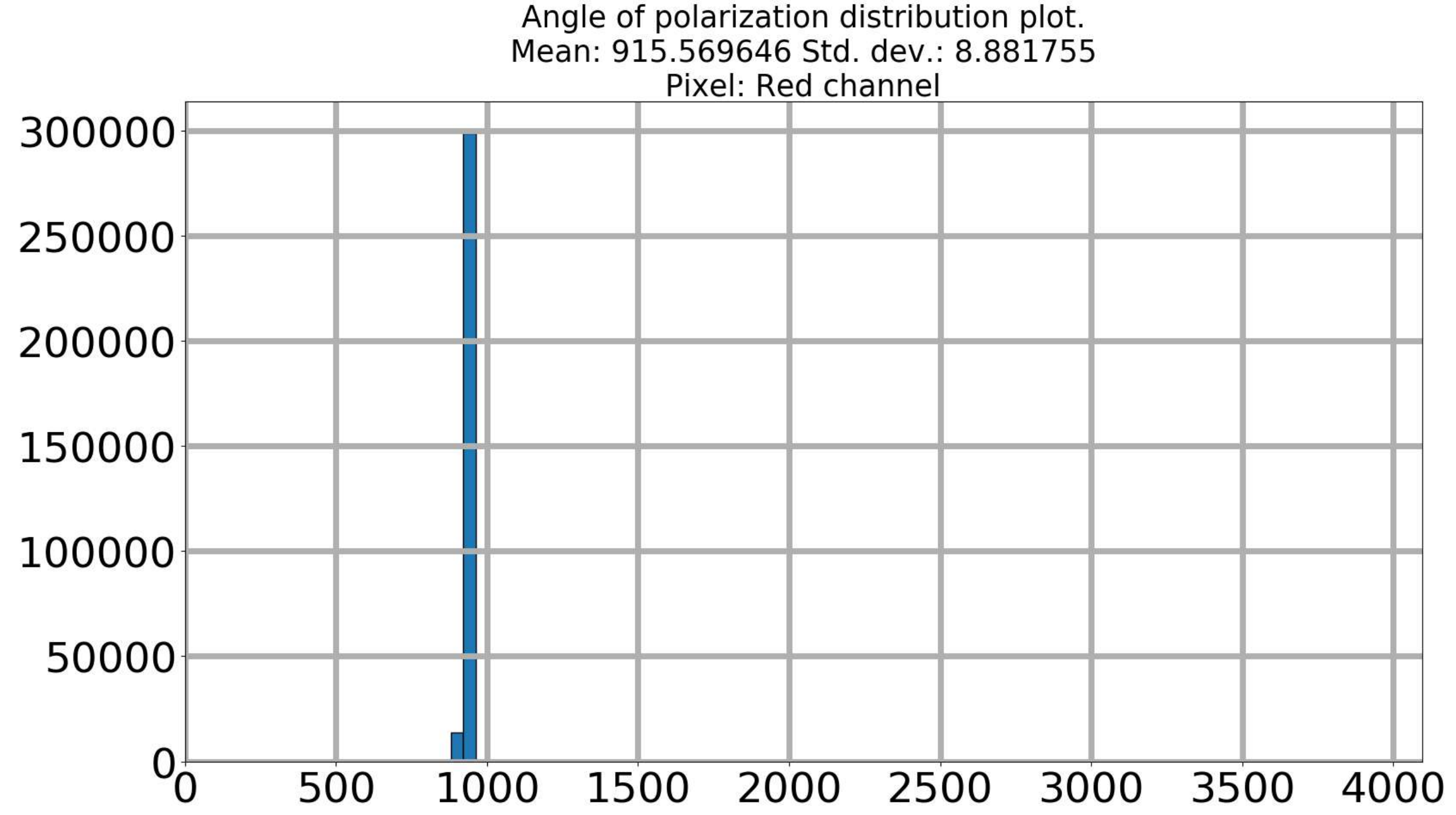} &
    \igw{0.24\textwidth}{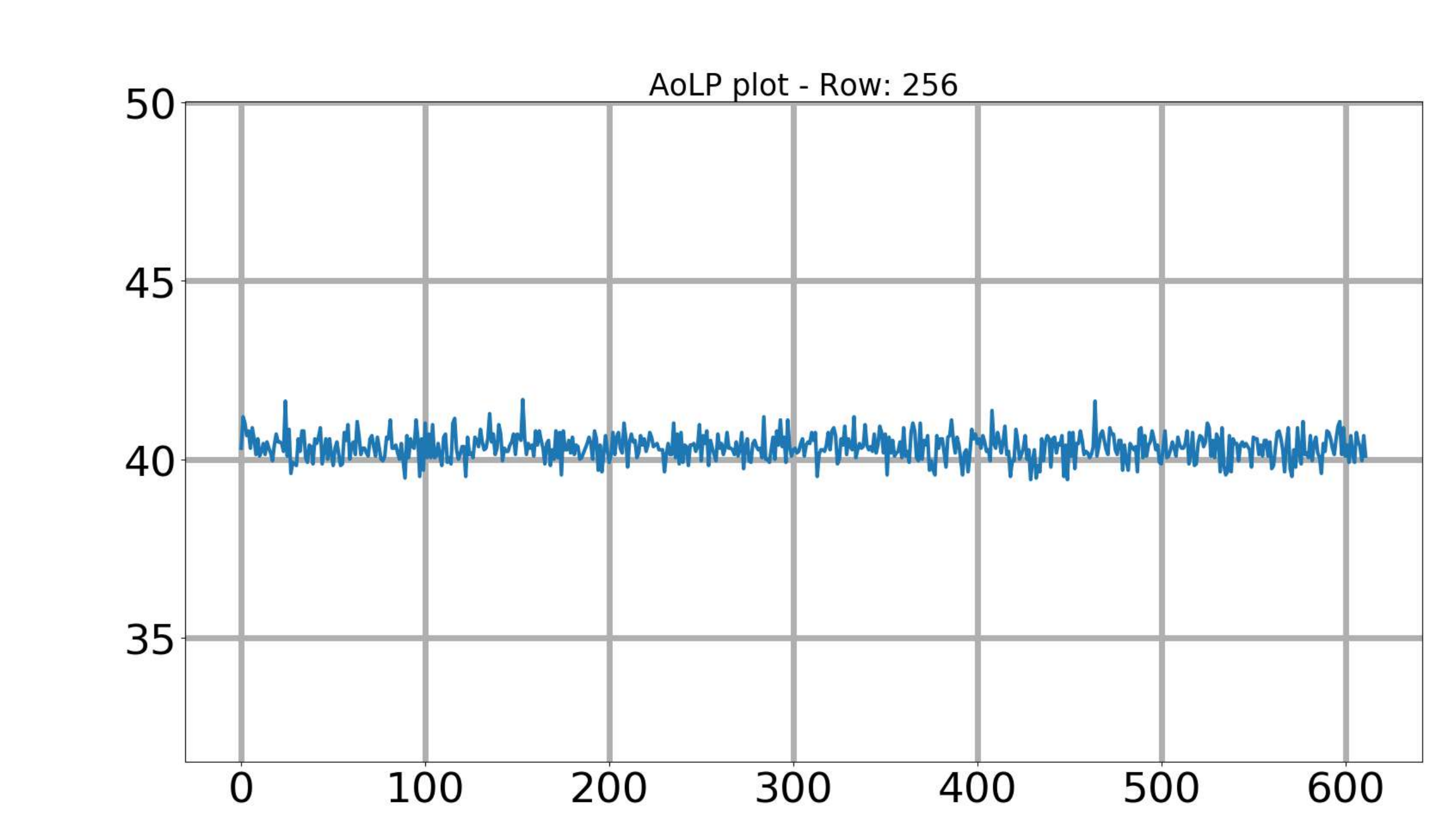} &
    \igw{0.24\textwidth}{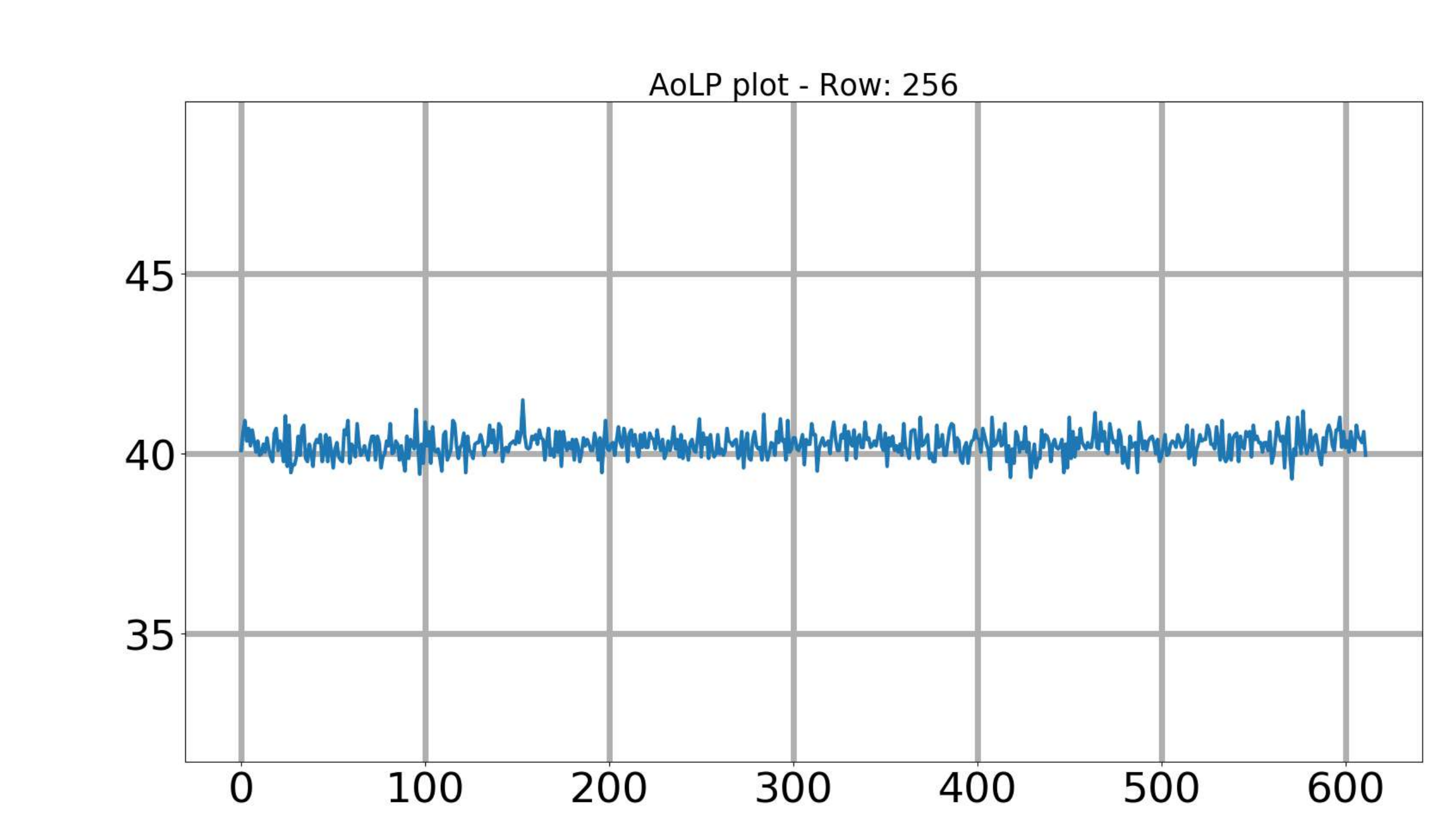} \\
    (b) & (e) & (h) & (k)\\

    \igw{0.24\textwidth}{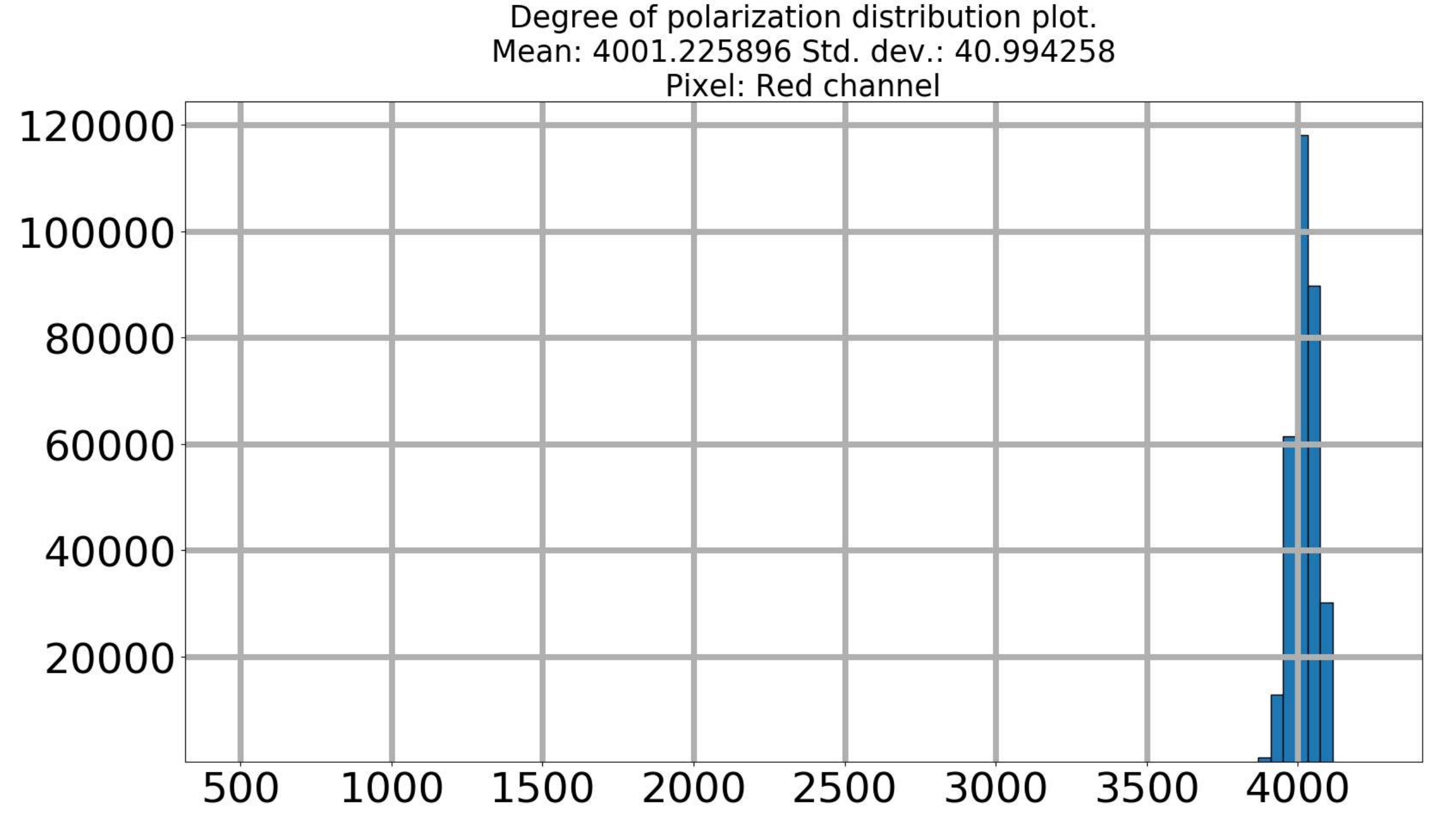} &
    \igw{0.24\textwidth}{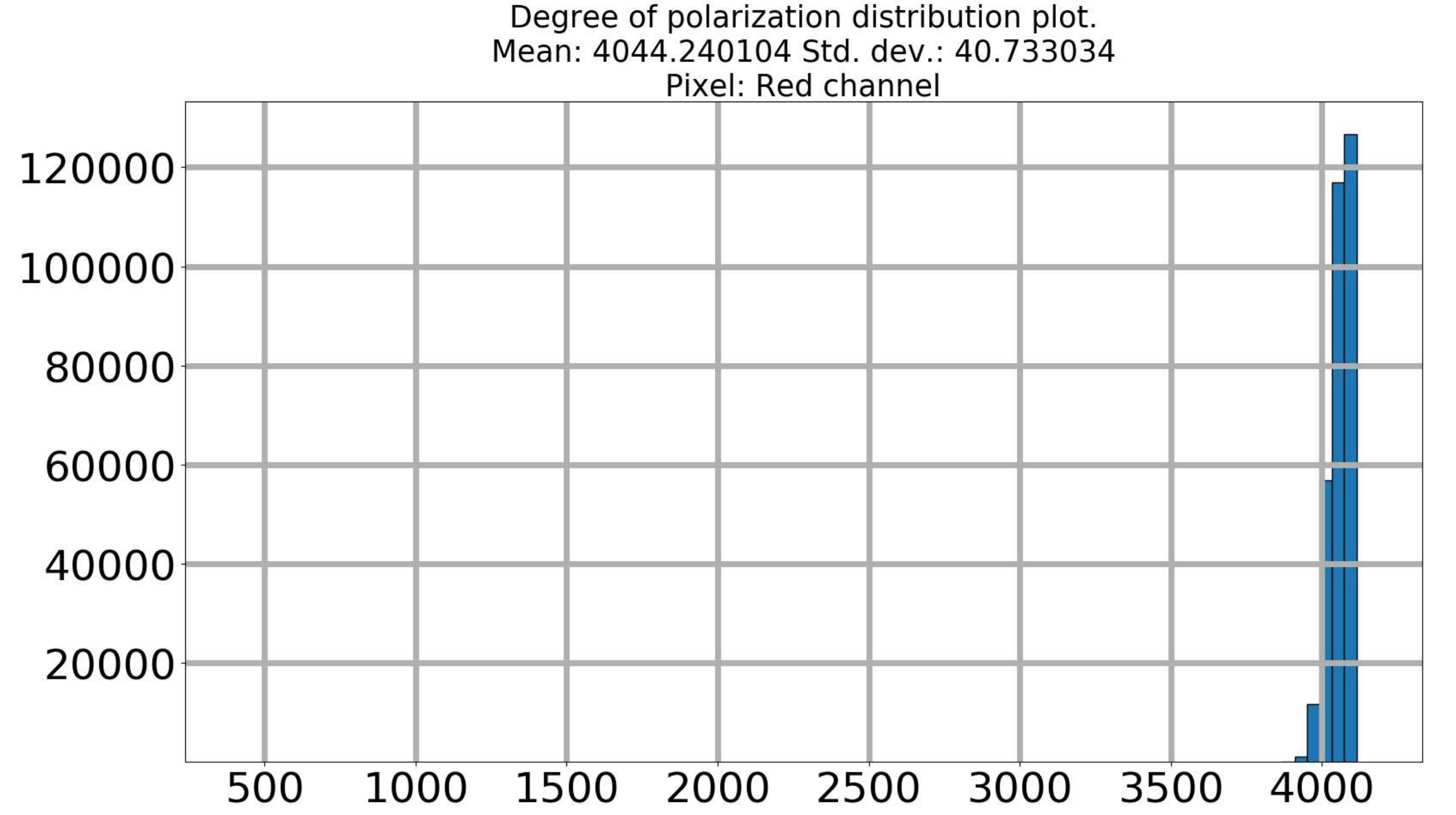} &
    \igw{0.24\textwidth}{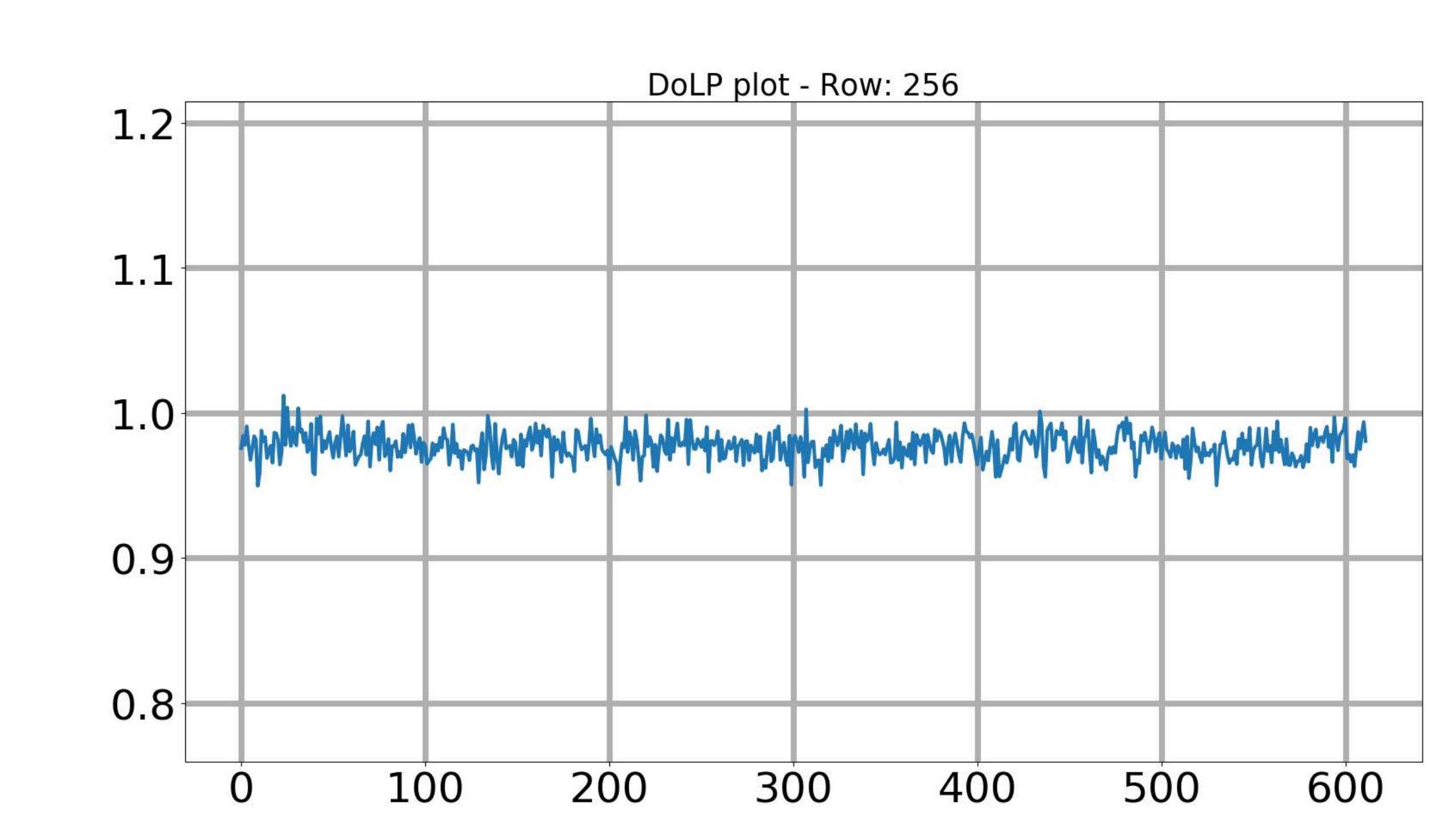} &
    \igw{0.24\textwidth}{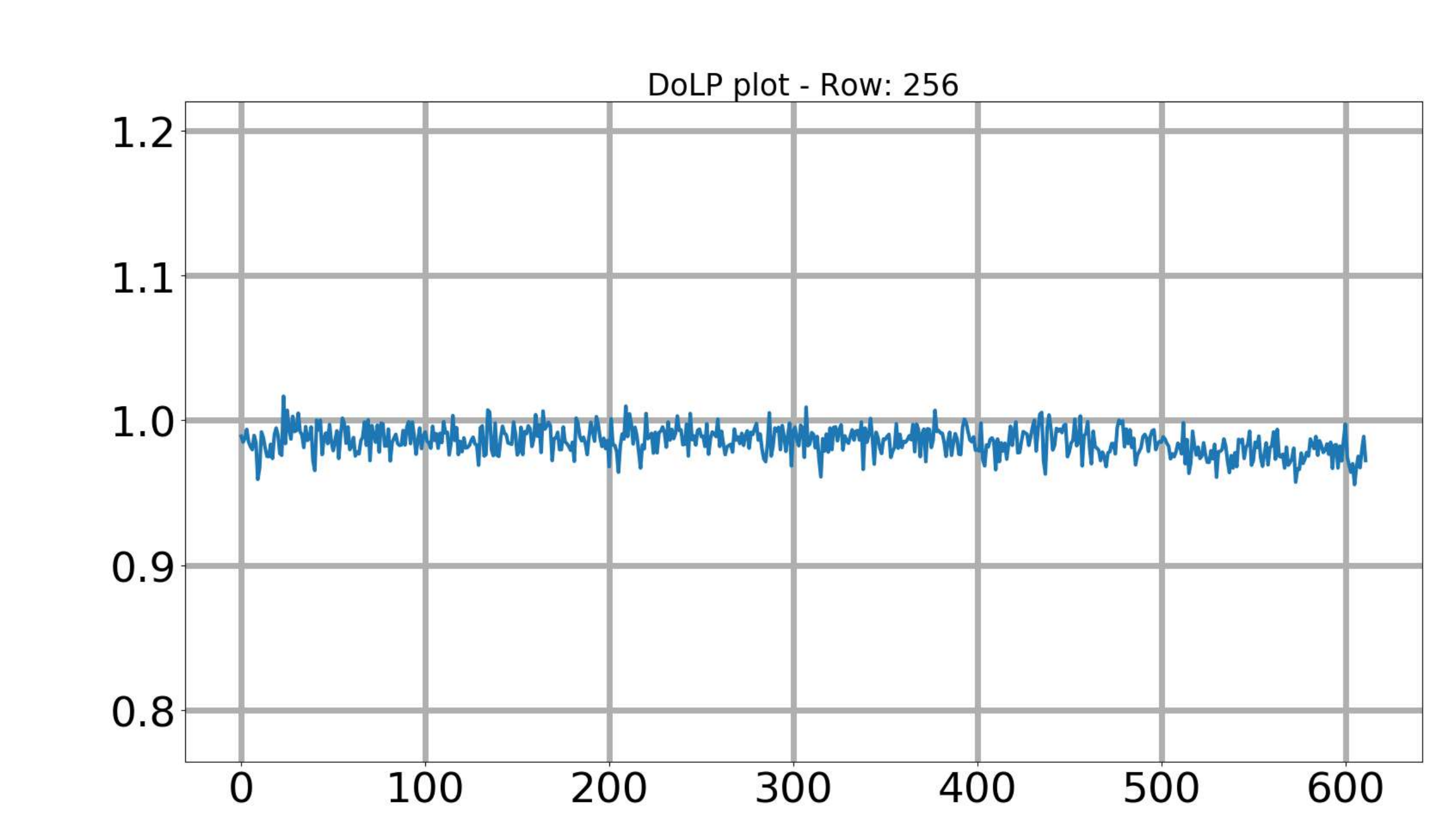} \\
    (c) & (f) & (i) & (l)\\
    \end{tabular}
    \caption[Calibrated camera quality evaluation plots.]
    {\changes{Calibrated camera quality evaluation. All the plots are computed before
    (first and third columns) and after (second and fourth columns) calibration.
    (a) and (d) Intensity histograms.
    (b) and (e) AoLP histograms.
    (c) and (f) DoLP histograms.
    (g) and (j) Intensity measurements plots over a single row.
    (h) and (k) AoLP measurements plots over a single row.
    (i) and (l) DoLP measurements plots over a single row.}}
    \label{fig:AllCalibPlots}
\end{figure*}

\changes{Among the available plots, the histograms of the intensity, the DoLP and the
AoLP of the incident light can be computed from the image currently displayed by
the software. This information allows to evaluate the quality of the calibration
results. After correction, these three parameters (intensity, DoLP and AoLP)
will have a narrower distribution than the uncalibrated case. These histograms
before and after calibration are represented in} \Cref{fig:AllCalibPlots}
\changes{(a) to (f). They have been computed with the same sample image, mentioned
before, of a polarized light with an AoLP of $40^\circ$.}

\changes{A real-time plot of these three parameters (intensity, DoLP, and AoLP) can be
done for a given row of pixels of the sensor. These graphs are displayed in the
last two columns of} \Cref{fig:AllCalibPlots}, \changes{and they correspond to the
measurements before and after applying the calibration. These plots illustrate
the vignetting effect and show how calibration can reduce its impact over the
three polarization parameters. It is important to note that this correction is
achieved since the pixel model used considers the polarization parameters of the
pixel, and not only the unbalanced sensing gain. A simple gain correction will
only affect the intensity image, but not the AoLP nor the DoLP images.}

\begin{figure*}[!t]
    \centering
    \begin{tabular}{ccc}
    \igw{0.3\textwidth}{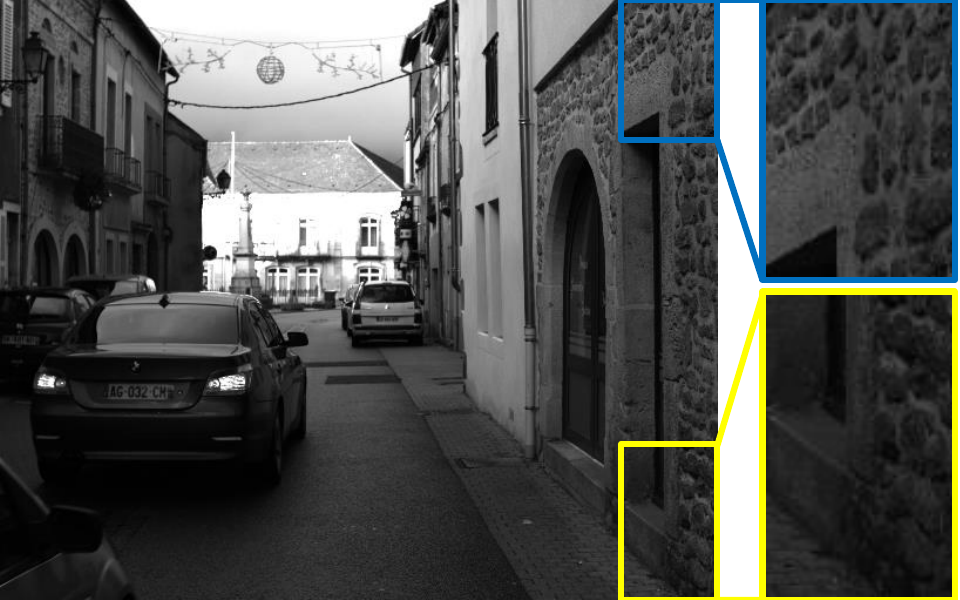} &
    \igw{0.3\textwidth}{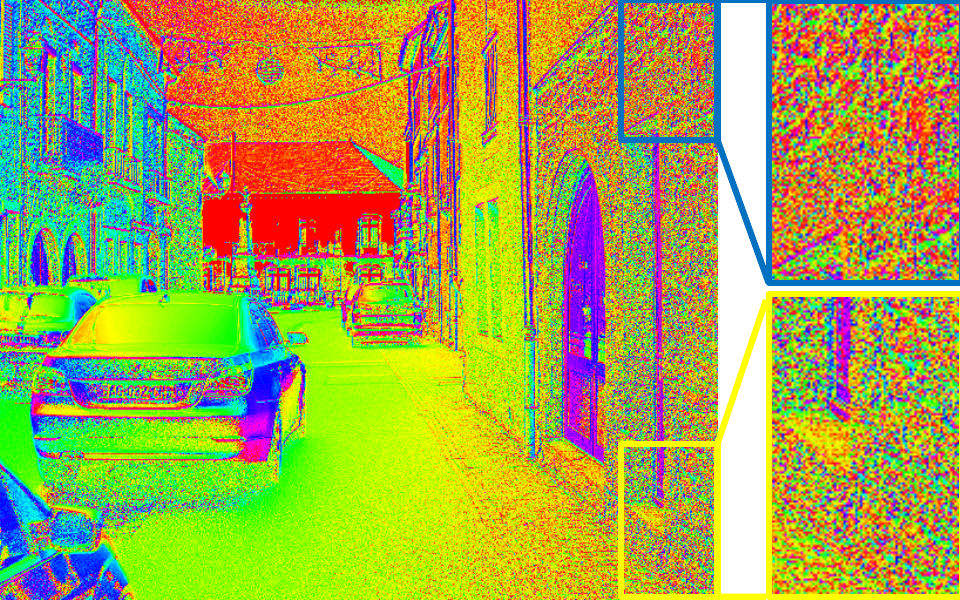} &
    \igw{0.3\textwidth}{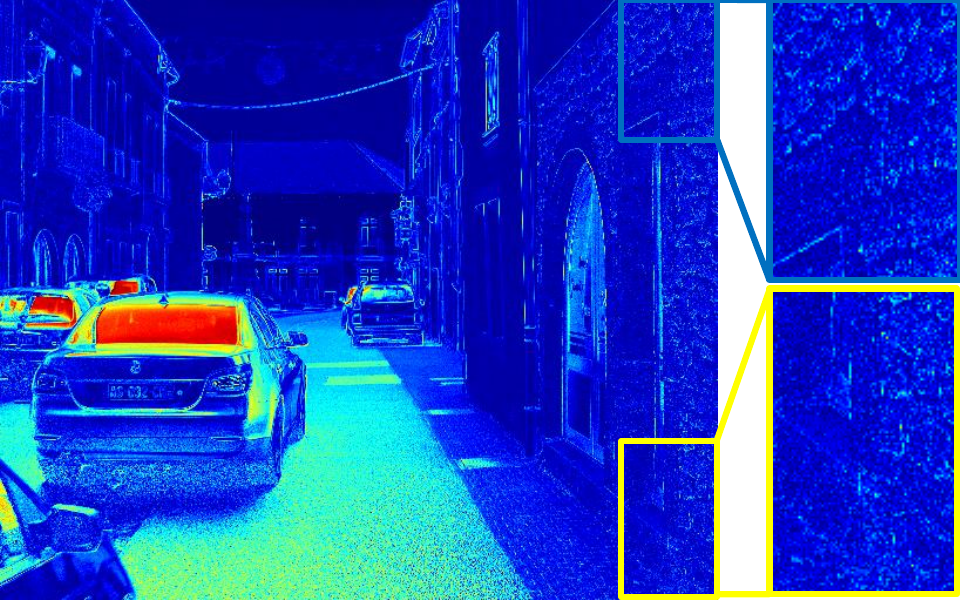} \\
    (a) & (b) & (c) \\
    \igw{0.3\textwidth}{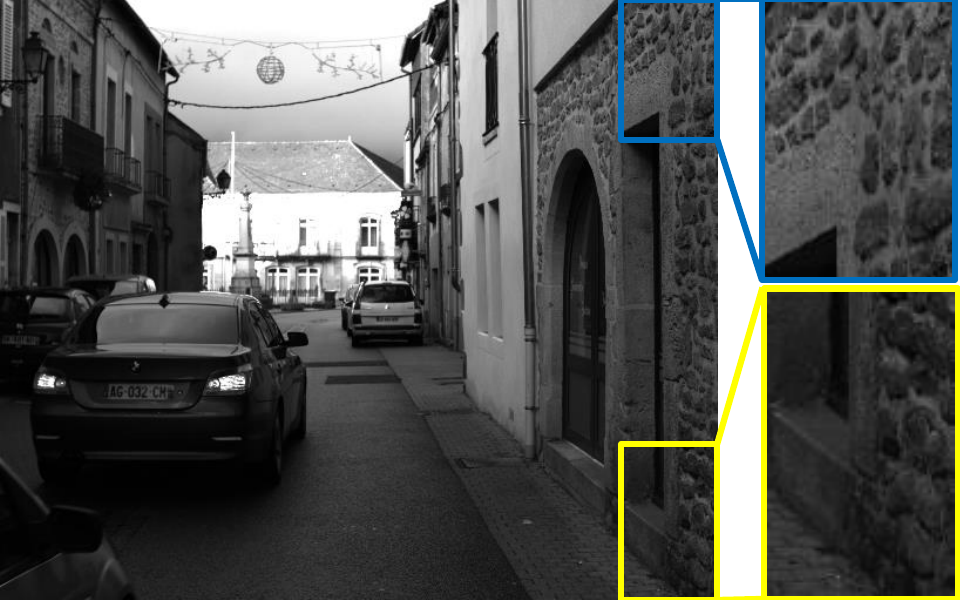} &
    \igw{0.3\textwidth}{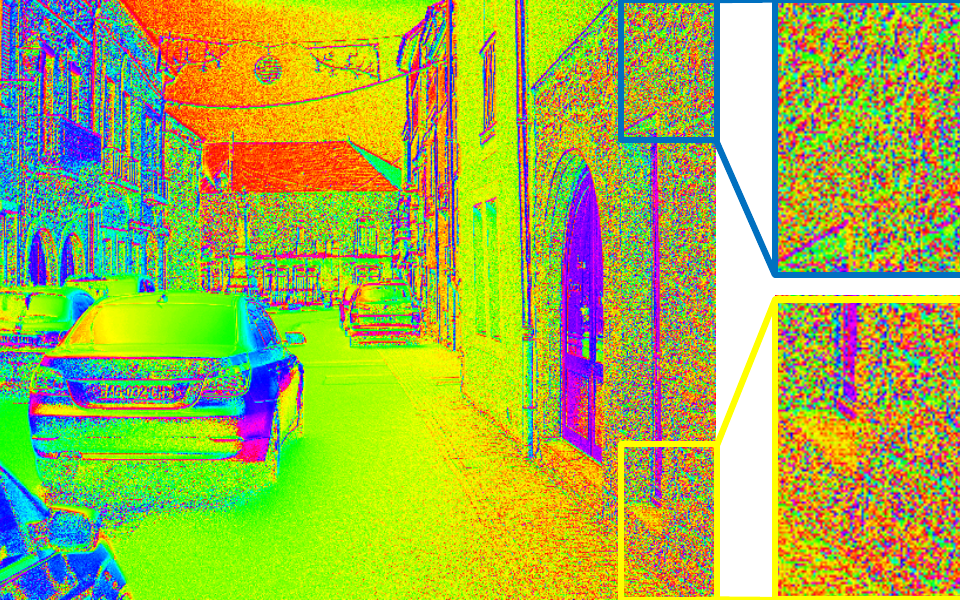} &
    \igw{0.3\textwidth}{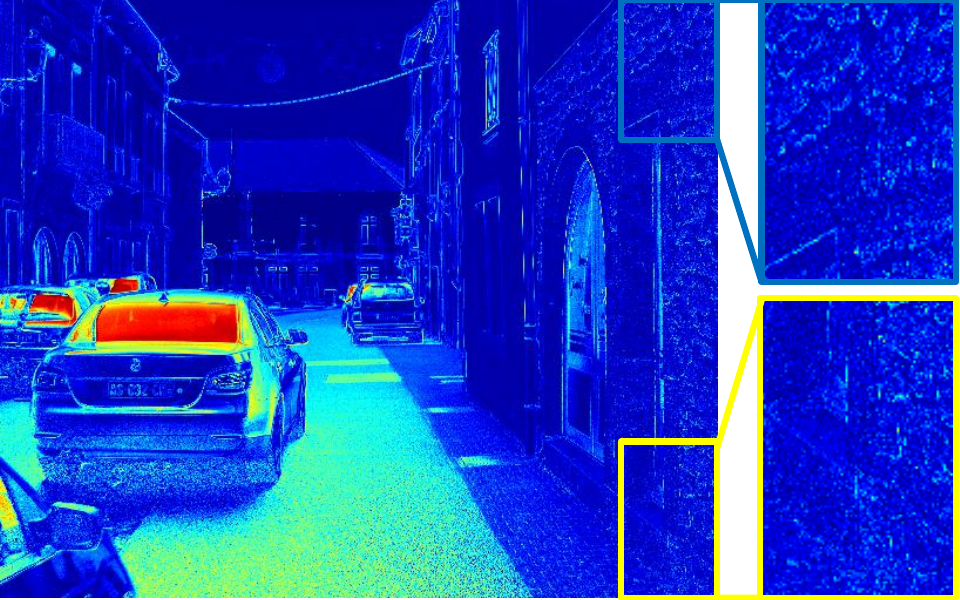} \\
    (d) & (e) & (f) \\
    \end{tabular}
    \caption[Calibration effects over the polarization images]
    {Calibration effects over the polarization images, for the red channel.
    Top row: uncalibrated images.
    Bottom row: calibrated images.
    (a) (d) Total intensity.
    (b) (e) Angle of linear polarization, colored with the HSV palette.
    (c) (f) Degree of linear polarization, colored with the Jet palette.
    }
    \label{fig:CalibUrban}
\end{figure*}

\changes{Finally, the consequence of applying the calibration can be observed in the
reference urban scene image. The effects of the calibration over the intensity,
AoLP, and DoLP images are shown in} \Cref{fig:CalibUrban}. \changes{A zoom has been
added to the top and bottom right areas of the image where it is possible
to see that the vignetting effect has been corrected.}
\changes{Particularly from these images, the contribution of the calibration can be
observed mostly in the AoLP and the intensity images. In the scene, there are
several walls that act as planes. Thus, they should reflect the same AoLP, i.e.,
they should have the same color. This happens only after calibration, mainly in
the building situated in the far region of the image, and in the walls on both
sides of the road. In the intensity image, since the pixel model includes a gain
factor, the vignetting effect is also corrected with this system, making the
darker areas in the borders to be as bright as the center of the image.}

\subsection{\label{sec:PolaAppsExplanation}Polarization processing algorithms}
\begin{figure*}[!t]
    \centering
    \begin{tabular}{cccc}
        \igw{0.24\textwidth}{images-software/05_orig_filtering.pdf} &
        \igw{0.24\textwidth}{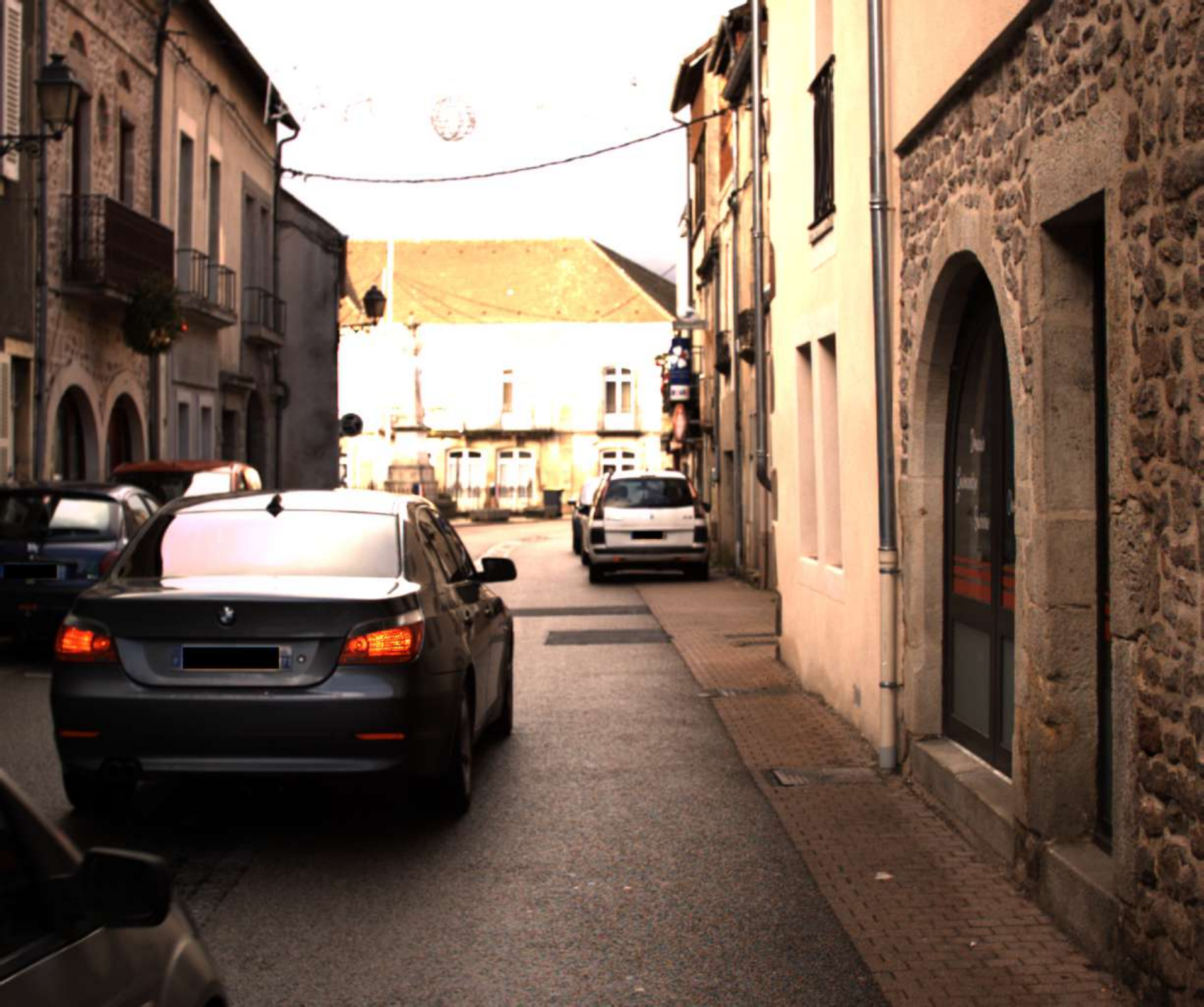} &
        \igw{0.24\textwidth}{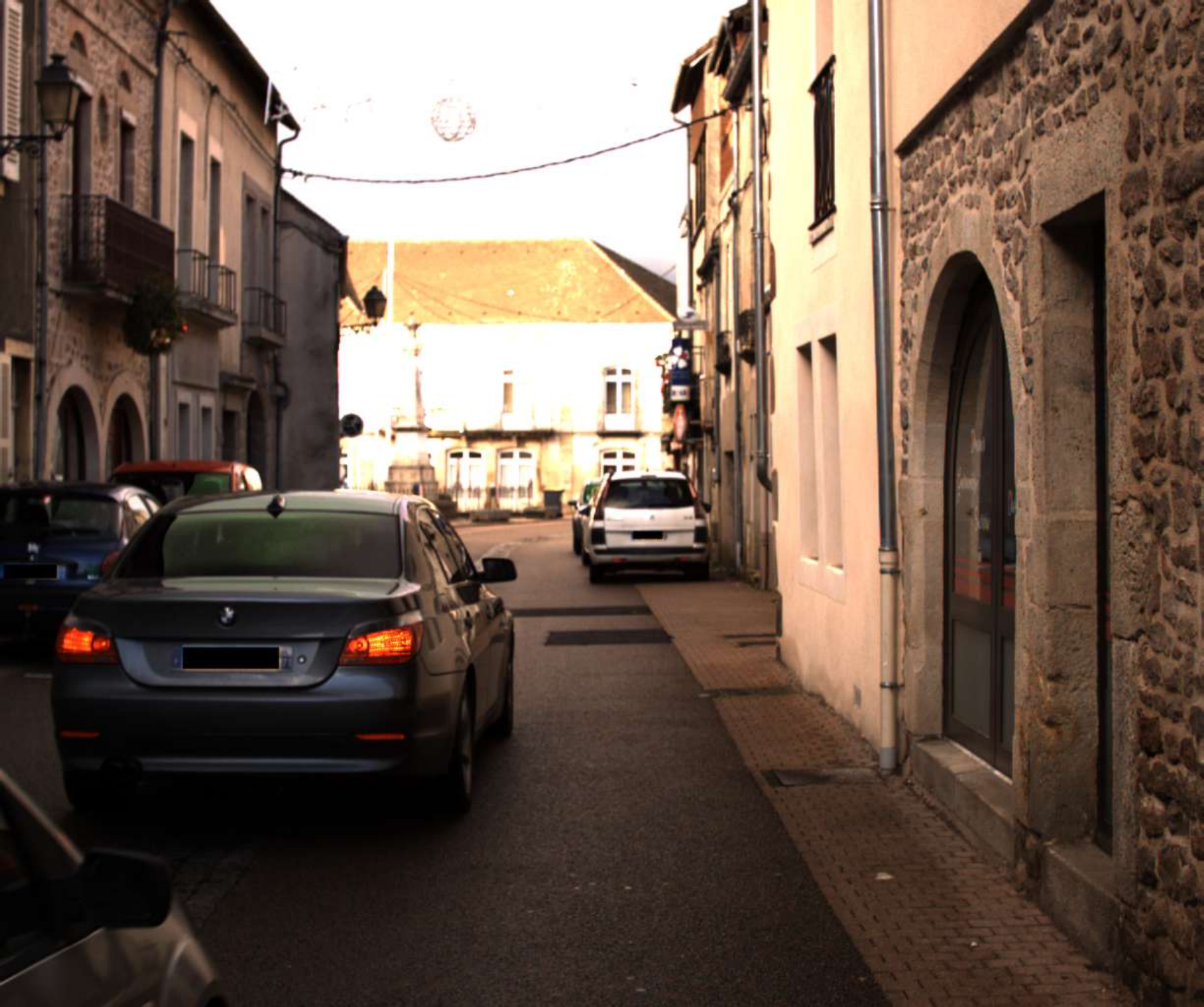} &
        \igw{0.24\textwidth}{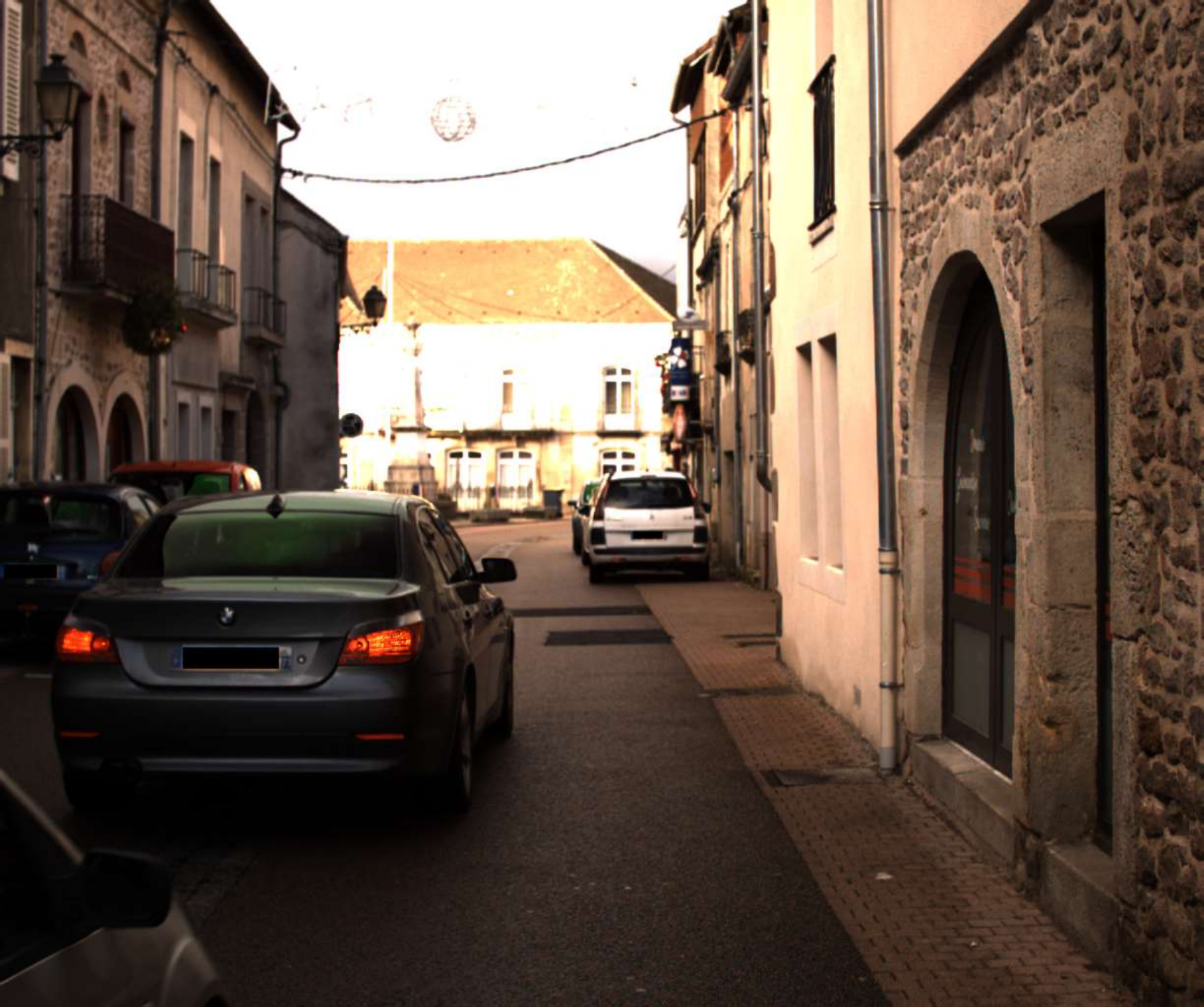} \\
        (a) & (b) & (c) & (d) \\
    \end{tabular}
    \caption[Specularity removal application results.]
    {Specularity removal application results. (a) Unfiltered, total
        intensity. (b) (c) Filtered image by simulating a linear polarization filter
        oriented at $60^\circ$, and $120^\circ$, respectively. (d) Polarized specularity
        removal by filtering with the degree of linear polarization.}
    \label{fig:ReflectionRemovalApps}
\end{figure*}

Differently from \Cref{sec:PolarimetricProcMod} where basic polarimetric
\changes{operations} can be done, in this module two applications of the polarization
concepts are implemented. The first one is the simulated polarization filter. As
explained in \Cref{sec:PolaIntro}, each super-pixel allows the computation of the
Stokes vector $\mathbf{S}=\left[S_{0},S_{1},S_{2}\right]^T$ of the incident
light. Now, let us consider a light, described by the Stokes vector
$\mathbf{S}_{in}$ that passes through a linear polarizer. The filter is
oriented at an angle $\theta$ and modeled by a Mueller matrix $\mathbf{M}$, as
explained in \Cref{sec:PolaIntro}. Therefore, the effects of a linear polarizer
in front of a normal camera can be simulated by computing \cref{eq:usedEq}. Thus
in this functionality, the inputs are a raw image from the camera and the
orientation of the filter $\theta$ that one would like to simulate. Then, this
algorithm returns two images: the input image, and the filtered image after
applying \cref{eq:usedEq} to all the super-pixels. This functionality is
commonly used in photography to remove annoying polarized reflections from the
environment. In a real system with a conventional RGB camera, the filter is
physically placed on top of the lens, and turned until the reflection is
removed. With this software, the filter and its effects are simulated after the
image has been captured, and the exact angle for reflection removal can be
found. This orientation is equal to the AoLP of the incident light shifted by 90
degrees. \changes{As an example, the results of simulating a linear filter with the
angles $\theta=60^\circ$ and $\theta=120^\circ$ are shown in}
\Cref{fig:ReflectionRemovalApps} \changes{(b) and (c), respectively.}
\changes{Note that the choice of an angle or another will either reinforce or erase
the reflections on the windshields.}

The second functionality of this module is the polarized
specularity removal. It is an extension of the simulated polarization filter,
explained above. In the previous case, all the pixels are affected by a single
polarization filter. But, in a scene, there might be several objects that
produce this type of reflection, with different AoLP. To erase them all at once,
let us consider the Stokes vector of the observed light. This vector can be
split into two other Stokes vectors: one that represents totally unpolarized
light and another that represents a totally linearly polarized light
$\mathbf{S=S_{unpol}+S_{pol}}$ such as:
\begin{equation}
    \left[\begin{array}{c}
        S_{0} \\
        S_{0} \rho \cos\left(2\phi\right) \\
        S_{0} \rho \sin\left(2\phi\right)
    \end{array}\right]=\left(1-\rho\right)\left[\begin{array}{c}
        S_{0} \\
        0 \\
        0
    \end{array}\right] + \rho\left[\begin{array}{c}
        S_{0} \\
        S_{0}\cos\left(2\phi\right) \\
        S_{0}\sin\left(2\phi\right)
    \end{array}\right].
\end{equation}

Removing the polarized reflection means erasing the component corresponding to
$\mathbf{S_{pol}}$. This is equivalent to computing the $\mathbf{S_{unpol}}$
vector. This functionality returns two images: the input and the filtered
images, both demosaiced. In contrast to the previous case, this functionality
does not require the user to enter an angle to each filter. The filtering is
done based on the measured DoLP at each super-pixel. \changes{The results of the
filtered image is shown in} \Cref{fig:ReflectionRemovalApps} \changes{(d). One can
note that most reflections from the shiny surfaces, such as the windshield, the
road and the door glasses, have been removed.}

\section{Discussion}
Recent technological advances have made possible new sensors that allow for
capturing RGB and polarization information in a single snapshot. This is an
important step forward in the development of the polarization field since it
reduces the movement constraints and the acquisition time for this modality.
This is one of the main reasons why many research works leveraging polarization
have been published in recent years with impressive results. From a global point
of view, all of the works make use of the AoLP and the DoLP as principal added
features to an RGB system. The Fresnel equations are used as geometrical
constraints to better guide an optimization algorithm or a data-driven model
training towards more consistent results. This last point can be used in two
ways: either by computing the normal maps out of the Fresnel equations and
providing these maps as input cues for a deep learning network or by using these
relationships to compute an energy function that will be later optimized. Going
further, some works \cite{Lei_2020_CVPR,dehazingpaper} use the raw
measurements to avoid adding non-linearities to the relationships between the
polarization state and the intensity measurements. However, there is still much
to explore, since one of the reasons why polarization has not been widely
adopted in the computer vision field is the required controlled acquisition
conditions. In other words, many applications have been developed in laboratory
conditions, or in environments where there are no perturbations that might
affect the measurements. One example of this is in shape from polarization
algorithms \cite{Ichikawa_2021_CVPR,Fukao_2021_CVPR,Zhu_2019_CVPR}, where
often the properties of the object being reconstructed are known beforehand.
Even though experiments can be done in outdoor conditions, the fact of having a
single object to analyze prevents their generalization to a more generic scene
in which several objects, with different properties, are present at the same
time. Another example of the generalization limitation are algorithms developed
under simulated conditions, such as the underwater image enhancement algorithm
\CustomCite{Li}{UnderwaterRestoration}. In this particular case, the results
show a clear improvement with respect to RGB systems, but the experiments have
been done in a small-sized water tank, with known light, and no perturbations
such as waves or the presence of other objects inside the tank. These
constraints are not in line with the requirements of a real-world underwater
situation, leaving its actual applicability unknown. It is worth noting that for
data-driven algorithms, it is hard to make acquisitions of large number of
images of highly reflective surfaces. These types of surfaces are known to be
challenging for RGB-only algorithms and the polarization information can provide
valuable cues. However, to obtain a ground truth regarding the shape or the
depth, sensors such as the Microsoft Kinect or LiDAR must be used, but the
reflective or transparent nature of the objects can result in inaccurate
measurements by the sensors. Therefore, there is still challenges in handling
these types of situations. As also mentioned in the different categories of
\Cref{sec:relWorks}, there is a lack of common standards to share data,
acquisition and basic signal processing tools to process polarization
information. In this context, we have proposed a software library toolkit that
can serve to the objectives of facilitating the acquisition, application
deployment, and comparison of techniques with polarization cues. We expect this
toolkit to be a step further to help promoting the polarization imaging field in
less constrained conditions.
\section{Conclusion}
In this paper, we have done a comprehensive review of recent works that have
leveraged the polarization modality in robotics and computer vision fields,
summarizing some of their main contributions and limitations. The rich
information brought by RGB-polarization cameras has allowed a huge improvement
in the analysis of scenes where RGB-only systems generally fail. The number of
works in the recent years has increased due to the advent of new and cheap
RGB-polarization sensors, and data-driven algorithms. These new devices
facilitate data acquisition, in less constrained environments, and in real-time
with a single snapshot. However, there is still a long way to the vision
community to fully exploit the potential and the benefits of this modality.
There is no general rule to follow to extend most existing conventional RGB
algorithms to RGB-polarization cameras. As discussed before, just providing the
polarization information as an additional input to an algorithm is unlike to
provide better results. A good understanding of the physics and principles of
light polarization is required to be able to get the most out of this modality.
Furthermore, the lack of standards, common processing tools, large-scale
benchmarks limits the development of methods that extend the results obtained
for traditional vision systems since it is often not possible to train and
compare different approaches over a common dataset. To allow a greater number of
researchers to easily and rapidly acquire and analyze polarimetric images, we
have developed an open-source software library toolkit that makes available
commonly used processing algorithms. This software has been carefully developed
to allow easy maintenance and the addition of new features. We hope this work
will encourage other researchers to contribute to this field and to participate
in the expansion of the functionalities included in the presented toolkit, with
the aim of making it accessible to anyone who wants to explore polarization.

\subsection*{Acknowledgements}
We thank the Conseil Régional de Bourgogne-Franche-
Comté for providing financial support for this research through the project ANER MOVIS, and the French government for the Plan France Relance initiative which also provided fundings via the European Union under contract ANR-21-PRRD-0047-01. We would also like to thank the IDRIS CNRS for granting us access to their High Performance
Computing resources (Grant No. 2021-AD011013154).

\end{spacing}
\begin{spacing}{0.95}    

\bibliography{11_refs}
\bibliographystyle{spiejour}

\end{spacing}
\end{document}